%% file: main.tex
\documentclass{article}

\usepackage[final,nonatbib]{neurips_data_2022}

\usepackage[utf8]{inputenc} 
\usepackage[T1]{fontenc}    
\usepackage{hyperref}       
\usepackage{url}            
\usepackage{booktabs}       
\usepackage{amsfonts}       
\usepackage{nicefrac}       
\usepackage{microtype}      
\usepackage{xcolor}         
\usepackage{enumitem}       
\usepackage{amsmath,bm,bbm}
\usepackage{pifont}
\usepackage[perpage]{footmisc}
\usepackage{makecell}

\usepackage{amsmath}        

\usepackage{mathtools}

\usepackage{graphicx}
\usepackage{caption}
\usepackage{subcaption}
\usepackage{amssymb}
\usepackage{bbding}
\usepackage{multirow}
\usepackage[export]{adjustbox}

\usepackage{soul}

\input{dfn}

\title{\system: Anomaly Detection Benchmark}

\author{%
  Songqiao Han$^{1,\ast}$, 
  Xiyang Hu$^{2,\ast}$,
  Hailiang Huang$^{1,\ast}$, 
  Minqi Jiang$^{1,\ast}$,
  Yue Zhao$^{2,}$\thanks{All authors contribute equally and are listed alphabetically. Direct questions to Minqi Jiang and Yue Zhao.}\\
  $^1$ Shanghai University of Finance and Economics 
  $^2$ Carnegie Mellon University \\
  \texttt{\{han.songqiao,hlhuang\}@shufe.edu.cn},
  \texttt{\{2020310191\}@live.sufe.edu.cn},\\
  \texttt{\{xiyanghu,zhaoy\}@cmu.edu}
}

\begin{document}

\maketitle

\begin{abstract}
\input{00abstract}
\end{abstract}

\vspace{-0.1in}
\section{Introduction}
\label{sec:intro}
\input{1intro}
\vspace{-0.1in}

\section{Related Work}
\label{sec:related}
\input{2related}
\vspace{-0.1in}

\section{\system: AD Benchmark Driven by Research and Application Needs}
\label{sec:settings}
\input{3setting}

\section{Experiment Results and Analyses}
\label{sec:experiments}
\input{4exp}

\section{Conclusions and Future Work}
\label{sec:conclusion}
\input{5discussion}




\section*{Aknowledgement}
We briefly describe the authors' contributions. \textit{Problem scoping}: M.J., Y.Z., S.H, X.H., and H.H.; \textit{Experiment and Implementation}: M.J. and Y.Z.; \textit{Result Analysis}: M.J., Y.Z., and X.H.; \textit{Paper Drafting}: M.J., Y.Z., S.H, X.H., and H.H.; \textit{Paper Revision}: M.J., Y.Z., S.H, and X.H.

We thank anonymous reviewers for their insightful feedback and comments. We appreciate the suggestions of Xueying Ding, Kwei-Herng (Henry) Lai, Meng-Chieh Lee, Ninghao Liu, Yuwen Yang, and Allen Zhu.
Y.Z. is partly supported by the Norton Graduate Fellowship.


\bibliographystyle{abbrv}
\small
\bibliography{reference}

\clearpage
\newpage

\clearpage
\newpage
\normalsize 
\input{appendix}

\end{document}

%% file: dfn.tex
\usepackage{xspace}
\newcommand{\system}{{ADBench}\xspace}

\newcommand{\nexps}{{98,436}\xspace}

\newcommand{\ndatasets}{{57}\xspace}
\newcommand{\nmodels}{{30}\xspace}

\newcommand{\ndatasetssimple}{{47}\xspace}
\newcommand{\ndatasetscomplex}{{10}\xspace}

\newcommand{\nunsup}{{14}\xspace}
\newcommand{\nsemi}{{7}\xspace}
\newcommand{\nsup}{{9}\xspace}

\newcommand{\bal}{\begin{align}}
\newcommand{\ean}{\end{align}}
\newcommand{\bit}{\begin{itemize}}
\newcommand{\eit}{\end{itemize}}
\newcommand{\ben}{\begin{enumerate}}
\newcommand{\een}{\end{enumerate}}
\newcommand{\beq}{\begin{equation}}
\newcommand{\eeq}{\end{equation}}

\newcommand{\cmark}{\ding{51}}%
\newcommand{\xmark}{\ding{55}}%

\newcommand\crule[3][black]{\textcolor{#1}{\rule{#2}{#3}}}

\definecolor{aliceblue}{rgb}{0.91796875, 0.91796875, 0.999}
\definecolor{aliceyellow}{rgb}{0.999, 0.96875, 0.91796875}
\definecolor{alicegreen}{rgb}{0.91796875, 0.95703125, 0.91796875}

%% file: 00abstract.tex
Given a long list of anomaly detection algorithms developed in the last few decades, how do they perform with regard to (\textit{i}) varying levels of supervision, (\textit{ii}) different types of anomalies, and (\textit{iii}) noisy and corrupted data? In this work, we answer these key questions by conducting (to our best knowledge) the most comprehensive \underline{a}nomaly \underline{d}etection \underline{bench}mark with \nmodels algorithms on \ndatasets benchmark datasets, named \system.
Our extensive experiments (\nexps in total) identify meaningful insights into the role of supervision and anomaly types, and unlock future directions for researchers in algorithm selection and design. With \system, researchers can efficiently conduct comprehensive and fair evaluations for newly proposed methods on the datasets (including our contributed ones from natural language and computer vision domains) against the existing baselines. To foster accessibility and reproducibility, we fully open-source \system and the corresponding results.

%% file: 1intro.tex
\vspace{-0.1in}
Anomaly detection (AD), which is also known as outlier detection, is a key machine learning (ML) task with numerous applications, including anti-money laundering \cite{lee2020autoaudit}, rare disease detection \cite{zhao2021suod}, social media analysis \cite{yu2017ring,zhao2020multi}, and intrusion detection \cite{lazarevic2003comparative}. AD algorithms aim to identify data instances that deviate significantly from the majority of data objects \cite{grunau2020adapting,qiu2022latent,rebjock2021online,shen2020timeseries}, and numerous methods have been developed in the last few decades \cite{aggarwal2017introduction,revisiting_time_series,liu2022pygod,liu2021event,pang2021deep,schirrmeister2020understanding,wang2020further,zhao2019pyod}. Among them, the majority are designed for tabular data (i.e., no time dependency and graph structure). Thus, we focus on the \textit{tabular} AD algorithms and datasets in this work. 

Although there are already some benchmark and evaluation works for tabular AD \cite{campos2016evaluation,comparative_evaluation,meta_analysis,goldstein2016comparative,realistic_synthetic_data}, they generally have the limitations as follows: 
(\textit{i}) primary emphasis on unsupervised methods only without including emerging (semi-)supervised AD methods; 
(\textit{ii}) limited analysis of the algorithm performance concerning anomaly types (e.g., local vs. global);
\textit{(iii)} the lack of analysis on model robustness (e.g.,  noisy labels and irrelevant features);
\textit{(iv)} the absence of using statistical tests for algorithm comparison; and 
\textit{(v)} no coverage of more complex CV and NLP datasets, which have attracted extensive attention nowadays.

\begin{figure*}[!t]
    \centering
    \includegraphics[width=\linewidth]{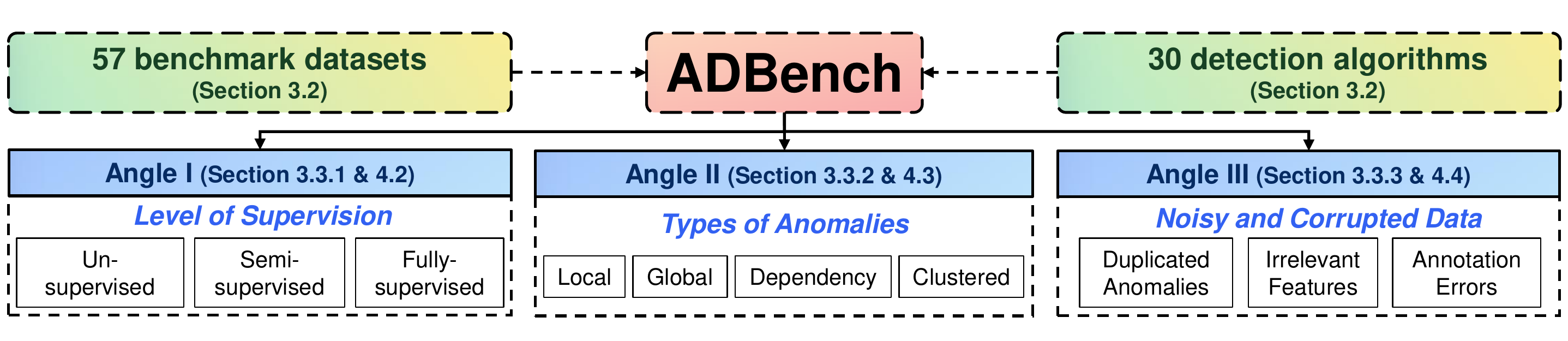}
    \vspace{-0.2in}
    \caption{The design of the proposed \system is driven by research and application needs.}
    \vspace{-0.25in}
    \label{fig:flowchart}
\end{figure*}

To address these limitations, we design (to our best knowledge) the most comprehensive tabular \underline{a}nomaly \underline{d}etection \underline{bench}mark called \system. 
By analyzing both research needs and deployment requirements in the industry, we design the experiments with three major angles in anomaly detection (see \S \ref{subsec:angles}): (\textit{i}) the availability of supervision (e.g., ground truth labels) by including \nunsup unsupervised, \nsemi semi-supervised, and \nsup supervised methods; (\textit{ii}) algorithm performance under different types of anomalies by simulating the environments with four types of anomalies; and  (\textit{iii}) algorithm robustness and stability under three settings of data corruptions. 
Fig.~\ref{fig:flowchart} provides an overview of \system.

\textbf{Key takeaways}: Through extensive experiments, we find 
(\textit{i}) surprisingly none of the benchmarked unsupervised algorithms is statistically better than others, emphasizing the importance of algorithm selection;
(\textit{ii}) with merely 1\% labeled anomalies, most semi-supervised methods can outperform the best unsupervised method, justifying the importance of supervision;
(\textit{iii}) in controlled environments, we observe that the best unsupervised methods for specific types of anomalies are even better than semi- and fully-supervised methods, revealing the necessity of understanding data characteristics;
(\textit{iv}) semi-supervised methods show potential in achieving robustness in noisy and corrupted data, possibly due to their efficiency in using labels and feature selection.
See \S \ref{sec:experiments} for additional results and insights. 

We summarize the primary contributions of \system as below:
\setlist{nolistsep}
\begin{enumerate}[leftmargin=*]
    \item \textbf{The most comprehensive AD benchmark}. 
    \system examines \nmodels detection algorithms' performance on \ndatasets benchmark datasets (of which \ndatasetssimple are existing ones and we create \ndatasetscomplex). 
    \item \textbf{Research and application-driven benchmark angles}. By analyzing the needs of research and real-world applications, we focus on three critical comparison angles: availability of supervision, anomaly types, and algorithm robustness under noise and data corruption.
    \item \textbf{Insights and future directions for researchers and practitioners}. With extensive results, we show the necessity of algorithm selection, and the value of supervision and prior knowledge.
    \item \textbf{Fair and accessible AD evaluation}. We open-source \system with BSD-2 License at \url{https://github.com/Minqi824/ADBench}, for benchmarking newly proposed methods.
\end{enumerate}


%% file: 2related.tex
\vspace{-0.1in}

\subsection{Anomaly Detection Algorithms}
\label{subsec:algorithms}
\vspace{-0.1in}

\textbf{Unsupervised Methods by Assuming Anomaly Data Distributions}.
\textit{Unsupervised AD methods are proposed with different assumptions of data distribution} \cite{aggarwal2017introduction}, e.g., anomalies located in low-density regions, and their performance depends on the agreement between the input data and the algorithm assumption(s).
Many unsupervised methods have been proposed in the last few decades \cite{aggarwal2017introduction,bergman2019classification,pang2021deep,unifying_shallow_deep,zhao2019pyod}, which can be roughly categorized into shallow and deep (neural network) methods. The former often carries better interpretability, while the latter handles large, high-dimensional data better. Please see Appx. \S \ref{appendix:subsec:unsupervised}, recent book \cite{aggarwal2017introduction}, and surveys \cite{pang2021deep,unifying_shallow_deep} for additional information. 

\textbf{Supervised Methods by Treating Anomaly Detection as Binary Classification}.
\textit{With the accessibility of full ground truth labels (which is rare), supervised classifiers may identify known anomalies at the risk of missing unknown anomalies}. 
Arguably, there are no specialized supervised anomaly detection algorithms, and people often use existing classifiers for this purpose \cite{aggarwal2017introduction,vargaftik2021rade} such as Random Forest \cite{liaw2002classification} and neural networks \cite{lecun2015deep}. 
One known risk of supervised methods is that ground truth labels are not necessarily sufficient to capture all types of anomalies during annotation. These methods are therefore limited to detecting unknown types of anomalies \cite{aggarwal2017introduction}. 
Recent machine learning books \cite{DBLP:books/sp/Aggarwal18,goodfellow2016deep} and scikit-learn \cite{pedregosa2011scikit} may serve as good sources of supervised ML methods.

\textbf{Semi-supervised Methods with Efficient Use of Labels.} \textit{Semi-supervised AD algorithms can capitalize the supervision from partial labels, while keeping the ability to detect unseen types of anomalies.}
To this end, some recent studies investigate using partially labeled data for improving detection performance and leveraging unlabeled data to facilitate representation learning. For instance, some semi-supervised models are trained only on normal samples, and detect anomalies that deviate from the normal representations learned in the training process \cite{GANomaly, Skip-GANomaly,ALAD}. 
In \system, semi-supervision mostly refers to \textit{incomplete label learning} in weak-supervision (see \cite{zhou2018brief}). More discussions on semi-supervised AD are deferred to Appx. \S \ref{appendix:subsec:semisupervised}.
\vspace{-0.1in}

\input{tables/benchmark}

\subsection{Existing Datasets and Benchmarks for Tabular AD} \label{subsec:benchmarks}

\textbf{AD Datasets in Literature}.
Existing benchmarks mainly evaluate 
a part of the datasets derived from the ODDS Library \cite{Rayana2016},
DAMI Repository \cite{campos2016evaluation},
ADRepository \cite{pang2021deep}, 
and Anomaly Detection Meta-Analysis Benchmarks \cite{meta_analysis}. 
In \system, we include almost all publicly available datasets,
and 
add larger datasets adapted from CV and NLP domains, for a more holistic view.
See details in \S \ref{subsec:algo_datasets}.

\textbf{Existing Benchmarks}.
There are some notable works that take effort to benchmark AD methods on tabular data, e.g., \cite{campos2016evaluation,comparative_evaluation,meta_analysis,unifying_shallow_deep,realistic_synthetic_data} (see Appx. \ref{appendix:benchmarks}). How does \system differ from them?

First, previous studies mainly focus on benchmarking the shallow unsupervised AD methods. Considering the rapid advancement of ensemble learning and deep learning methods, we argue that a comprehensive benchmark should also consider them.
Second, most existing works only evaluate public benchmark datasets and/or some fully synthetic datasets; we organically incorporate both of them to unlock deeper insights.
More importantly, existing benchmarks primarily focus on direct performance comparisons, while the settings may not be sufficiently complex to understand AD algorithm characteristics.
We strive to address the above issues in \system, and illustrate the main differences between the proposed \system and existing AD benchmarks in Table \ref{tab:benchmark comparison}.

Also,
``anomaly detection'' is an overloaded term;
there are AD benchmarks for time-series \cite{revisiting_time_series,lavin2015evaluating,paparrizos2022tsb}, graph \cite{liu2022benchmarking}, CV \cite{akcay2022anomalib,chan2021segmentmeifyoucan, zheng2022benchmarking} and NLP \cite{rahman2021information}, but they are different from tabular AD in nature.
\vspace{-0.1in}

\subsection{Connections with Related Fields and Other Opportunities}
\label{subsec:novelty_detection}

While \system focuses on the AD tasks, we note that there are some closely related problems, including out-of-distribution (OOD) detection \cite{yangopenood, yang2021generalized}, novelty detection \cite{markou2003novelty, pimentel2014review}, and open-set recognition (OSR) \cite{geng2020recent, mahdavi2021survey}. 
Uniquely, AD usually 
does not assume the train set is anomaly-free, while other related tasks may do. 
Some methods designed for these related fields, e.g., OCSVM \cite{scholkopf1999support}, can be used for AD as well;
future benchmark can consider including: (\textit{i}) OOD methods: MSP \cite{hendrycks17baseline}, energy-based EBO \cite{liu2020energy}, and Mahalanobis distance-based MDS \cite{lee2018simple}; (\textit{ii}) novelty detection methods: OCGAN \cite{perera2019ocgan} and Adversarial One-Class Classifier \cite{sabokrou2018adversarially}; and (\textit{iii}) OSR methods: OpenGAN \cite{kong2021opengan} and PROSER \cite{zhou2021learning}. See \cite{salehi2021unified} for deeper connections and differences between AD and these fields.

We consider saliency detection (SD) \cite{fan2018salient,fan2021re} and camouflage detection (CD) \cite{fan2020camouflaged} as good inspirations and applications of AD tasks. 
Saliency detection identifies important regions in the images, where explainable AD algorithms \cite{pang2021toward}, e.g., FCDD \cite{liznerski2020explainable}, may help the task.
Camouflage detection finds concealed objects in the background, e.g., camouflaged anomalies blurred with normal objects \cite{ma2021comprehensive}, 
where camouflage-resistant AD methods \cite{dou2020enhancing}
help detect concealed objects (that look normal but are abnormal).
Future work can explore the explainability of detected objects in AD.

%% file: tables/benchmark.tex
\begin{table}[!t]
  \centering
  \caption{Comparison among \system and existing benchmarks, where \system comprehensively includes the most datasets and algorithms, uses both benchmark and synthetic datasets, covers both shallow and deep learning (DL) algorithms, and considers multiple comparison angles.
  }
  \scalebox{0.74}{
    \begin{tabular}{l|cc|cc|cc|ccc}
    \toprule
    \multirow{2}*{\textbf{Benchmark}} &
    \multicolumn{2}{c}{\textbf{Coverage} (\S \ref{subsec:algo_datasets})} &
    \multicolumn{2}{c}{\textbf{Data Source}} &
    \multicolumn{2}{c}{\textbf{Algorithm Type}} &  \multicolumn{3}{c}{\textbf{Comparison Angle} (\S \ref{subsec:angles})} \\
    \cmidrule{2-10} & \# datasets & \# algo. & Real-world & Synthetic     & Shallow & DL  & Supervision & Types & Robustness \\
    \midrule
    Ruff et al. \cite{unifying_shallow_deep} &3&9 &\cmark       &\cmark     &\cmark         &\cmark  &\xmark &\cmark       &\xmark  \\
    Goldstein et al. \cite{goldstein2016comparative} &10&19 &\cmark       &\xmark   &\cmark       &\xmark  &\xmark      &\cmark &\xmark  \\
    Domingues et al. \cite{comparative_evaluation} &15&14 &\cmark       &\xmark     &\cmark      &\xmark  &\xmark   &\xmark    &\cmark  \\
    Soenen et al. \cite{soenen2021effect} &16&6 &\cmark       &\xmark   &\cmark       &\xmark &\xmark      &\xmark &\xmark  \\
    Steinbuss et al. \cite{realistic_synthetic_data} &19&4 &\xmark       &\cmark &\cmark &\xmark &\xmark   &\cmark    &\xmark  \\
    Emmott et al. \cite{meta_analysis} &19&8 &\cmark        &\cmark   &\cmark &\xmark &\xmark        &\cmark &\cmark  \\
    Campos et al. \cite{campos2016evaluation} &23&12 &\cmark       &\xmark   &\cmark      &\xmark  &\xmark       &\xmark &\xmark  \\
    \midrule
    \textbf{\system (ours)} &\ndatasets&\nmodels & \textcolor{blue}{\cmark}      & \textcolor{blue}{\cmark} & \textcolor{blue}{\cmark}     & \textcolor{blue}{\cmark}  & \textcolor{blue}{\cmark}     &\textcolor{blue}{\cmark} & \textcolor{blue}{\cmark} \\
    \bottomrule
    \end{tabular}%
    }
  \label{tab:benchmark comparison}%
  \vspace{-0.2in}
\end{table}%

%% file: 3setting.tex
\vspace{-0.1in}

\subsection{Preliminaries and Problem Definition} \label{subsec:Preliminaries}
\textbf{Unsupervised AD} often presents a collection of $n$  samples  $\mathbf{X}=\left\{\boldsymbol{x}_{1}, \ldots, \boldsymbol{x}_{n}\right\} \in \mathbbm{R}^{n \times d}$, where each sample has $d$ features. Given the inductive setting, the goal is to train an AD model $M$ 
to output anomaly score $\mathbf{O} \coloneqq M(\mathbf{X}) \in \mathbbm{R}^{n\times 1}$, where higher scores denote for more outlyingness. In the inductive setting, we need to predict on $\mathbf{X}_\text{test} \in \mathbbm{R}^{m\times d}$, so to return  $\mathbf{O}_\text{test} \coloneqq M(\mathbf{X}_\text{test}) \in \mathbbm{R}^{m\times 1}$.

\textbf{Supervised AD} also has the (binary) ground truth labels of $\mathbf{X}$, i.e., $\mathbf{y} \in \mathbbm{R}^{n \times 1}$. A supervised AD model $M$ is first trained on $\{\mathbf{X}, \mathbf{y} \}$, and then returns anomaly scores for the $\mathbf{O}_\text{test} \coloneqq M(\mathbf{X}_\text{test})$.

\textbf{Semi-supervised AD} only has the partial label information $\mathbf{y}^l \in \mathbf{y}$
. The AD model $M$ is trained on the entire feature space  $\mathbf{X}$ with the partial label $\mathbf{y}^l$, i.e., $\{\mathbf{X}, \mathbf{y}^l \}$, and then outputs $\mathbf{O}_\text{test} \coloneqq M(\mathbf{X}_\text{test})$.

\textbf{Remark}. Irrespective of the types of underlying AD algorithms, the goal of \system is to understand  AD algorithms' performance under the inductive setting. Collectively, we refer semi-supervised and supervised AD methods as ``label-informed'' methods. Refer to \S \ref{subsec:exp_setting} for specific experiment settings.

\vspace{-0.1in}

\subsection{The Largest AD Benchmark with \nmodels Algorithms and \ndatasets Datasets}
\label{subsec:algo_datasets}
\vspace{-0.1in}
\begin{figure*}[!t]
    \centering
    \includegraphics[width=1\linewidth]{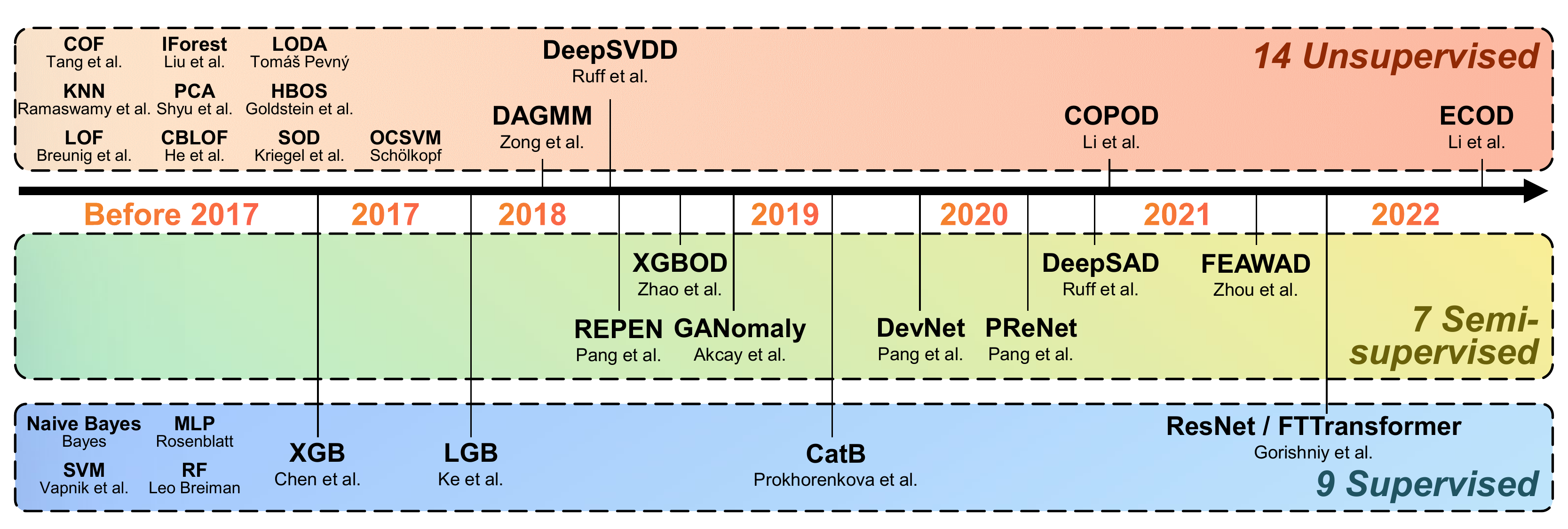}
    \vspace{-0.15in}
    \caption{
    \system covers a wide range of AD algorithms. See Appx. \ref{appendix:algorithms} for more details.
    }
    \vspace{-0.25in}
    \label{fig:algorithms}
\end{figure*}

\textbf{Algorithms}. 
Compared to the previous benchmarks, 
we have a larger algorithm collection with
(\textit{i}) the latest unsupervised AD algorithms like DeepSVDD \cite{DeepSVDD} and ECOD \cite{ECOD};
(\textit{ii}) SOTA semi-supervised algorithms, including DeepSAD \cite{DeepSAD} and DevNet \cite{devnet};
(\textit{iii}) latest network architectures like ResNet \cite{resnet} in computer vision (CV) and Transformer \cite{vaswani2017attention} in the natural language processing (NLP) domain---we adapt ResNet and FTTransformer models \cite{gorishniy2021revisiting} for tabular AD in the proposed \system; and 
(\textit{iv}) ensemble learning methods like LightGBM \cite{ke2017lightgbm}, XGBoost \cite{XGBoost}, and CatBoost \cite{prokhorenkova2018catboost} that have shown effectiveness in AD tasks \cite{vargaftik2021rade}. 
Fig. \ref{fig:algorithms} shows the 30 algorithms (\nunsup unsupervised, \nsemi semi-supervised, and \nsup supervised algorithms) evaluated in \system, where we provide more information about them in Appx. \ref{appendix:algorithms}.

\textbf{Algorithm Implementation}. Most unsupervised algorithms are readily available in our early work Python Outlier Detection (PyOD) \cite{zhao2019pyod}, and some supervised methods are available in scikit-learn \cite{pedregosa2011scikit} and corresponding libraries. Supervised ResNet and FTTransformer tailored for tabular data have been open-sourced in their original paper \cite{gorishniy2021revisiting}. We implement the semi-supervised methods and release them along with \system.

\textbf{Public AD Datasets}. 
In \system, we gather more than 40 benchmark datasets \cite{campos2016evaluation,meta_analysis,pang2021deep,Rayana2016}, for model evaluation, as shown in Appx. Table \ref{table:datasets}.
These datasets cover many application domains, including healthcare (e.g., disease diagnosis), audio and language processing (e.g., speech recognition), image processing (e.g., object identification), finance (e.g., financial fraud detection), etc. For due diligence, we keep the datasets where the anomaly ratio is below 40\% (Appx. Fig. \ref{fig:ratio}).

\textbf{Newly-added Datasets in \system}. 
Since most of these datasets are relatively small, we introduce \ndatasetscomplex more complex datasets from CV and NLP domains with more samples and richer features in \system (highlighted in Appx. Table \ref{table:datasets}).
Pretrained models are applied to extract data embedding from CV and NLP datasets to access more complex representations, which has been widely used in AD literature \cite{deecke2021transfer,manolache2021date,DeepSAD} and shown better results than using the raw features.
For NLP datasets, we use BERT \cite{BERT} pretrained on the BookCorpus and English Wikipedia to extract the embedding of the [CLS] token.
For CV datasets, we use ResNet18 \cite{resnet} pretrained on the ImageNet \cite{ImageNet} to extract the embedding after the last average pooling layer. Following previous works \cite{DeepSVDD, DeepSAD}, we set one of the multi-classes as normal, downsample the remaining classes to $5\%$ of the total instances as anomalies, and report the average results over all the respective classes. Including these originally non-tabular datasets helps to see whether tabular AD methods can work on CV/NLP data after necessary preprocessing.
See Appx. \ref{appendix:datasets} for more details on datasets.
\vspace{-0.1in}

\subsection{Benchmark Angles in \system}
\label{subsec:angles}
\vspace{-0.1in}

\subsubsection{Angle I: Availability of Ground Truth Labels (Supervision)}
\label{Angle I}
\vspace{-0.1in}
\textbf{Motivation}. 
As shown in Table \ref{tab:benchmark comparison}, existing benchmarks only focus on the unsupervised setting, i.e., none of the labeled anomalies is available. Despite, in addition to unlabeled samples, one may have access to a limited number of labeled anomalies in real-world applications, e.g., a few anomalies identified by domain experts or human-in-the-loop techniques like active learning \cite{aissa2016semi,GANomaly,kiran2018overview,zha2020meta}. Notably, there is a group of semi-supervised AD algorithms \cite{REPEN,devnet_explainable,PReNet,devnet,DeepSAD,AABiGAN, FEAWAD} that have not been covered by existing benchmarks. 

\textbf{Our design}: We first benchmark existing unsupervised anomaly detection methods, and then evaluate both semi-supervised and fully-supervised methods with varying levels of supervision following the settings in \cite{REPEN,devnet,FEAWAD} to provide a fair comparison. For example, labeled anomalies $\gamma_{l}=10\%$ means that $10\%$ anomalies in the train set are known while other samples remain unlabeled. The complete experiment results of un-, semi-, and full-supervised algorithms are presented in \S \ref{subsec:overall_performance}.
\vspace{-0.1in}

\subsubsection{Angle II: Types of Anomalies}
\label{subsub:types}
\vspace{-0.1in}
\textbf{Motivation}. While extensive public datasets can be used for benchmarking, they often consist of a mixture of different types of anomalies, making it challenging to understand the pros and cons of AD algorithms regarding specific types of anomalies \cite{gopalan2019pidforest,realistic_synthetic_data}. In real-world applications, one may know specific types of anomalies of interest. To better understand the impact of anomaly types, we create synthetic datasets based on public datasets by injecting specific types of anomalies to analyze the response of AD algorithms.

\textbf{Our design}: In \system, we create \textit{realistic} synthetic datasets from benchmark
datasets by injecting specific types of anomalies. 
Some existing works, such as PyOD \cite{zhao2019pyod}, generate fully synthetic anomalies by assuming their data distribution, which fails to create complex anomalies.
We follow and enrich the approach in \cite{realistic_synthetic_data} to generate ``realistic'' synthetic data; ours supports more types of anomaly generation.
The core idea is to build a generative model (e.g., Gaussian mixture model GMM used in \cite{realistic_synthetic_data}, Sparx \cite{zhang2022sparx}, and \system) using the normal samples from a benchmark dataset and discard its original anomalies as we do not know their types. Then, We could generate normal samples and different types of anomalies based on their definitions by tweaking the generative model. The generation of normal samples is the same in all settings if not noted, and we provide the generation process of four types of anomalies below (also see our codebase for details).

\begin{figure}[!t]
    \vspace{-0.1in}
    \begin{minipage}[c]{0.8\textwidth}
     \centering
     \begin{subfigure}[b]{0.242\textwidth}
         \centering
         \includegraphics[width=\textwidth]{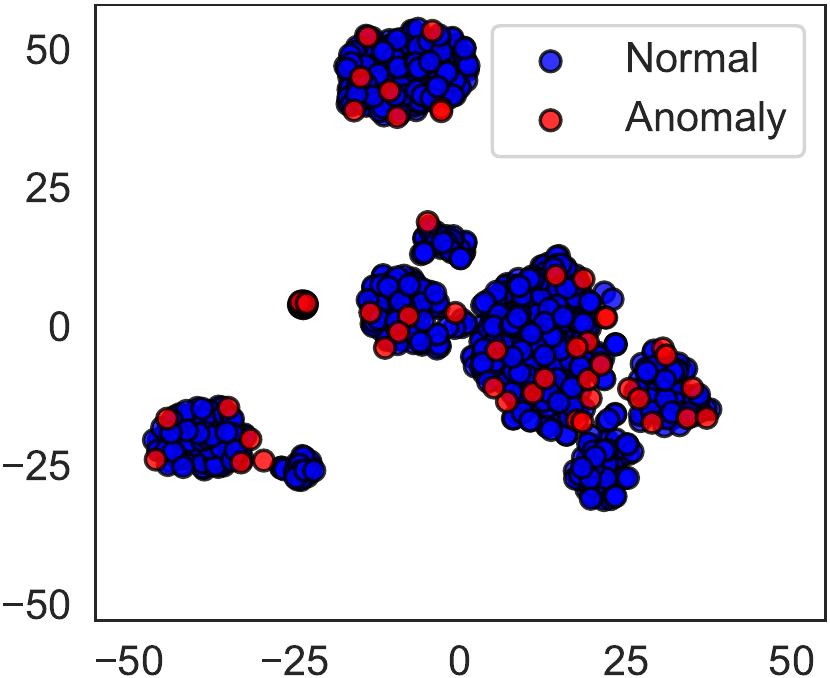}
         \caption{Local}
         \label{fig:demo_local}
     \end{subfigure}
     \hfill
     \begin{subfigure}[b]{0.242\textwidth}
         \centering
         \includegraphics[width=\textwidth]{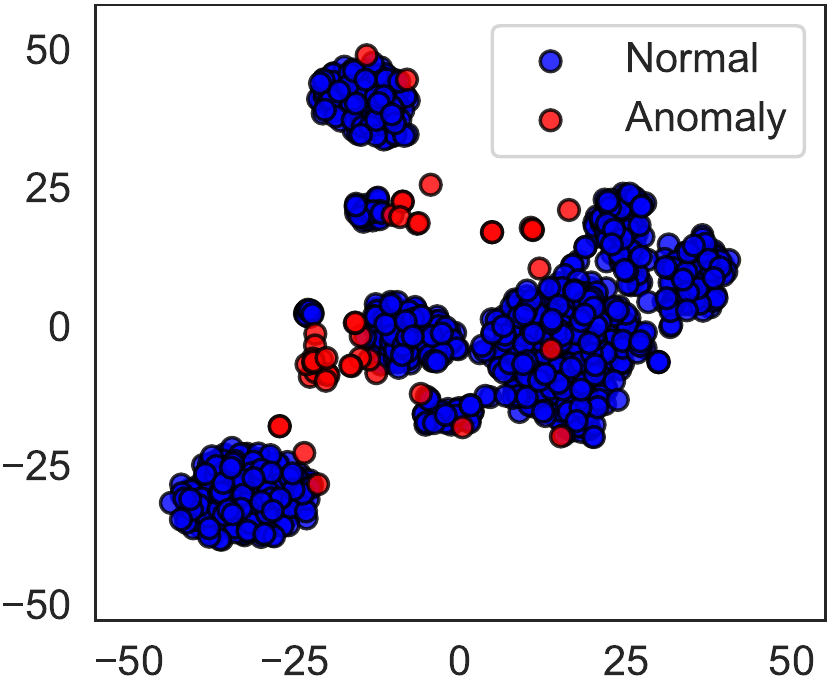}
         \caption{Global}
         \label{fig:demo_global}
     \end{subfigure}
     \hfill
     \begin{subfigure}[b]{0.242\textwidth}
         \centering
         \includegraphics[width=\textwidth]{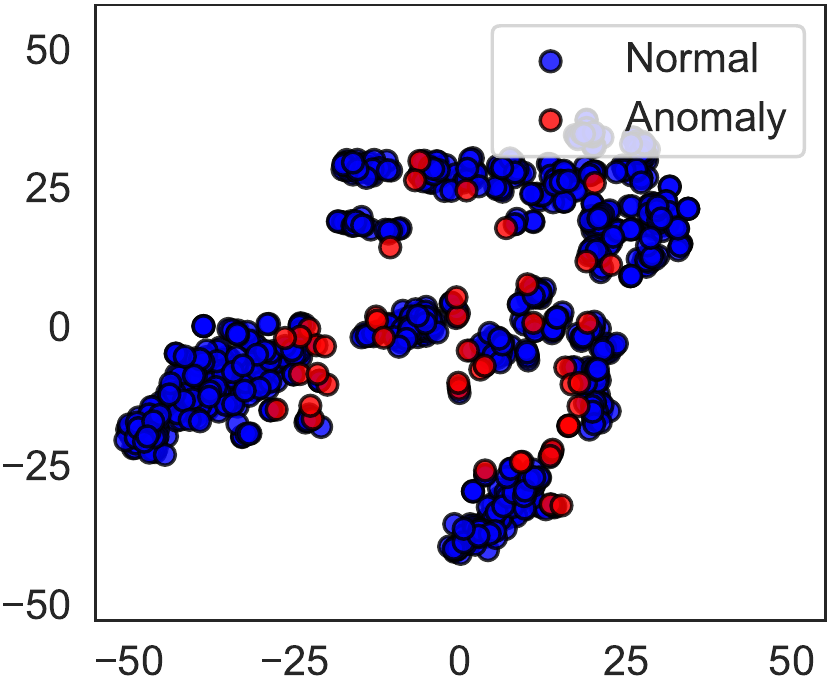}
         \caption{Dependency}
         \label{fig:demo_dependency}
     \end{subfigure}
     \hfill
     \begin{subfigure}[b]{0.242\textwidth}
         \centering
         \includegraphics[width=\textwidth]{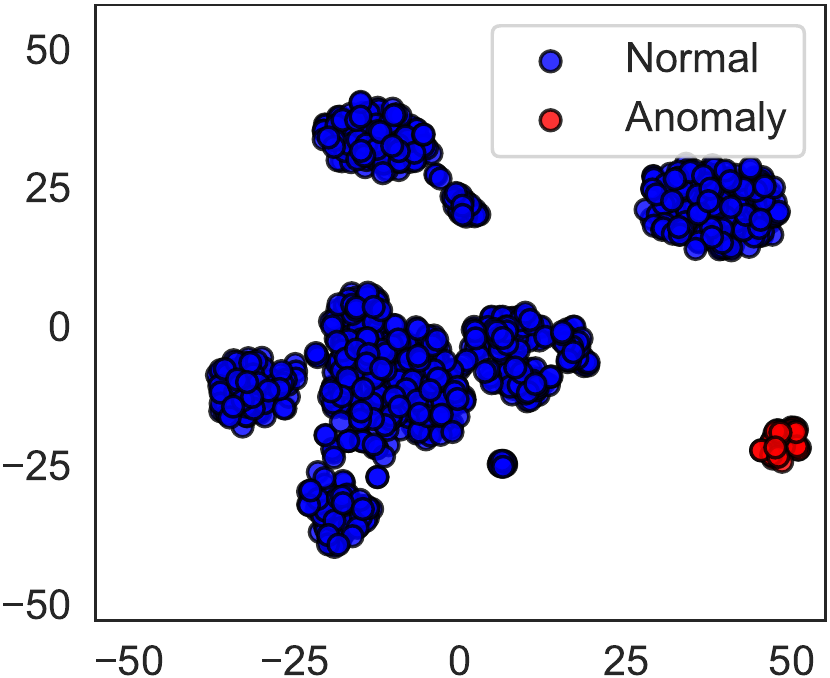}
         \caption{Clustered}
         \label{fig:demo_clusdtered}
     \end{subfigure}
    \end{minipage}
    \hfill
    \begin{minipage}[c]{0.18\textwidth}
     \caption{Illustration of four types of synthetic anomalies shown on Lymphography dataset. See the additional demo in Appx. Fig. \ref{fig:synthetic anomalies add}.}
     \label{fig:synthetic anomalies}
    \end{minipage}
    \vspace{-0.45in}
\end{figure}

\textbf{Definition and Generation Process of Four Types of Common Anomalies Used in \system}: 

\setlist{nolistsep}
\begin{itemize}[leftmargin=*,noitemsep]
\item \textbf{Local anomalies} refer to the anomalies that are deviant from their local neighborhoods \cite{LOF}. We follow the GMM procedure \cite{milligan1985algorithm,realistic_synthetic_data} to generate synthetic normal samples, and then scale the covariance matrix $\hat{\boldsymbol{\Sigma}}=\alpha \hat{\boldsymbol{\Sigma}}$ by a scaling parameter $\alpha=5$ to generate local anomalies.

\item \textbf{Global anomalies} are more different from the normal data \cite{huang2014physics}, generated from a uniform distribution $\text{Unif}\left(\alpha \cdot \min \left(\mathbf{X^{k}}\right), \alpha \cdot \max \left(\mathbf{X^{k}}\right)\right)$, where the boundaries are defined as the \textit{min} and \textit{max} of an input feature, e.g., $k$-th feature $\mathbf{X^{k}}$, and $\alpha=1.1$ controls the outlyingness of anomalies. 

\item \textbf{Dependency anomalies} refer to the samples that do not follow the dependency structure which normal data follow \cite{martinez2016fault}, i.e., the input features of dependency anomalies are assumed to be independent of each other. Vine Copula \cite{aas2009pair} method is applied to model the dependency structure of original data, where the probability density function of generated anomalies is set to complete independence by removing the modeled dependency (see \cite{martinez2016fault}).
We use Kernel Density Estimation (KDE) \cite{hastie2009elements} to estimate the probability density function of features and generate normal samples.

\item \textbf{Clustered anomalies}, also known as group anomalies \cite{lee2021gen}, 
exhibit similar characteristics \cite{meta_analysis, liu2022unsupervised}. We scale the mean feature vector of normal samples by $\alpha=5$, i.e., $\hat{\boldsymbol{\mu}}=\alpha \hat{\boldsymbol{\mu}}$, where $\alpha$ controls the distance between anomaly clusters and the normal, and use the scaled GMM to generate anomalies.
\end{itemize}

Fig.~\ref{fig:synthetic anomalies} shows 2-d t-SNE \cite{TSNE} visualization of the four types of synthetic outliers generated from Lymphography dataset, where they generally satisfy the expected characteristics. 
Local anomalies (Fig.~\ref{fig:demo_local}) are well overlapped with the normal samples.
Global anomalies (Fig.~\ref{fig:demo_global}) are more deviated from the normal samples and on the edges of normal clusters. 
The other two types of anomalies are as expected, with no clear dependency structure in Fig.~\ref{fig:demo_dependency} and having anomaly cluster(s) in Fig.~\ref{fig:demo_clusdtered}.
In \system, we analyze the algorithm performances under all four types of anomalies above (\S \ref{exp:types}). 

\vspace{-0.1in}

\subsubsection{Angle III: Model Robustness with Noisy and Corrupted Data} 
\label{Angle III}
\vspace{-0.1in}
\textbf{Motivation}. Model robustness has been an important aspect of anomaly detection and adversarial machine learning \cite{cai2021learned,du2021learning,fan2021does,kim2020adversarial,wu2021wider}. Meanwhile, the input data likely suffers from noise and corruption to some extent in real-world applications \cite{meta_analysis,gopalan2019pidforest,guha2016robust,pang2021homophily}.
However, this important view has not been well studied in existing benchmarks, and we try to understand this by evaluating AD algorithms under three noisy and corruption settings (see results in \S \ref{exp:robustness}):

\setlist{nolistsep}
\begin{itemize}[leftmargin=*,noitemsep]
\item \textbf{Duplicated Anomalies}. In many applications, certain anomalies likely repeat multiple times in the data for reasons such as recording errors \cite{kwon2018empirical}. The presence of duplicated anomalies is also called the ``anomaly masking'' \cite{gopalan2019pidforest,guha2016robust,liu2008isolation}, posing challenges to many AD algorithms \cite{campos2016evaluation}, e.g., the density-based KNN \cite{angiulli2002fast,ramaswamy2000efficient}. Besides, the change of anomaly frequency would also affect the behavior of detection methods \cite{meta_analysis}. Therefore, we simulate this setting by splitting the data into train and test set, then duplicating the anomalies (both features and labels) up to 6 times in both sets, and observing how AD algorithms change.

\item \textbf{Irrelevant Features}. Tabular data may contain irrelevant features caused by measurement noise or inconsistent measuring units \cite{chang2022data,gopalan2019pidforest}, where these noisy dimensions could hide the characteristics of anomaly data and thus make the detection process more difficult \cite{devnet_explainable,unifying_shallow_deep}. We add irrelevant features up to $50\%$ of the total input features (i.e., $d$ in the problem definition) by generating uniform noise features from $\text{Unif}\left(\min \left(\mathbf{X^{k}}\right), \max \left(\mathbf{X^{k}}\right)\right)$ of randomly selected $k$-th input feature $\mathbf{X^{k}}$ while the labels stay correct, and summarize the algorithm performance changes.

\item \textbf{Annotation Errors}. While existing studies \cite{devnet,DeepSAD} explored anomaly contamination in the unlabeled samples,
we further discuss the more generalized impact of label contamination on the algorithm performance, where the label flips \cite{nguyen2019self,zheng2021meta} between the normal samples and anomalies are considered (up to $50\%$ of total labels). Note this setting does not affect unsupervised methods as they do not use any labels. Discussion of annotation errors is meaningful since manual annotation or some automatic labeling techniques are always noisy while being treated as perfect.
\end{itemize}

%% file: 4exp.tex
\vspace{-0.1in}

We conduct \nexps experiments (Appx. \ref{appendix:exp_setting}) to answer \textbf{Q1} (\S \ref{subsec:overall_performance}): How do AD algorithms perform with varying levels of supervision? \textbf{Q2} (\S \ref{exp:types}): How do AD algorithms respond to different types of anomalies?  \textbf{Q3} (\S \ref{exp:robustness}): How robust are AD algorithms with noisy and corrupted data? 
In each subsection, we first present the key results and analyses (please refer to the additional points in Appx. \ref{appendix:exp_results}), and then propose a few open questions and future research directions.

\vspace{-0.1in}

\subsection{Experiment Setting}
\label{subsec:exp_setting}
\vspace{-0.1in}

\textbf{Datasets, Train/test Data Split, and Independent Trials}. As described in \S \ref{subsec:algo_datasets} and Appx. Table \ref{table:datasets}, \system includes \ndatasets existing and freshly proposed datasets, which cover different fields including healthcare, security, and more.
Although unsupervised AD algorithms are primarily designed for the transductive setting (i.e., outputting the anomaly scores on the input data only other than making predictions on the newcoming data), we adapt all the algorithms for the inductive setting to predict the newcoming data, which is helpful in applications and also common in popular AD library PyOD \cite{zhao2019pyod}, TODS \cite{lai2021tods}, and PyGOD \cite{liu2022pygod}. Thus, we use $70\%$ data for training and the remaining $30\%$ as the test set. We use stratified sampling to keep the anomaly ratio consistent. We repeat each experiment 3 times and report the average. Detailed settings are described in Appx. \ref{appendix:exp_setting}.

\textbf{Hyperparameter Settings}. For all the algorithms in  \system, we use their default hyperparameter (HP) settings in the original paper for a fair comparison. 
Refer to the Appx. \ref{appendix:exp_setting} for more information.

\textbf{Evaluation Metrics and Statistical Tests}. We evaluate different AD methods by two widely used metrics:  AUCROC (Area Under Receiver Operating Characteristic Curve) and AUCPR (Area Under Precision-Recall Curve) value\footnote{We present the results based on AUCROC and observe similar results for AUCPR; See Appx. \ref{appendix:exp_results} for all.}. Besides, the critical difference diagram (CD diagram) \cite{demvsar2006statistical,ismail2019deep} based on the Wilcoxon-Holm method is used for comparing groups of AD methods statistically ($p\leq 0.05$).

\begin{figure}[h]
     \centering
     \begin{subfigure}[t]{0.37\textwidth}
         \centering
         \includegraphics[width=\textwidth]{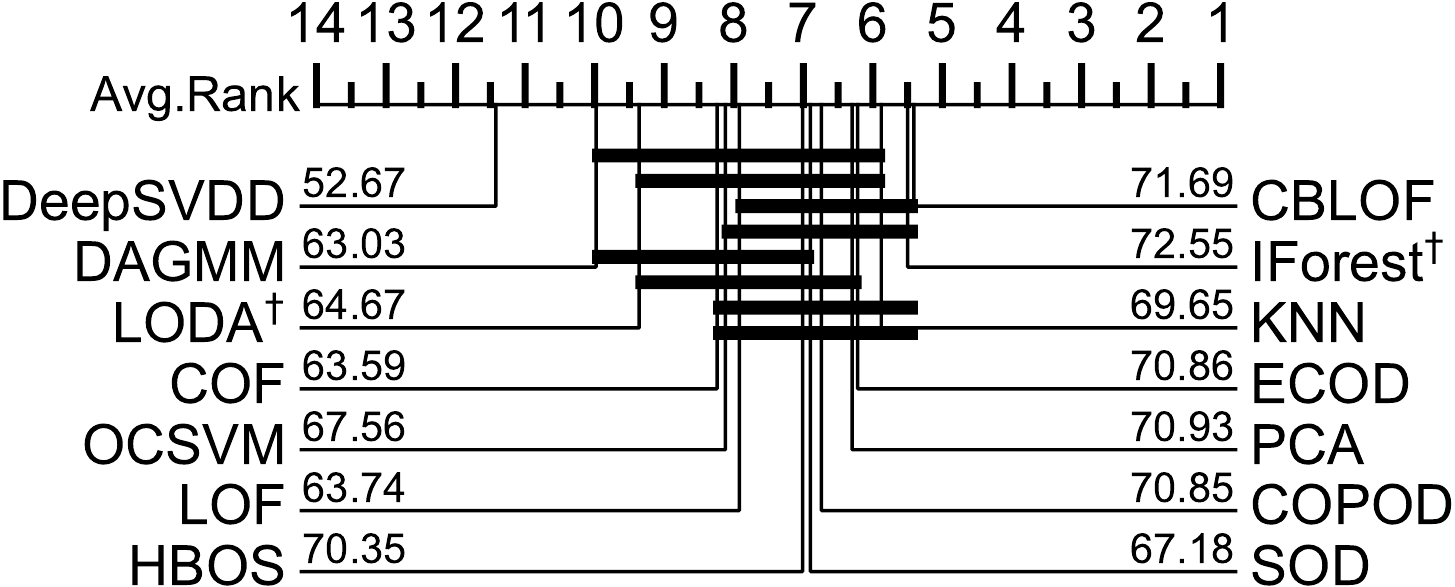}
         \caption{Avg. rank (lower the better) and avg. AUCROC (on each line) of unsupervised methods;
         groups of algorithms not statistically different are connected horizontally.}
         \label{fig:overall_unsupervised}
     \end{subfigure}
     \hspace{0.01in}
     \begin{subfigure}[t]{0.61\textwidth}
         \centering
         \hfill
         \includegraphics[width=1\textwidth]{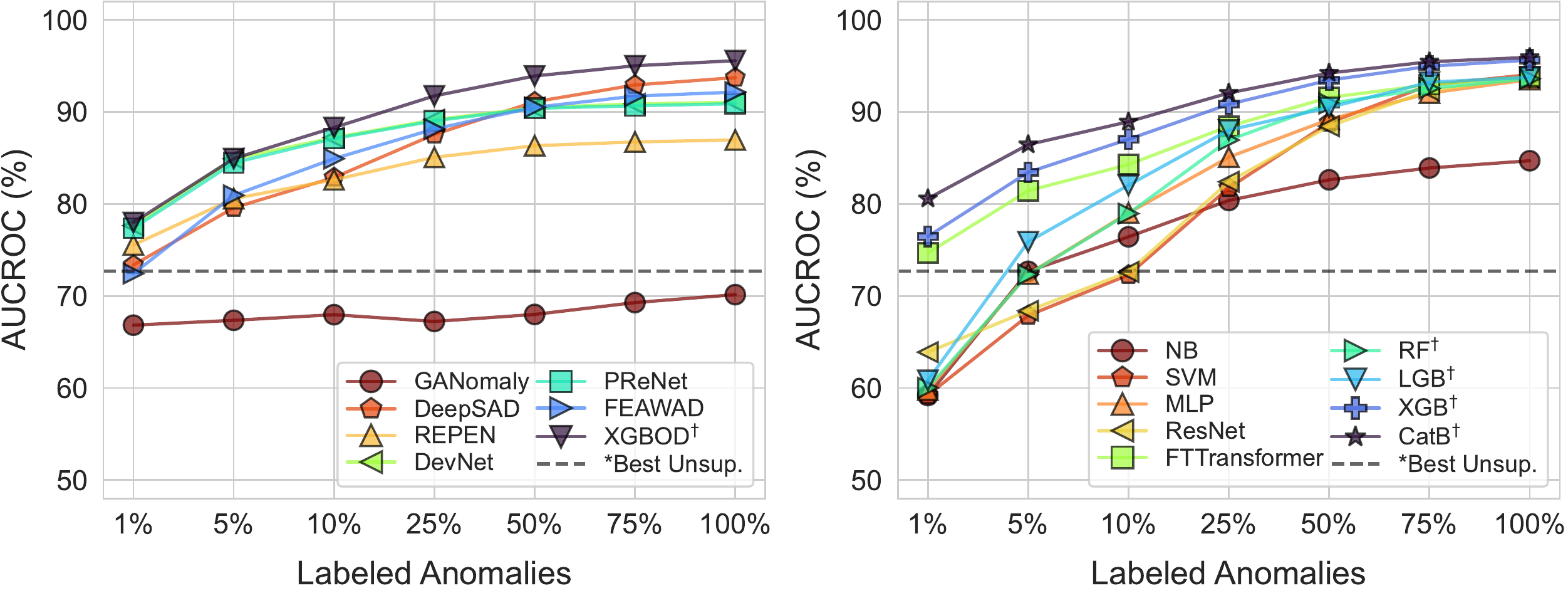}
         \caption{Avg. AUCROC (on \ndatasets datasets) vs. \% of labeled anomalies (x-axis); semi-supervised (left) and fully-supervised (right). Most label-informed algorithms outperform the best \textit{unsupervised} algorithm CBLOF (denoted as the dashed line) with 10\% labeled anomalies.}
         \label{fig:overall_supervised}
     \end{subfigure}
     \vspace{0.05in}
    
    \begin{subfigure}[t]{\dimexpr.8\linewidth-2em\relax}
    \centering
    \includegraphics[width=.95\linewidth,valign=t]{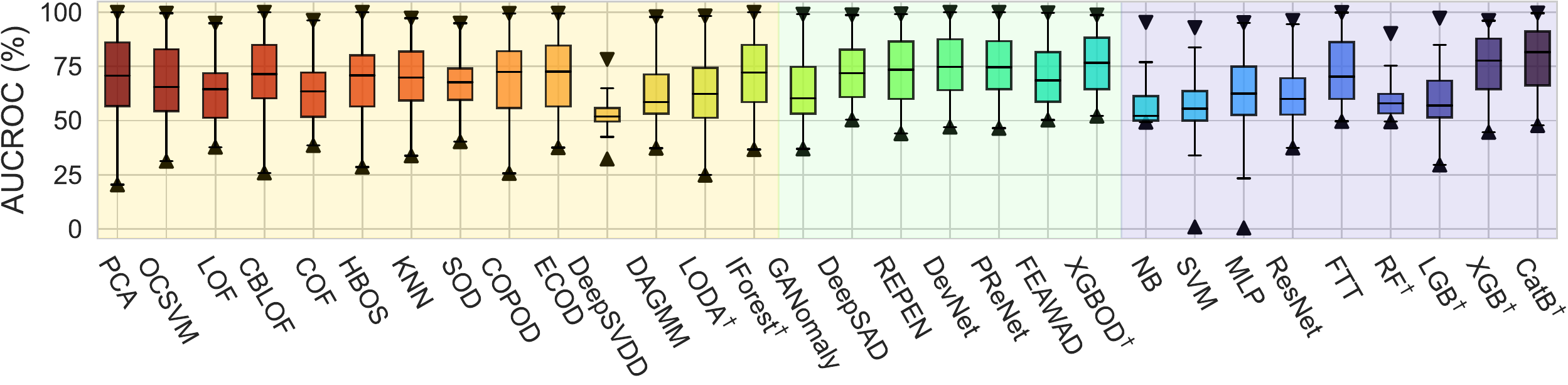}
    \end{subfigure}%
    \adjustbox{minipage=9em,valign=t}{\subcaption{Boxplot of AUCROC ($@$1\% labeled anomalies) on \ndatasets datasets; we denote un-, semi-, and fully supervised methods in light yellow, green, and purple.}}\label{sfig:auc_boxplot}%
    
    \begin{subfigure}[t]{\dimexpr.8\linewidth-2em\relax}
    \centering
    \includegraphics[width=.95\linewidth,valign=t]{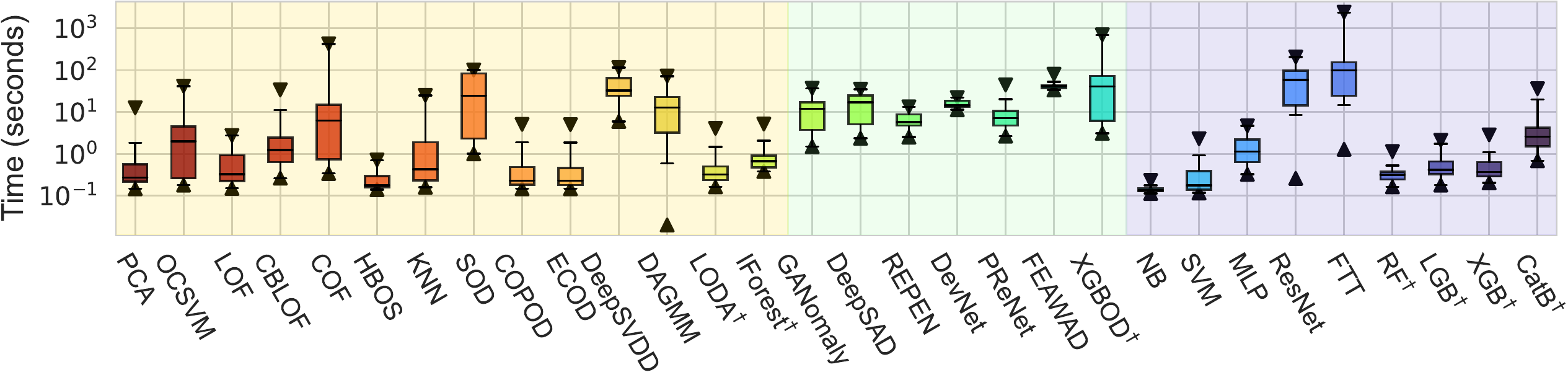}
    \end{subfigure}
    \adjustbox{minipage=9em,valign=t}{\subcaption{Boxplot of train time (see inf. time in Appx. Fig.~\ref{fig:boxplot_infertime}) on \ndatasets datasets; we denote un-, semi-, and fully supervised methods in light yellow, green, and purple.}\label{sfig:runtime}}%
    \vspace{-0.1in}
     
     \caption{Average AD model performance across \ndatasets benchmark datasets. (a) shows that no unsupervised algorithm statistically outperforms the rest. (b) 
     shows that semi-supervised methods leverage the labels more efficiently than fully-supervised methods with a small labeled anomaly ratio $\gamma_{l}$. (c) and (d) present the boxplots of AUCROC and runtime. Ensemble methods are marked with "$^{\dag}$".}
     \vspace{-0.3in}
     \label{fig:overall model performance}
\end{figure}

\subsection{Overall Model Performance on Datasets with Varying Degrees of Supervision}
\label{subsec:overall_performance}
\vspace{-0.1in}
As introduced in \S \ref{Angle I},
we first present the results of unsupervised methods on \ndatasets datasets in Fig.~\ref{fig:overall_unsupervised}, and then compare label-informed semi- and fully-supervised methods under varying degrees of supervision, i.e., different label ratios of $\gamma_{l}$ (from $1\%$ to $100\%$ full labeled anomalies) in Fig.~\ref{fig:overall_supervised}.

\textbf{None of the unsupervised methods is statistically better than the others}, as shown in the critical difference diagram of Fig.~\ref{fig:overall_unsupervised} (where most algorithms are horizontally connected without statistical significance). 
We also note that some DL-based unsupervised methods like DeepSVDD and DAGMM are surprisingly worse than shallow methods.
Without the guidance of label information, DL-based unsupervised algorithms are harder to train (due to more hyperparameters) and more difficult to tune hyperparameters, leading to unsatisfactory performance.

\textbf{Semi-supervised methods outperform supervised methods when limited label information is available}. For $\gamma_{l}\leq 5\%$, i.e., only less than $5\%$ labeled anomalies are available during training, the detection performance of semi-supervised methods (median AUCROC$=75.56\%$ for $\gamma_{l}=1\%$ and AUCROC$=80.95\%$ for $\gamma_{l}=5\%$) are generally better than that of fully-supervised algorithms (median AUCROC$=60.84\%$ for $\gamma_{l}=1\%$ and AUCROC$=72.69\%$ for $\gamma_{l}=5\%$). For most semi-supervised methods, merely $1\%$ labeled anomalies are sufficient to surpass the best unsupervised method (shown as the dashed line in Fig.~\ref{fig:overall_supervised}), while most supervised methods need $10\%$ labeled anomalies to achieve so. We also show the improvement of algorithm performances about the increasing $\gamma_{l}$, and notice that with a large number of labeled anomalies, both semi-supervised and supervised methods have comparable performance. Putting these together, we verify the assumed advantage of semi-supervised methods in leveraging limited label information more efficiently.

\textbf{Latest network architectures like Transformer and emerging ensemble methods yield competitive performance in AD}. 
Fig.~\ref{fig:overall_supervised} shows FTTransformer and ensemble methods like XGB(oost) and CatB(oost) provide satisfying detection performance among all the label-informed algorithms, even these methods are not specifically proposed for the anomaly detection tasks. For $\gamma_{l}=1\%$, the AUCROC of FTTransformer and the median AUCROC of ensemble methods are $74.68\%$ and $76.47\%$, respectively, outperforming the median AUCROC of all label-informed methods $72.91\%$. The great performance of tree-based ensembles (in tabular AD) is consistent with the findings in literature \cite{borisov2021deep,grinsztajn2022tree,vargaftik2021rade}, which may be credited to their capacity to handle imbalanced AD datasets via aggregation. Future research may focus on understanding the cause and other merits of ensemble trees in tabular AD, e.g., better model efficiency. 

\textbf{Runtime Analysis.} We present the train and inference time in Fig. \ref{sfig:runtime} and Appx. Fig.~\ref{fig:boxplot_infertime}. Runtime analysis finds that HBOS, COPOD, ECOD, and NB are the fastest as they treat each feature independently. In contrast, more complex representation learning methods like XGBOD, ResNet, and FTTansformer are computationally heavy. This should be factored in for algorithm selection.

\textit{\textbf{Future Direction 1: Unsupervised Algorithm Evaluation, Selection, and Design}}. For unsupervised AD, the results suggest that future algorithms should be evaluated on large testbeds like \system for statistical tests (such as via critical difference diagrams).
Meanwhile, the no-free-lunch theorem \cite{wolpert1997no} suggests there is no universal winner for all tasks, and more focus should be spent on understanding the suitability of each AD algorithm. Notably, algorithm selection and hyperparameter optimization
are important in unsupervised AD, but limited works \cite{bahri2022automl,ma2021large,zhao2022towards,MetaOD} have studied them. We may consider self-supervision \cite{qiu2021neural,sehwag2020ssd,shenkar2021anomaly,xiao2021we} and transfer learning \cite{deecke2021transfer} to improve tabular AD as well.
Thus, we call for attention to large-scale evaluation,  task-driven algorithm selection, and 
data augmentation/transfer for unsupervised AD.

\textit{\textbf{Future Direction 2: Semi-supervised Learning}}. By observing the success of using limited labels in AD, we would call for more attention to semi-supervised AD methods which can leverage both the guidance from labels efficiently and the exploration of the unlabeled data. Regarding backbones, the latest network architectures like Transformer and ensembling show their superiority in AD tasks.

\vspace{-0.1in}
\subsection{Algorithm Performance under Different Types of Anomalies}
\label{exp:types}
\begin{figure}[h!]
     \centering
     \begin{subfigure}[b]{0.24\textwidth}
         \centering
         \includegraphics[width=1.08\textwidth]{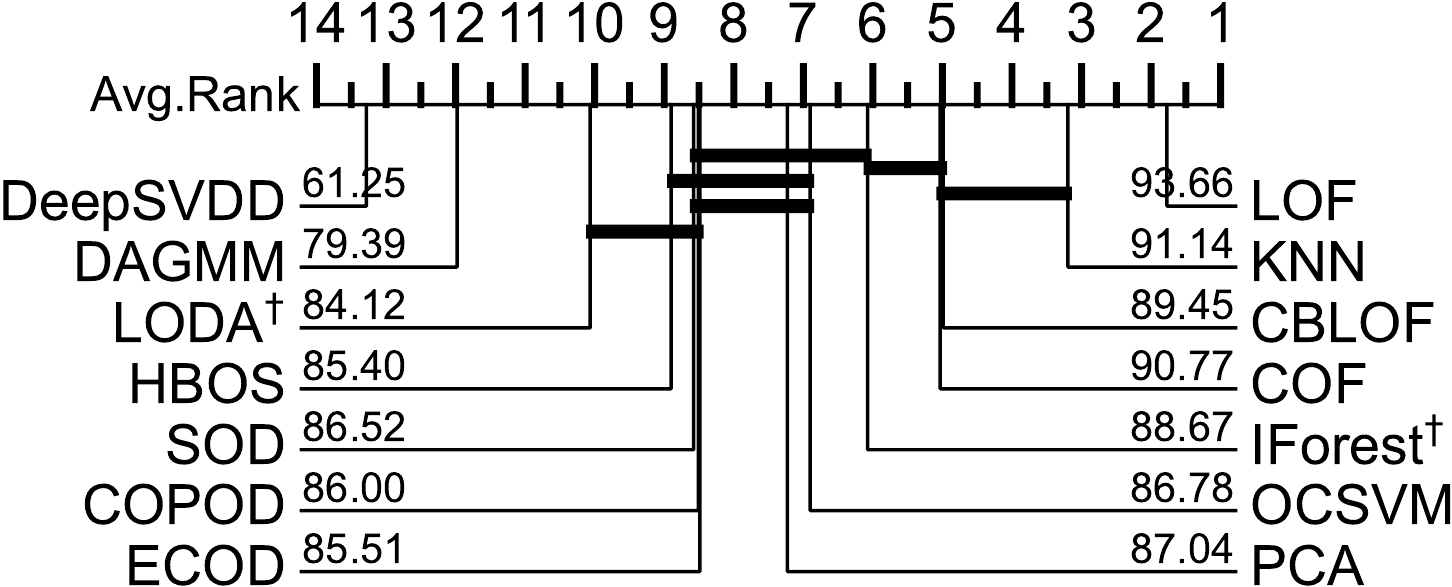}
         \caption{Local anomalies}
         \label{fig:local}
     \end{subfigure}
     \hfill
     \begin{subfigure}[b]{0.24\textwidth}
         \centering
         \includegraphics[width=1.08\textwidth]{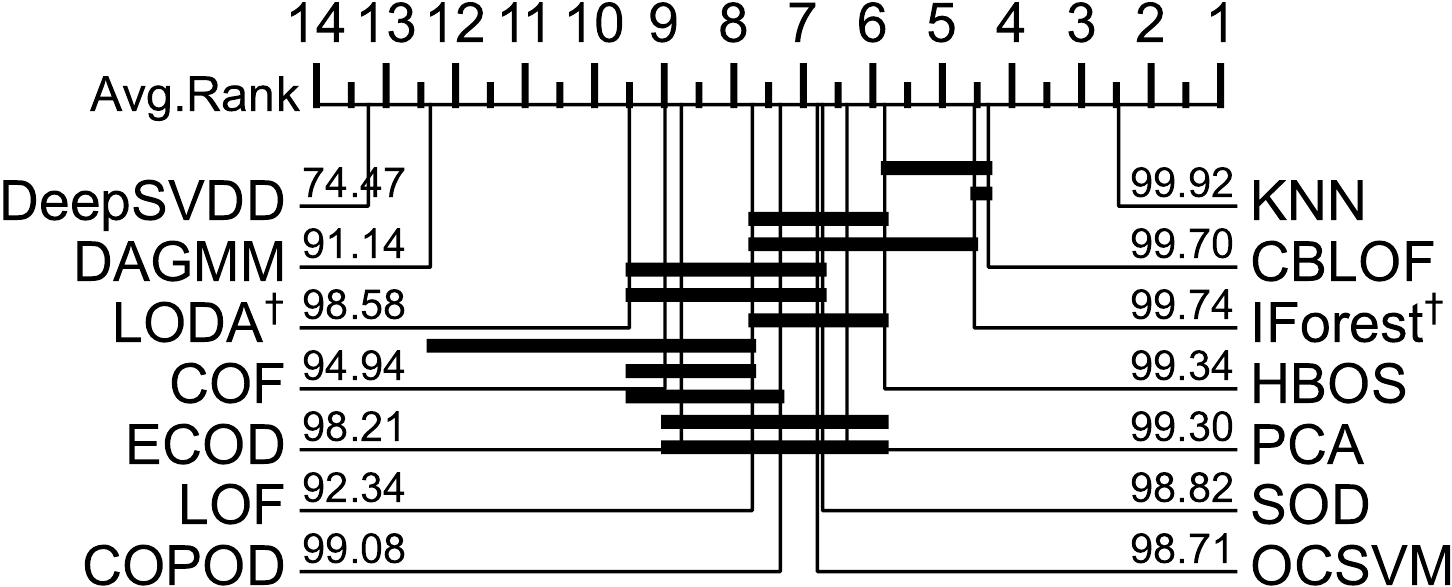}
         \caption{Global anomalies}
         \label{fig:global}
     \end{subfigure}
     \hfill
     \begin{subfigure}[b]{0.235\textwidth}
         \centering
         \includegraphics[width=1.08\textwidth]{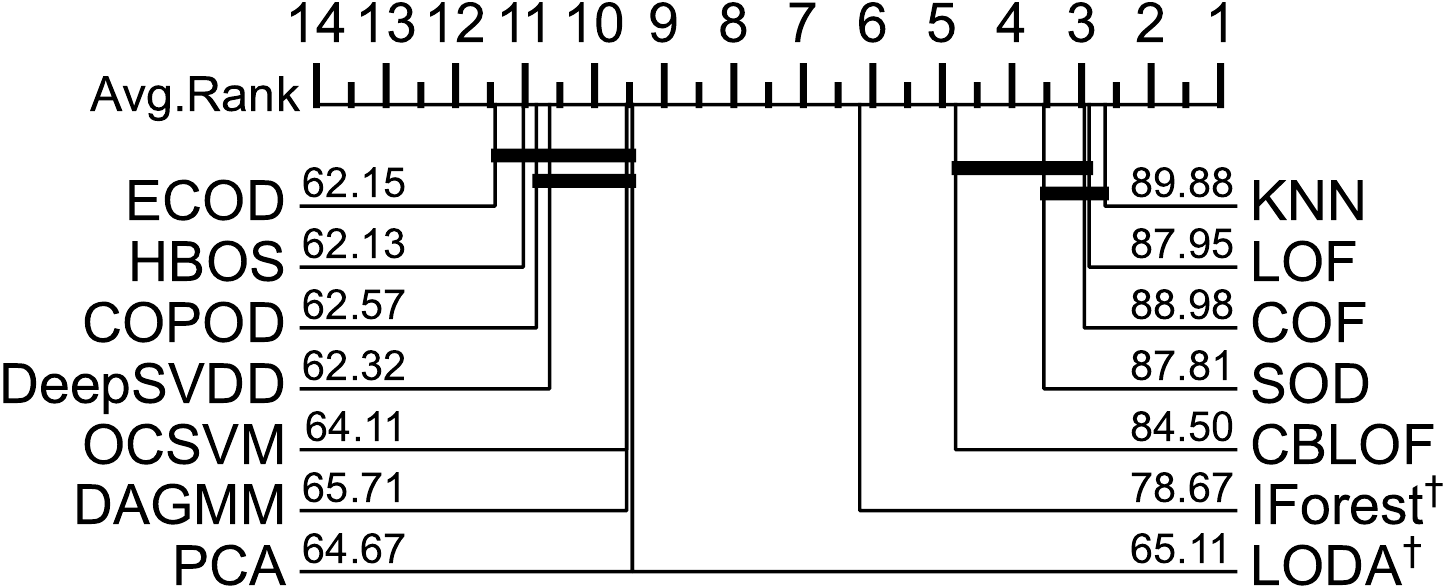}
         \caption{Dependency anomalies}
         \label{fig:dependency}
     \end{subfigure}
     \hfill
     \begin{subfigure}[b]{0.24\textwidth}
         \centering
         \includegraphics[width=1.08\textwidth]{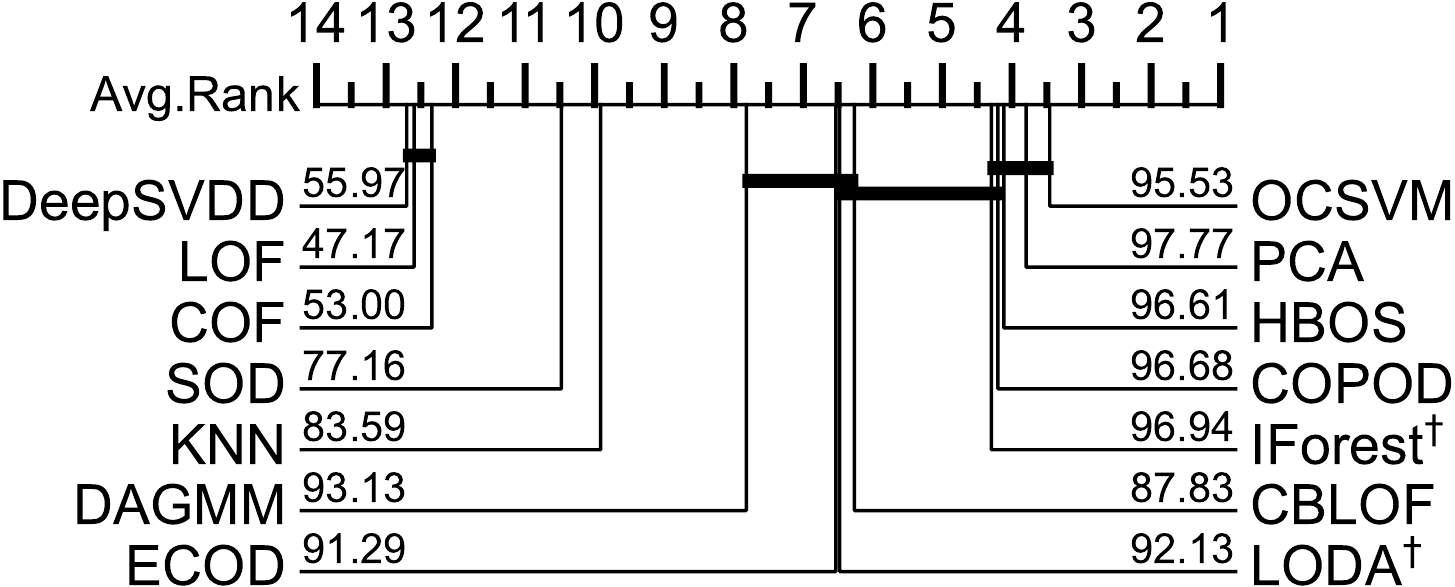}
         \caption{Clustered anomalies}
         \label{fig:clustered}
     \end{subfigure}
     \vspace{-0.05in}
     \caption{Avg. rank (lower the better) of unsupervised methods on different types of anomalies. Groups of algorithms not significantly different are connected horizontally in the CD diagrams. The unsupervised methods perform well when their assumptions conform to the underlying anomaly type.
     }
     \label{fig:type_unsupervised}
     \vspace{-0.1in}
\end{figure}

\begin{figure}[!ht]
     \centering
     \begin{subfigure}[b]{0.495\textwidth}
         \centering
         \includegraphics[width=\textwidth]{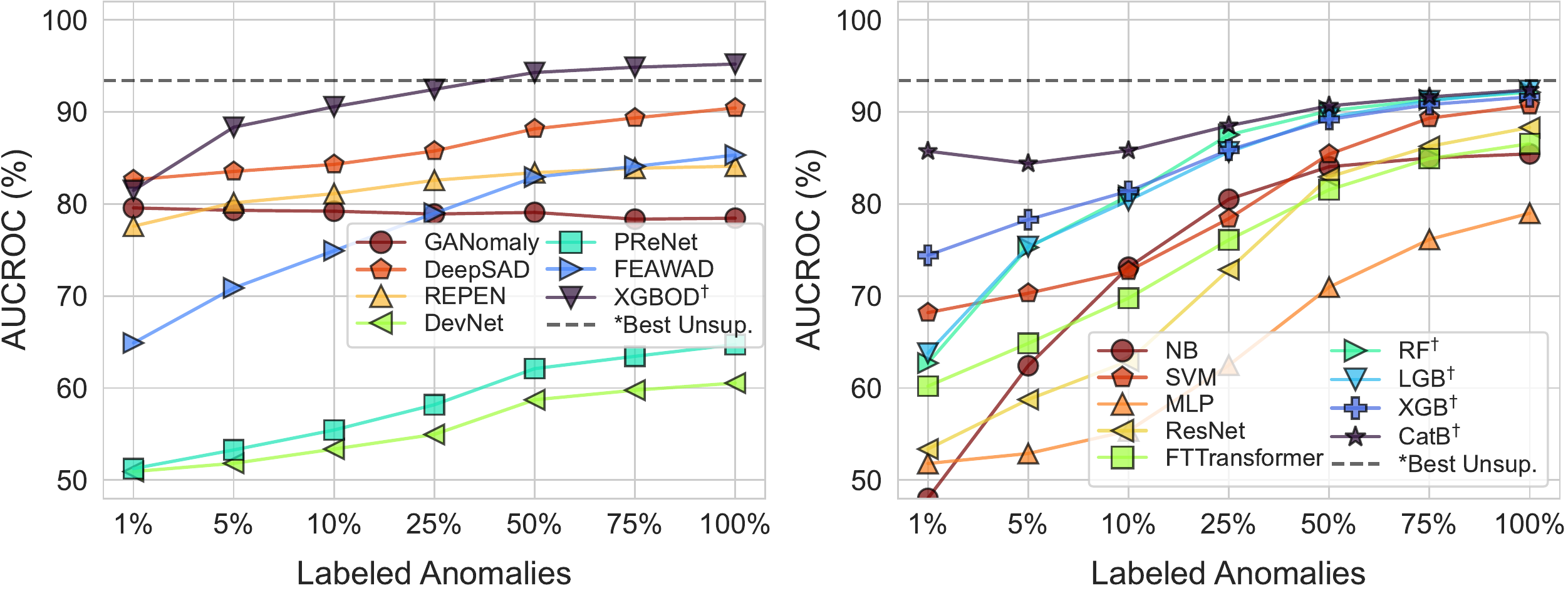}
         \caption{Local anomalies}
         \label{fig:local_label}
     \end{subfigure}
     \hfill
     \begin{subfigure}[b]{0.495\textwidth}
         \centering
         \includegraphics[width=\textwidth]{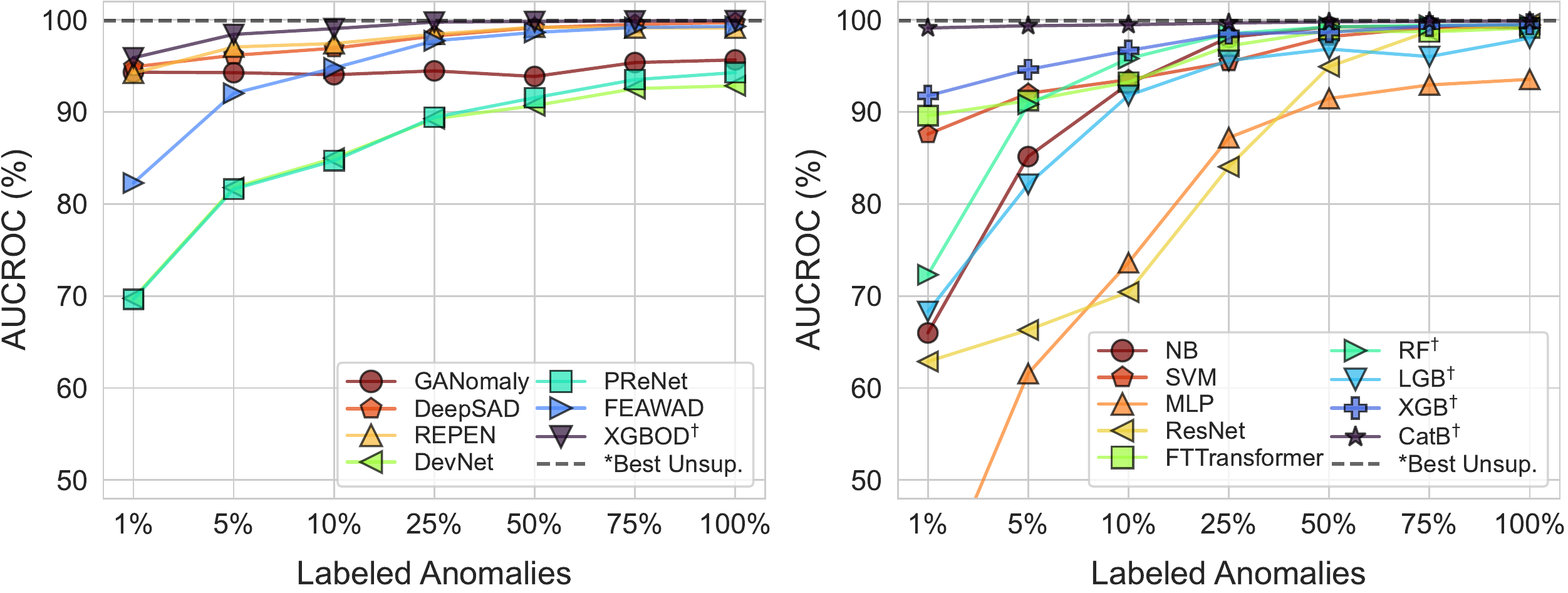}
         \caption{Global anomalies}
         \label{fig:global_label}
     \end{subfigure}
     \hfill
     \begin{subfigure}[b]{0.495\textwidth}
         \centering
         \includegraphics[width=\textwidth]{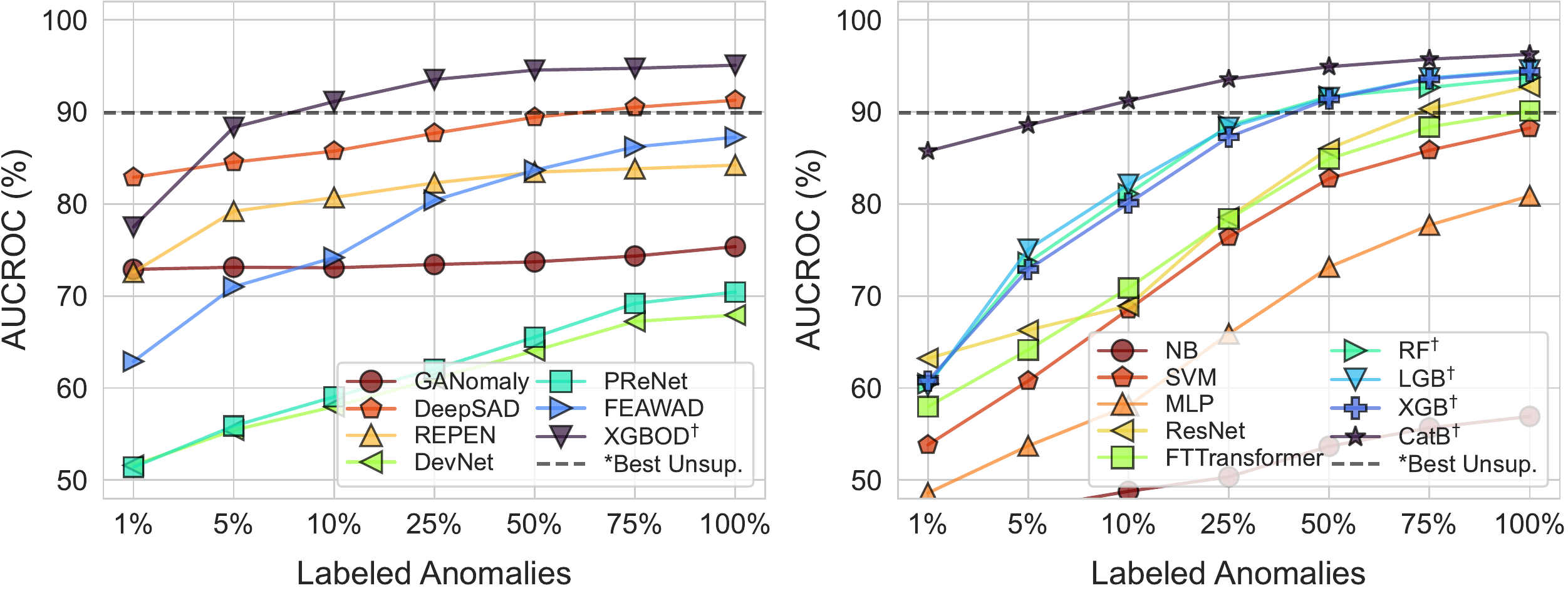}
         \caption{Dependency anomalies}
         \label{fig:dependency_label}
     \end{subfigure}
     \hfill
     \begin{subfigure}[b]{0.495\textwidth}
         \centering
         \includegraphics[width=\textwidth]{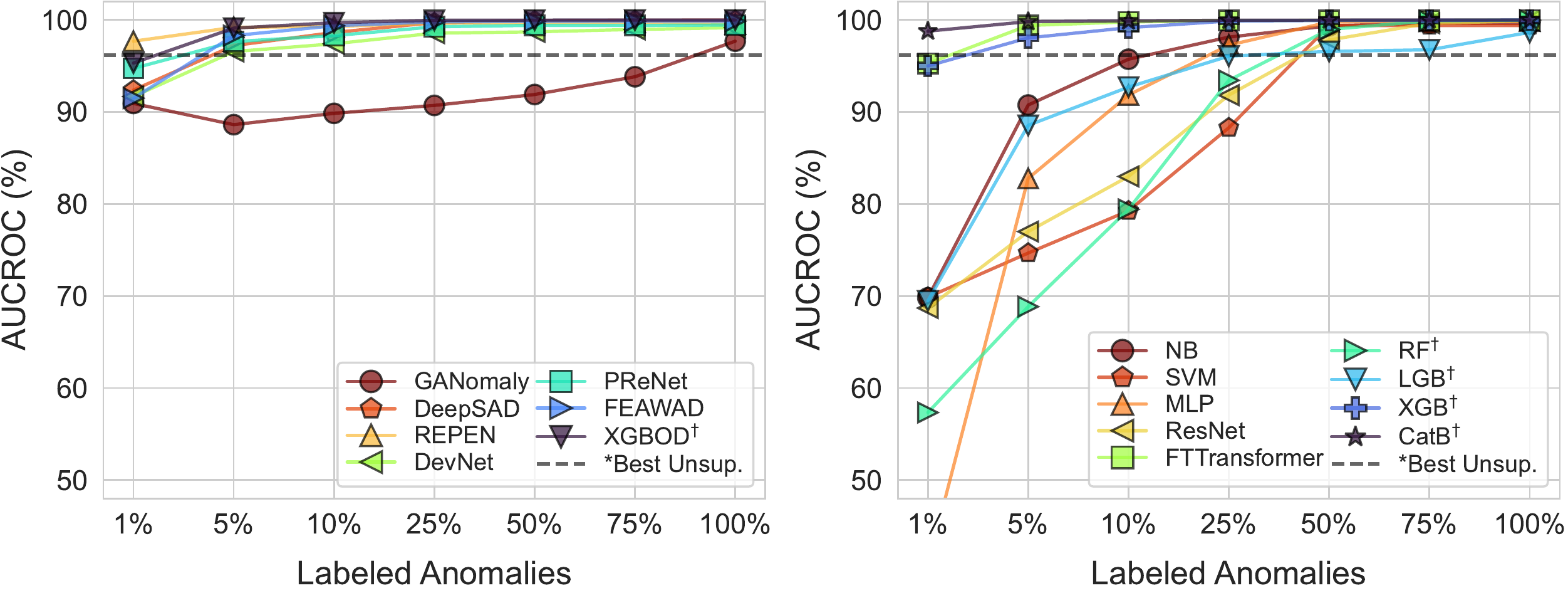}
         \caption{Clustered anomalies}
         \label{fig:clustered_label}
     \end{subfigure}
     \caption{Semi- (left of each subfigure) and supervised (right) algorithms' performance on different types of anomalies with varying levels of labeled anomalies. Surprisingly, these label-informed algorithms are \textit{inferior} to the best unsupervised method except for the clustered anomalies.
     }
     \vspace{-0.15in}
     \label{fig:type_supervised}
\end{figure}

\vspace{-0.1in}
Under four types of anomalies introduced in \S \ref{subsub:types}), we show the performances of unsupervised methods 
in Fig.~\ref{fig:type_unsupervised}, and then compare both semi- and fully-supervised methods in Fig.~\ref{fig:type_supervised}.

\vspace{-0.05in}

\textbf{Performance of unsupervised algorithms highly depends on the alignment of its assumptions and the underlying anomaly type}. 
As expected, \textit{local} anomaly factor (LOF) is statistically better than other unsupervised methods for the local anomalies (Fig.~\ref{fig:local}), and KNN, which uses $k$-th (\textit{global}) nearest neighbor's distance as anomaly scores, is the statistically best detector for global anomalies (Fig.~\ref{fig:global}). 
Again, there is no algorithm performing well on all types of anomalies; LOF achieves the best AUCROC on local anomalies (Fig.~\ref{fig:local}) and the second best AUCROC rank on dependency anomalies (Fig.~\ref{fig:dependency}), but performs poorly on clustered anomalies (Fig.~\ref{fig:clustered}). Practitioners should select algorithms based on the characteristics of the underlying task, and consider the algorithm which may cover more high-interest anomaly types \cite{lee2021gen}.

\textbf{The ``power'' of prior knowledge on anomaly types may overweigh the usage of partial labels}.
For the local, global, and dependency anomalies, most label-informed methods perform worse than the best unsupervised methods of each type (corresponding to LOF, KNN, and KNN). For example, the detection performance of XGBOD for the local anomalies is inferior to the best unsupervised method LOF when $\gamma_{l} \leq 50\%$, while other methods perform worse than LOF in all cases (See Fig.~\ref{fig:local_label}). 
Why could not label-informed algorithms beat unsupervised methods in this setting? 
We believe that partially labeled anomalies cannot well capture all characteristics of specific types of anomalies, and learning such decision boundaries is challenging. For instance, different local anomalies often exhibit various behaviors, as shown in Fig.~\ref{fig:demo_local}, which may be easier to identify by a generic definition of ``locality" in unsupervised methods other than specific labels. 
Thus, incomplete label information may bias the learning process of these label-informed methods, which explains their relatively inferior performances compared to the best unsupervised methods.
This conclusion is further verified by the results of clustered anomalies (See Fig.~\ref{fig:clustered_label}), where label-informed (especially semi-supervised) methods outperform the best unsupervised method OCSVM, as few labeled anomalies can already represent similar behaviors in the clustered anomalies (Fig.~\ref{fig:demo_clusdtered}).

\textit{\textbf{Future Direction 3: Leveraging Anomaly Types as Valuable Prior Knowledge}}. The above results emphasize the importance of knowing anomaly types in achieving high detection performance even without labels, and call for attention to designing anomaly-type-aware detection algorithms. In an ideal world, one may combine multiple AD algorithms based on the composition of anomaly types, via frameworks like dynamic model selection and combination \cite{zhao2019lscp}.
To our knowledge, the latest advancement in this end \cite{jerez2022equivalence} provides an equivalence criterion for measuring to what degree two anomaly detection algorithms detect the same kind of anomalies. Furthermore, future research may also consider designing semi-supervised AD methods capable of detecting different types of unknown anomalies while effectively improving performance by the partially available labeled data. Another interesting direction is to train an offline AD model using synthetically generated anomalies and then adapt it for online prediction on real-world datasets with likely similar anomaly types. Unsupervised domain adaption and transfer learning for AD \cite{deecke2021transfer,yang2020anomaly} may serve as useful references.


\subsection{Algorithm Robustness under Noisy and Corrupted Data}
\label{exp:robustness}
\begin{figure}[h!]
     \centering
     \begin{subfigure}[b]{0.245\textwidth}
         \centering
         \includegraphics[width=\textwidth]{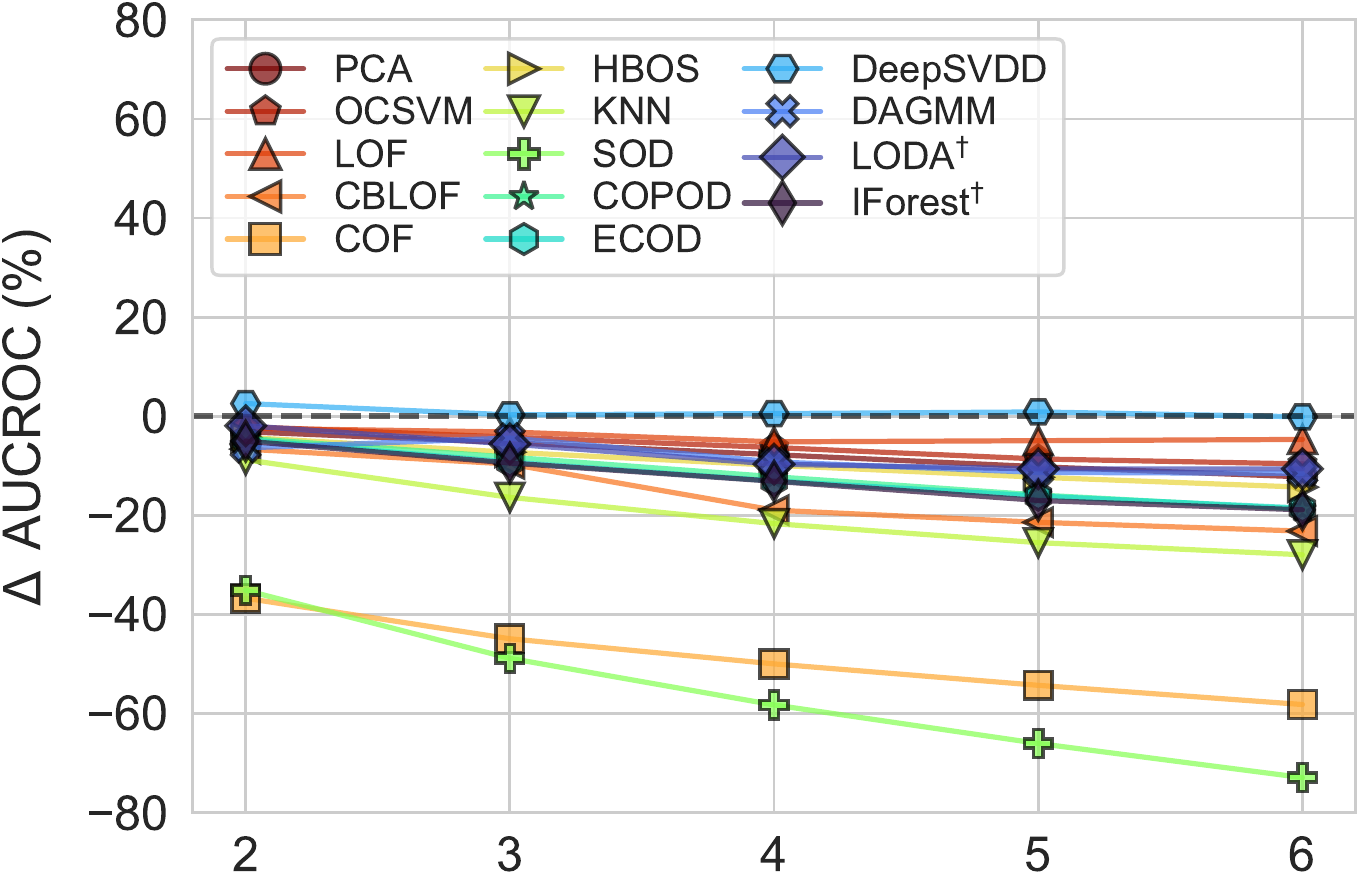}
         \caption{\scriptsize Duplicated Anomalies, unsup}
         \label{fig:dp_unsup}
     \end{subfigure}
     \begin{subfigure}[b]{0.245\textwidth}
         \centering
         \includegraphics[width=\textwidth]{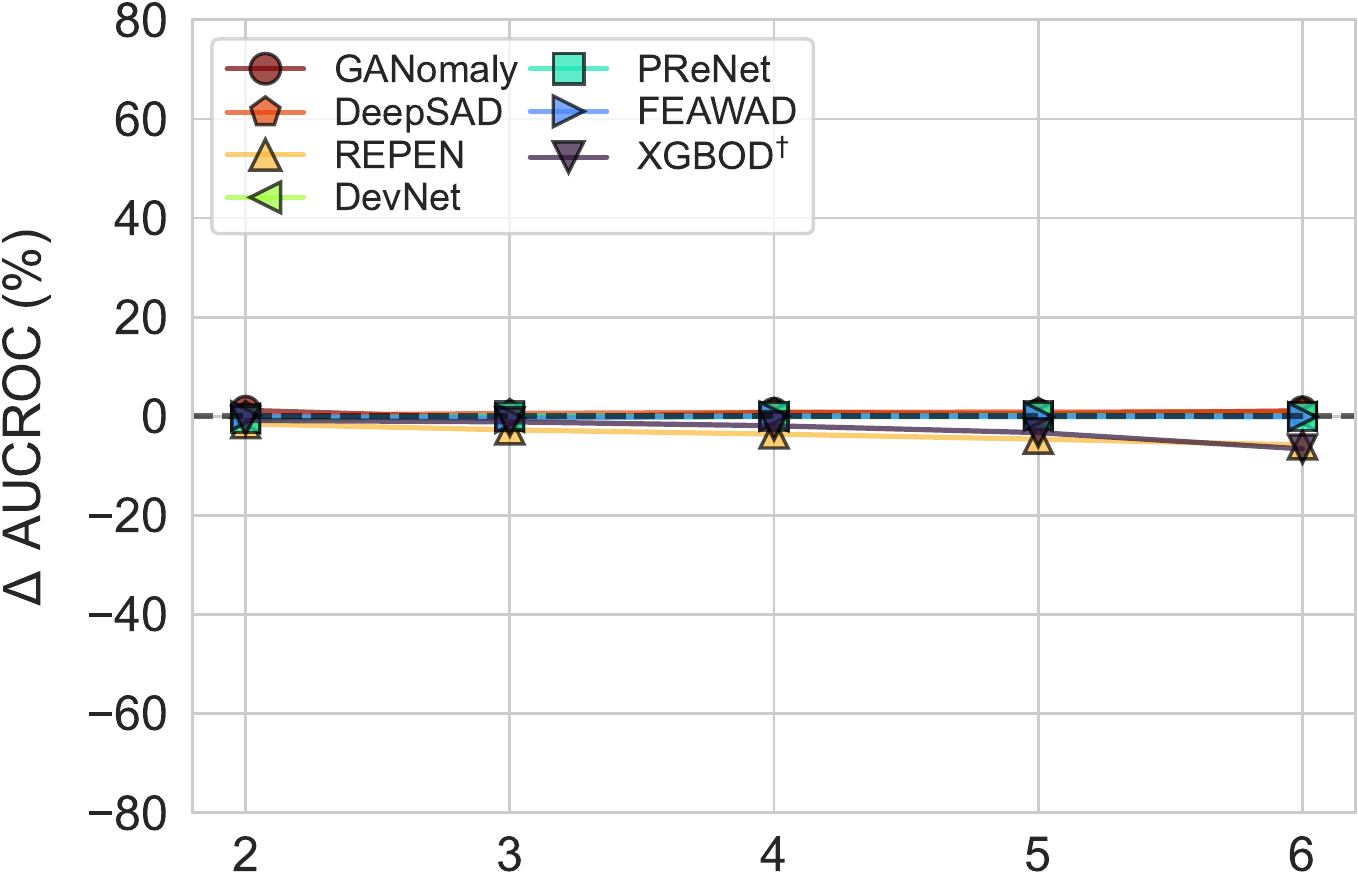}
         \caption{\scriptsize Duplicated Anomalies, semi}
         \label{fig:dp_semi}
     \end{subfigure}
     \begin{subfigure}[b]{0.245\textwidth}
         \centering
         \includegraphics[width=\textwidth]{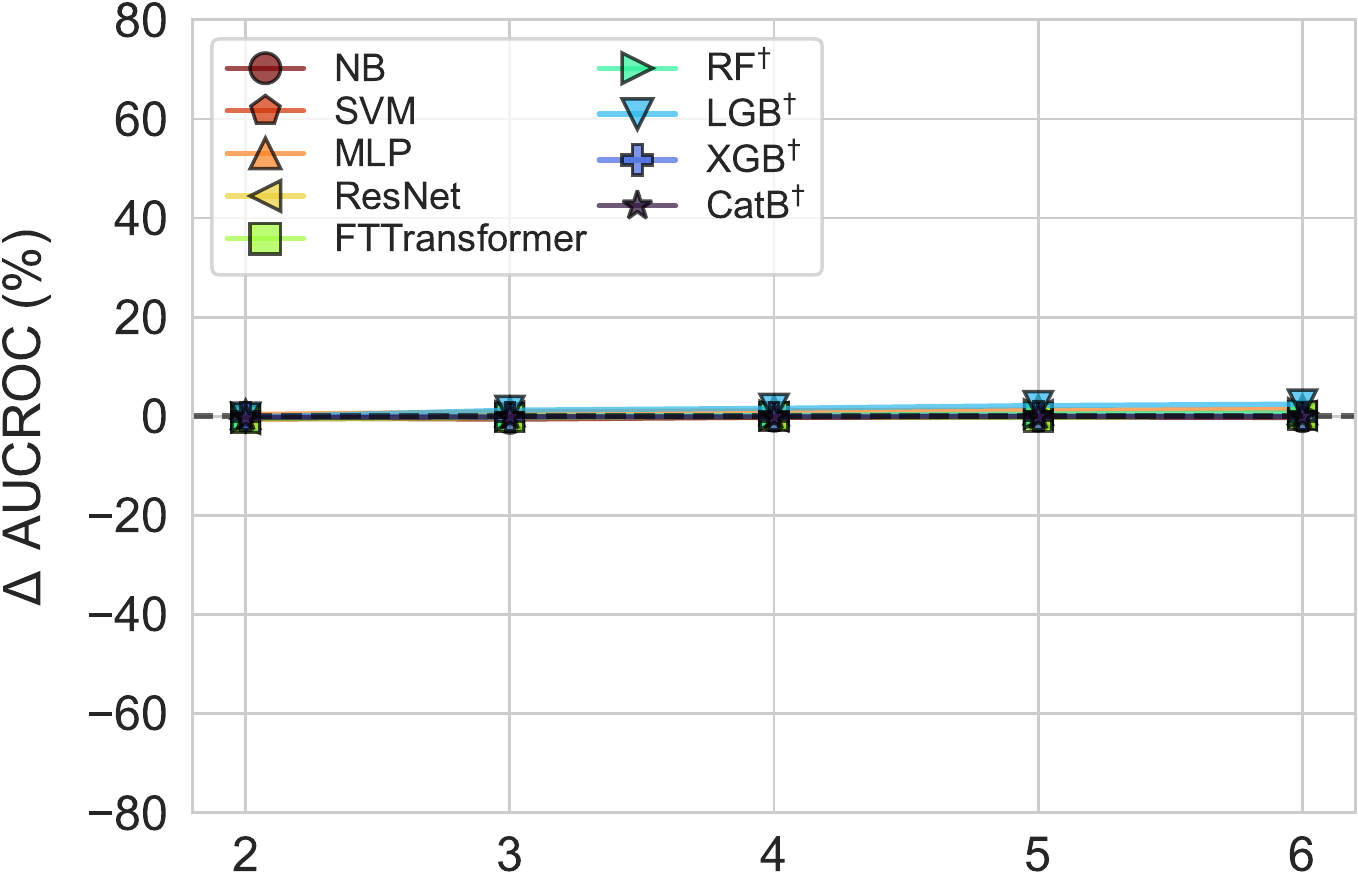}
         \caption{\scriptsize Duplicated Anomalies, sup}
         \label{fig:dp_sup}
     \end{subfigure}
     \begin{subfigure}[b]{0.245\textwidth}
         \centering
         \includegraphics[width=\textwidth]{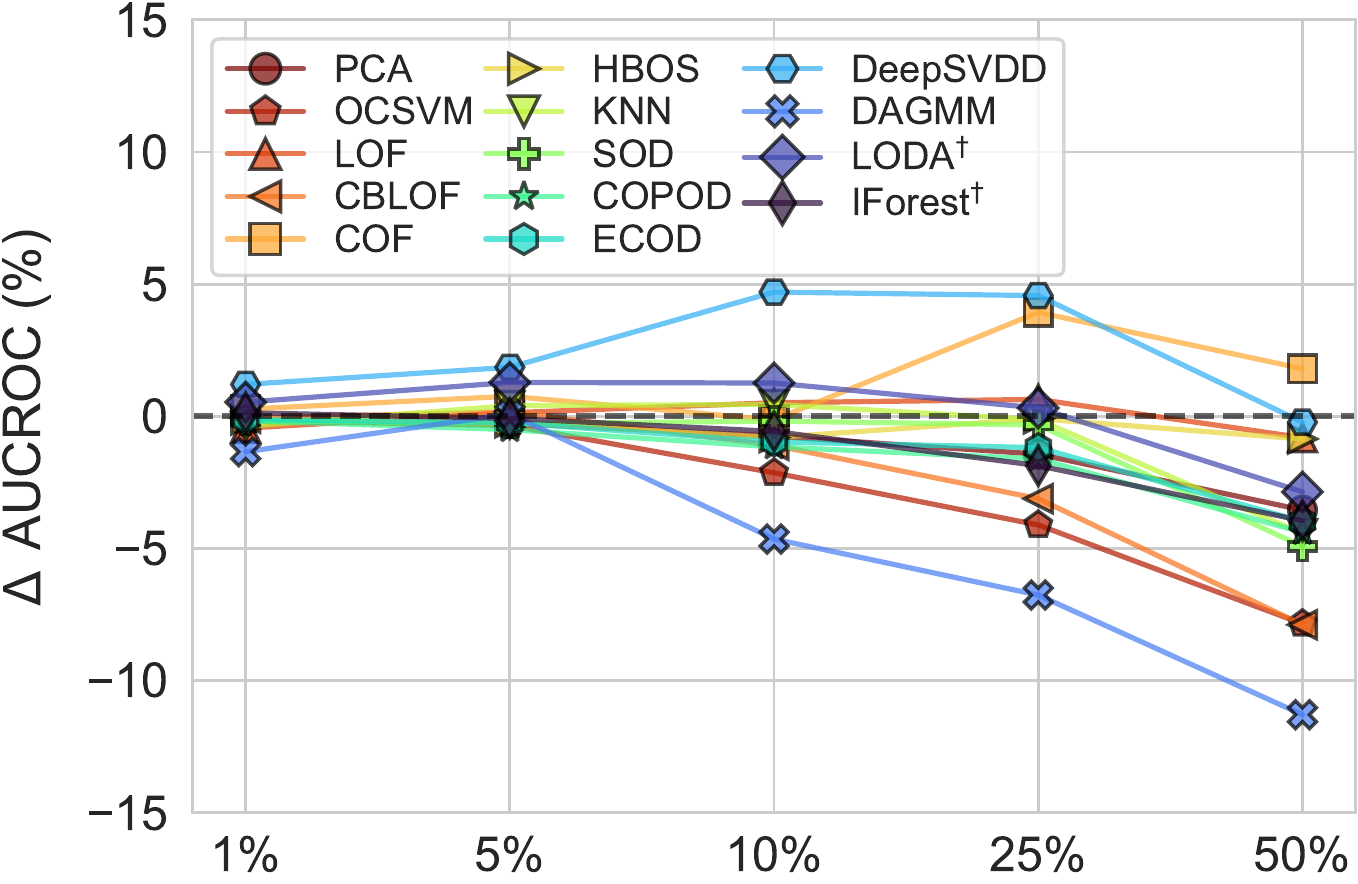}
         \caption{\scriptsize Irrelevant Features, unsup}
         \label{fig:ir_unsup}
     \end{subfigure}
     \begin{subfigure}[b]{0.245\textwidth}
         \centering
         \includegraphics[width=\textwidth]{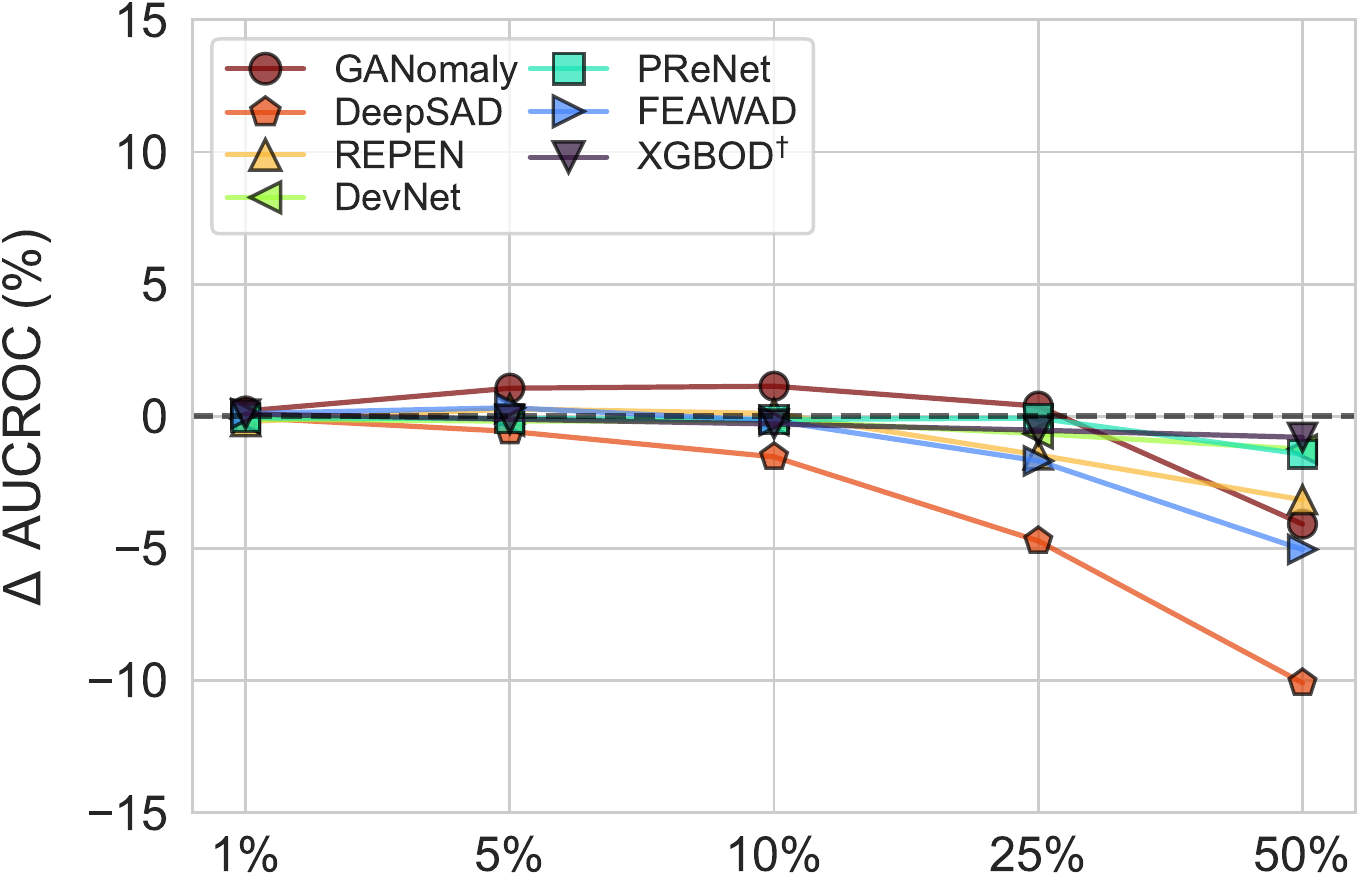}
         \caption{\scriptsize Irrelevant Features, semi}
         \label{fig:ir_semi}
     \end{subfigure}
     \begin{subfigure}[b]{0.245\textwidth}
         \centering
         \includegraphics[width=\textwidth]{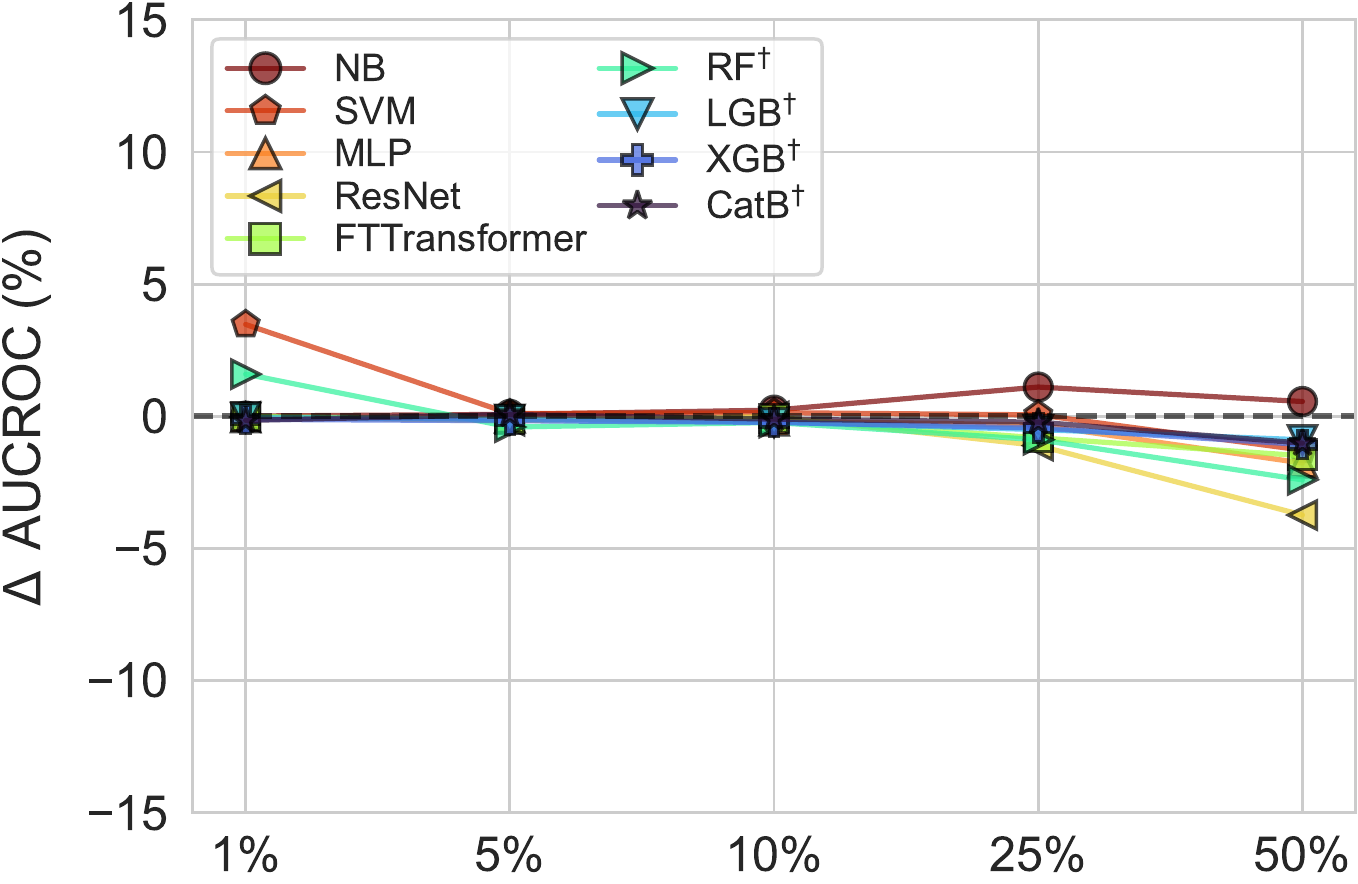}
         \caption{\scriptsize Irrelevant Features, sup}
         \label{fig:ir_sup}
     \end{subfigure}
     \begin{subfigure}[b]{0.245\textwidth}
         \centering
         \includegraphics[width=\textwidth]{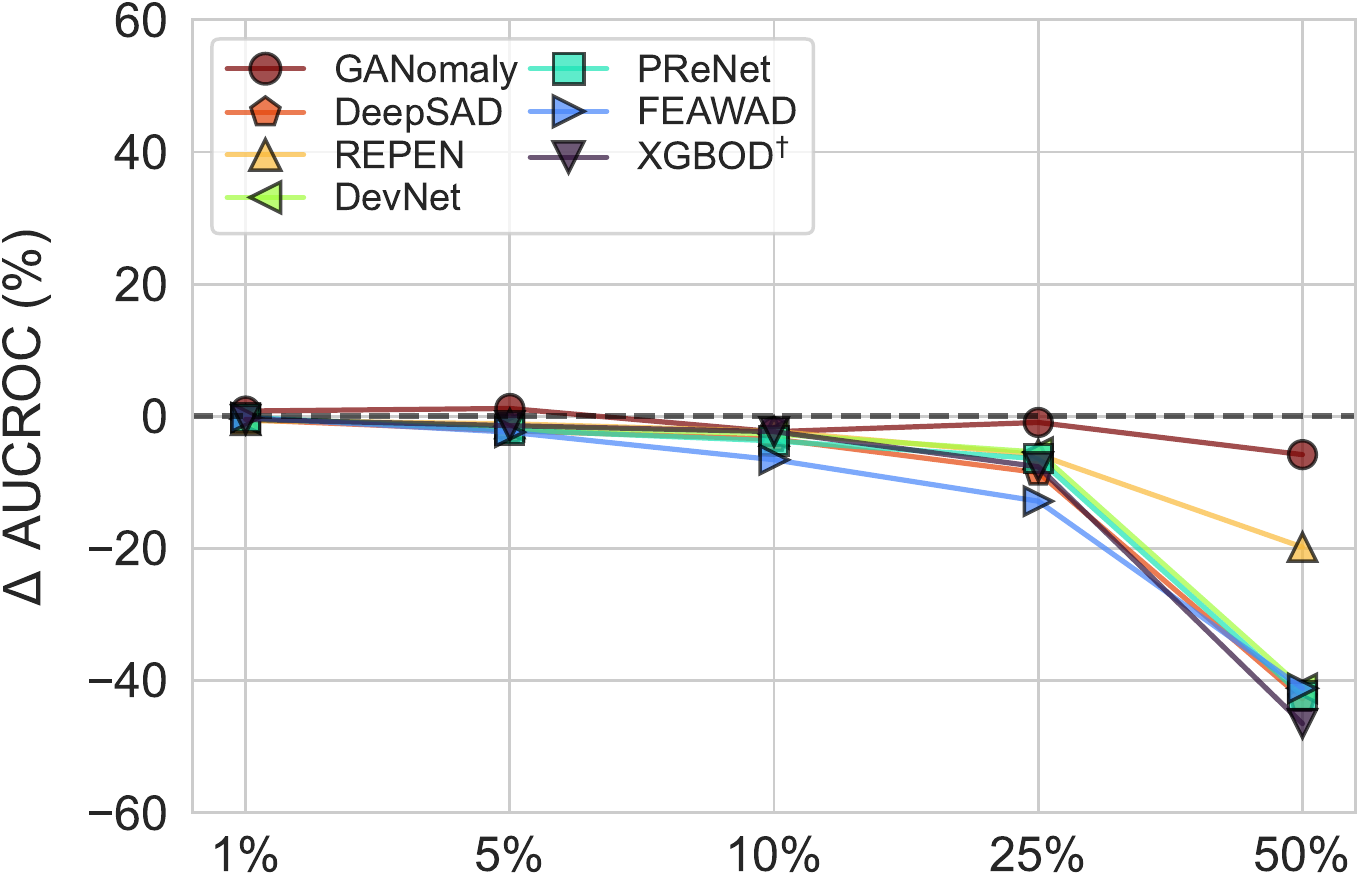}
         \caption{\scriptsize Annotation Errors, semi}
         \label{fig:ae_semi}
     \end{subfigure}
     \begin{subfigure}[b]{0.245\textwidth}
         \centering
         \includegraphics[width=\textwidth]{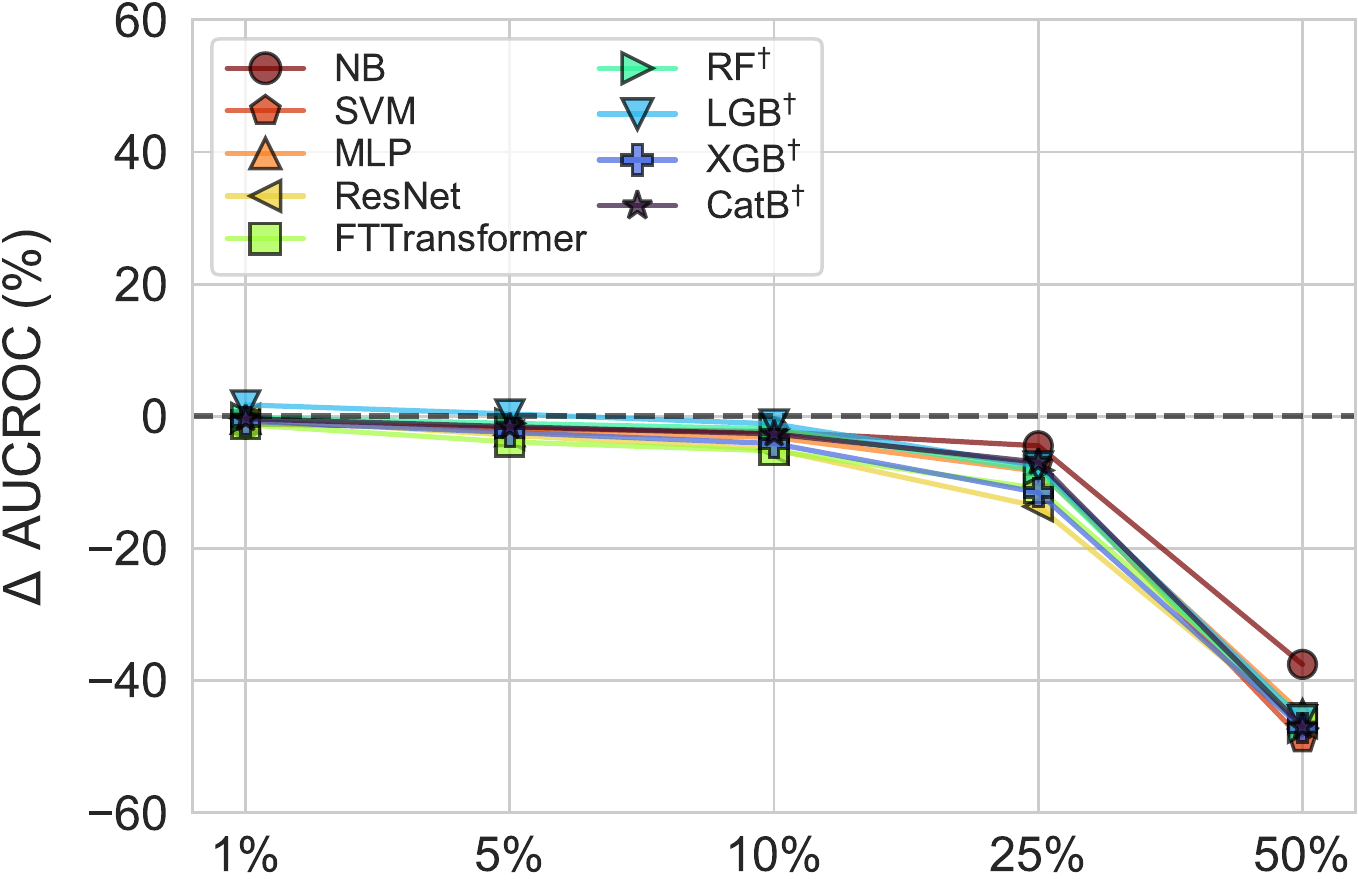}
         \caption{\scriptsize Annotation Errors, sup}
         \label{fig:ae_sup}
     \end{subfigure}
     \vspace{-0.15in}
    \caption{Algorithm performance change under noisy and corrupted data (i.e., duplicated anomalies for (a)-(c), irrelevant features for (d)-(f), and annotation errors for (g) and (h)). X-axis denotes either the duplicated times or the noise ratio. Y-axis denotes the \% of performance change ($\Delta \text{AUCROC}$), and its range remains consistent across different algorithms. The results reveal unsupervised methods' susceptibility to duplicated anomalies and the usage of label information in defending irrelevant features. Un-, semi-, and fully-supervised methods are denoted as \textit{unsup}, \textit{semi}, and \textit{sup}, respectively.
    }
    \vspace{-0.1in}
    \label{fig:model_robustness}
\end{figure}

\vspace{-0.1in}
In this section, we investigate the algorithm robustness (i.e., $\Delta \text{performance}$; see absolute performance plot in Appx. \ref{fig:robustness:absolute}) of different AD algorithms under noisy and data corruption described in \S \ref{Angle III}. The default $\gamma_{l}$ is set to $100\%$ since we only care about the relative change of model performance. Fig.~\ref{fig:model_robustness} demonstrates the results.

\textbf{Unsupervised methods are more susceptible to duplicated anomalies}. As shown in Fig.~\ref{fig:dp_unsup}, almost all unsupervised methods are severely impacted by duplicated anomalies. Their AUCROC deteriorates proportionally with the increase in duplication. When anomalies are duplicated by $6$ times, the median $\Delta \text{AUCROC}$ of unsupervised methods is $-16.43\%$, compared to that of semi-supervised methods $-0.05\%$ (Fig.~\ref{fig:dp_semi}) and supervised methods $0.13\%$ (Fig.~\ref{fig:dp_sup}).
One explanation is that unsupervised methods often assume the underlying data is imbalanced with only a smaller percentage of anomalies---they rely on this assumption to detect anomalies. 
With more duplicated anomalies, the underlying data becomes more balanced, and the minority assumption of anomalies is violated, causing the degradation of unsupervised methods. Differently, more balanced datasets do not affect the performance of semi- and fully-supervised methods remarkably, with the help of labels.

\textbf{Irrelevant features cause little impact on supervised methods due to feature selection}. Compared to the unsupervised and most semi-supervised methods, the training process of supervised methods is fully guided by the data labels ($\mathbf{y}$), therefore performing robustly to the irrelevant features (i.e., corrupted $\mathbf{X}$) due to the direct (or indirect) feature selection process. For instance, ensemble trees like XGBoost can filter irrelevant features. As shown in Fig.~\ref{fig:ir_sup}, even the worst performing supervised algorithm (say ResNet) in this setting yields $\leq 5\%$ degradation when 50\% of the input features are corrupted by the uniform noises, while the un- and semi-supervised methods could face up to $10\%$ degradation.
Besides, the robust performances of supervised methods (and some semi-supervised methods like DevNet) indicate that the label information can be beneficial for feature selection. Also, Fig.~\ref{fig:ir_sup} shows that minor irrelevant features (e.g., 1\%) help supervised methods as regularization to generalize better.
\vspace{-0.05in}

\textbf{Both semi- and fully-supervised methods show great resilience to minor annotation errors}. Although the detection performance of these methods is significantly downgraded when the annotation errors are severe (as shown in Fig.~\ref{fig:ae_semi} and \ref{fig:ae_sup}), their degradation with regard to minor annotation errors is acceptable. The median $\Delta \text{AUCROC}$ of semi- and fully-supervised methods for $5\%$ annotation errors is $-1.52\%$ and $-1.91\%$, respectively. That being said, label-informed methods are still acceptable in practice as the annotation error should be relatively small \cite{li2018weakly, xu2021dp}.
\vspace{-0.05in}

\textit{\textbf{Future Direction 4: Noise-resilient AD Algorithms}}. 
Our results indicate there is an improvement space for robust unsupervised AD algorithms.
One immediate remedy is to incorporate unsupervised feature selection 
\cite{cheng2020outlier,pang2016unsupervised, pang2017learning} to combat irrelevant features. 
Moreover, label information could serve as effective guidance for model training against data noise, and it helps semi- and fully-supervised methods to be more robust. Given the difficulty of acquiring full labels, we suggest using semi-supervised methods as the backbone for designing more robust AD algorithms.

\vspace{-0.05in}

%% file: 5discussion.tex
In this paper, we introduce \system, the most comprehensive tabular anomaly detection benchmark with \nmodels algorithms and \ndatasets benchmark datasets. Based on the analyses on multiple comparison angles, we unlock insights into the role of supervision, the importance of prior knowledge of anomaly types, and the principles of designing robust detection algorithms. On top of them, we summarize a few promising future research directions for anomaly detection, along with the fully released benchmark suite for evaluating new algorithms. 

\system can extend to understand the algorithm performance with (\textit{i}) mixed types of anomalies; (\textit{ii}) different levels of  (intrinsic) anomaly ratio; and (\textit{iii}) more data modalities.
Also, future benchmarks can consider the latest algorithms \cite{chang2022data,liu2022unsupervised,shenkar2021anomaly}, and curate datasets from emerging fields like drug discovery \cite{huang2021therapeutics}, molecule optimization \cite{fu2021mimosa,fu2020core}, interpretability and explainability \cite{pang2021toward,xu2021beyond}, and bias and fairness
\cite{davidson2020framework,hu2022uncovering,pang2021toward,shekhar2021fairod,song2021deep,zhang2021towards}.

%% file: appendix.tex
\appendix
\section*{Supplementary Material for \textit{\system: Anomaly Detection Benchmark}}

\textit{Additional information on related works, algorithms, datasets, and additional experiment settings and results}

\setcounter{table}{0}
\setcounter{figure}{0}

\renewcommand{\thetable}{\Alph{section}\arabic{table}}
\renewcommand{\thefigure}{\Alph{section}\arabic{figure}}

\section{Related Works with More Details}

We provide more details on existing AD algorithms and benchmarks, and the primary content discussed in \S \ref{sec:related}. 

\subsection{Unsupervised Methods}
\label{appendix:subsec:unsupervised}

\textbf{Unsupervised Methods by Assuming Anomaly Data Distributions}.
Unsupervised AD methods are proposed with different assumptions of data distribution \cite{aggarwal2017introduction}, e.g., anomalies located in low-density regions, and their performance often depends on the agreement between the input data and the algorithm assumption(s).
Many unsupervised methods have been proposed in the last few decades \cite{aggarwal2017introduction,bergman2019classification,pang2021deep,unifying_shallow_deep,zhao2019pyod}, which can be roughly categorized into shallow and deep (neural network) methods. The former often carry better interpretability, while the latter handles large, high-dimensional data better. Recent book \cite{aggarwal2017introduction} and surveys \cite{pang2021deep,unifying_shallow_deep} provide great details on these algorithms, while we further elaborate
on a few representative unsupervised methods. More algorithm details and hyperparameter settings are illustrated in Appx. \S \ref{appendix:algorithms}

\textbf{Representative Shallow Methods}. Some representative shallow methods include: (\textit{i}) Isolation Forest (IForest) \cite{liu2008isolation} builds an ensemble of trees to isolate the data points and defines the anomaly score as the distance of an individual instance to the root;
(\textit{ii})
One-class SVM (OCSVM) \cite{scholkopf1999support} maximizes the margin between origin and the normal samples, where the decision boundary is the hyper-plane that determines the margin;
and (\textit{iii})
Empirical-Cumulative-distribution-based Outlier Detection (ECOD) \cite{ECOD} computes the empirical cumulative distribution per dimension of the input data, and then aggregates the tail probabilities per dimension for calculating the anomaly score. 

\textbf{Representative Deep Methods}. Deep (neural network) methods have gained more attention recently, and we briefly review some representative ones in this section. 
Deep Autoencoding Gaussian Mixture Model (DAGMM) \cite{DAGMM} jointly optimizes the deep autoencoder and the Gaussian mixture model in an end-to-end neural network fashion. The joint optimization balances autoencoding reconstruction, density estimation of latent representation, and regularization and helps the autoencoder escape from less attractive local optima and further reduce reconstruction errors, avoiding pre-training.
Deep Support Vector Data Description (DeepSVDD) \cite{DeepSVDD} trains a neural network to learn a transformation that minimizes the volume of a data-enclosing hypersphere in the output space, and calculates the anomaly score as the distance of transformed embedding to the center of the hypersphere.

\subsection{Supervised Methods}

Due to the difficulty and cost of collecting large-scale labeled data, fully-supervised anomaly detection is often impractical as it assumes the availability of labeled training data with both normal and anomaly samples \cite{pang2021deep}. Although some loss functions (e.g., focal loss \cite{lin2017focal}) are devised to address the class imbalance problem, they are often not specific for AD tasks. There also exist a few works \cite{gornitz2013toward, kawachi2018complementary} discussing the relationship between fully-supervised and semi-supervised AD methods, and argue that semi-supervised AD needs to be ground on the unsupervised learning paradigm instead of the supervised one for detecting both known and unknown anomalies. We implement several representative supervised classification algorithms in \system (as shown in Appx. \S \ref{appendix:algorithms}), and recommend interesting readers to recent machine learning books \cite{DBLP:books/sp/Aggarwal18,goodfellow2016deep} and scikit-learn \cite{pedregosa2011scikit} for more details about recent supervised methods designed for the classification tasks.

\subsection{Semi-supervised Methods}
\label{appendix:subsec:semisupervised}

Semi-supervised AD algorithms are designed to capitalize the supervision from partial labels, while keeping the ability to detect unseen types of anomalies.
To this end, some recent studies investigate efficiently using partially labeled data for improving detection performance, and leverage the unlabeled data to facilitate representation learning. 
We further provide some technical details on representative semi-supervised AD methods here. Please see Appx. \S \ref{appendix:algorithms} for more algorithm details and hyperparameter settings in \system.

\textbf{Representative Methods}. Extreme Gradient Boosting Outlier Detection (XGBOD) \cite{zhao2018xgbod} uses multiple unsupervised AD algorithms to extract useful representations from the underlying data that augment the predictive capabilities of an embedded supervised classifier on an improved feature space.
Deep Semi-supervised Anomaly Detection (DeepSAD) \cite{DeepSAD} is an end-to-end methodology considered the state-of-the-art method in semi-supervised anomaly detection. DeepSAD improves the DeepSVDD \cite{DeepSVDD} model by the inverse loss function for the labeled anomalies.
REPresentations for a random nEarest Neighbor distance-based method (REPEN) \cite{REPEN} proposes a ranking model-based framework, which unifies representation learning and anomaly detection to learn low-dimensional representations tailored for random distance-based detectors.
Deviation Networks (DevNet) \cite{devnet} constructs an end-to-end neural network for learning anomaly scores, which forces the network to produce statistically higher anomaly scores for identified anomalies than that of unlabeled data.
\textbf{P}airwise \textbf{R}elation prediction-based ordinal regression \textbf{Net}work (PReNet) \cite{PReNet} formulates the anomaly detection problem as a pairwise relation prediction task, which defines a two-stream ordinal regression neural network to learn the relation of randomly sampled instance pairs.
Feature Encoding with AutoEncoders for Weakly-supervised Anomaly Detection (FEAWAD) \cite{FEAWAD} leverages an autoencoder to encode the input data and utilize hidden representation, reconstruction residual vector and reconstruction error as the new representations for improving the DevNet \cite{devnet} and DAGMM \cite{DAGMM}.

\subsection{Existing AD Benchmarks}
\label{appendix:benchmarks}

As we show in Table \ref{tab:benchmark comparison}, there is a line of existing AD benchmarks. 
\cite{unifying_shallow_deep} discusses a unifying review of both the shallow and deep anomaly detection methods, but they mainly focus on the theoretical perspective and thus lack results from the experimental views.
\cite{campos2016evaluation} benchmarks 19 different unsupervised methods on 10 datasets, and analyzes the characteristics of density-based and clustering-based algorithms.
\cite{comparative_evaluation} tests 14 unsupervised anomaly detection methods on 15 public datasets, and analyzes the scalability, memory consumption, and robustness of different methods.
\cite{realistic_synthetic_data} proposes a generic process for the generation of realistic synthetic data. The synthetic normal instances are reconstructed from existing real-world benchmark data, while synthetic anomalies are in line with a characterizable deviation from the modeling of synthetic normal data.
\cite{meta_analysis} evaluates 8 unsupervised methods on 19 public  datasets, and produces a large corpus of synthetic anomaly detection datasets that vary in their construction across several dimensions that are important to real-world applications.
\cite{campos2016evaluation} compares 12 unsupervised anomaly detection approaches on 23 datasets, providing a characterization of benchmark datasets and their suitability as anomaly detection benchmark sets.

All these existing works lay the foundation of AD algorithm design, and we further improve the foundation by considering more datasets, algorithms, and comparison aspects.

\section{More Details on \system}

\subsection{\system Algorithm List}
\label{appendix:algorithms}

We organize all the algorithms in \system into the following three categories and report their hyperparameter settings which mainly refer to the settings of their original papers or repositories (e.g., PyOD\footnote{https://pyod.readthedocs.io/en/latest/pyod.html} and scikit-learn\footnote{https://scikit-learn.org/stable/}).

\textit{(i) \nunsup unsupervised algorithms}:
\begin{enumerate} [leftmargin=*,noitemsep]
    \item \textbf{Principal Component Analysis (PCA)} \cite{shyu2003novel}. PCA is a linear dimensionality reduction using singular value decomposition of the data to project it to a lower dimensional space. When used for AD, it projects the data to the lower dimensional space and then uses the reconstruction errors as the anomaly scores. If not specified, the default hyperparameters in PyOD are used for the PCA (and the other unsupervised algorithms deployed by PyOD).
    \item \textbf{One-class SVM (OCSVM)} \cite{scholkopf1999support}. OCSVM maximizes the margin between the origin and the normal samples, and defines the decision boundary as the hyperplane that determines the margin.
    \item \textbf{Local Outlier Factor (LOF)} \cite{LOF}. LOF measures the local deviation of the density of a given sample with respect to its neighbors.
    \item \textbf{Clustering Based Local Outlier Factor (CBLOF)} \cite{he2003discovering}. CBLOF calculates the anomaly score by first assigning samples to clusters, and then using the distance among clusters as anomaly scores.
    \item \textbf{Connectivity-Based Outlier Factor (COF)} \cite{tang2002enhancing}. COF uses the ratio of the average chaining distance of data points and the average chaining distance of $k$-th nearest neighbor of the data point, as the anomaly score for observations.
    \item \textbf{Histogram- based outlier detection (HBOS)} \cite{goldstein2012histogram}. HBOS assumes feature independence and calculates the degree of outlyingness by building histograms.
    \item \textbf{K-Nearest Neighbors (KNN)} \cite{ramaswamy2000efficient}. KNN views the anomaly score of the input instance as the distance to its $k$-th nearest neighbor.
    \item \textbf{Subspace Outlier Detection (SOD)} \cite{kriegel2009outlier}. SOD aims to detect outliers in varying subspaces of high-dimensional feature space.
    \item \textbf{Copula Based Outlier Detector (COPOD)} \cite{COPOD}. COPOD is a hyperparameter-free, highly interpretable outlier detection algorithm based on empirical copula models.
    \item \textbf{Empirical-Cumulative-distribution-based Outlier Detection (ECOD)} \cite{ECOD}. ECOD is a hyperparameter-free, highly interpretable outlier detection algorithm based on empirical CDF functions. Basically, it uses ECDF to estimate the density of each feature independently, and assumes that outliers locate the tails of the distribution.
    \item \textbf{Deep Support Vector Data Description (DeepSVDD)} \cite{DeepSVDD}. DeepSVDD trains a neural network while minimizing the volume of a hypersphere that encloses the network representations of the data, forcing the network to extract the common factors of variation.
    \item \textbf{Deep Autoencoding Gaussian Mixture Model (DAGMM)} \cite{DAGMM}. DAGMM utilizes a deep autoencoder to generate a low-dimensional representation and reconstruction error for each input data point, which is further fed into a Gaussian Mixture Model (GMM). We train the DAGMM for 200 epochs with 256 batch size, where the patience of early stopping is set to 50. The learning rate of Adam \cite{kingma2014adam} optimizer is 0.0001 and is decayed once the number of epochs reaches 50. The latent dimension of DAGMM is set to 1 and the number of Gaussian mixture components is set to 4. The $\lambda_{1}$ and $\lambda_{2}$ for energy and covariance in the objective function are set to 0.1 and 0.005, respectively.
    \item \textbf{Lightweight on-line detector of anomalies (LODA)} \cite{pevny2016loda}. LODA is an ensemble method and is particularly useful in domains where a large number of samples need to be processed in real-time or in domains where the data stream is subject to concept drift and the detector needs to be updated online.
    \item \textbf{Isolation Forest (IForest)} \cite{liu2008isolation}. IForest isolates observations by randomly selecting a feature and then randomly selecting a split value between the maximum and minimum values of the selected feature.
\end{enumerate}

\textit{(ii) \nsemi semi-supervised algorithms}:
\begin{enumerate} [leftmargin=*,noitemsep]
    \item \textbf{Semi-Supervised Anomaly Detection via Adversarial Training (GANomaly)} \cite{GANomaly}. A GAN-based method that defines the reconstruction error of the input instance as the anomaly score. We replace the convolutional layer in the original GANomaly with the dense layer with tanh activation function for evaluating it on the tabular data, where the hidden size of the encoder-decoder-encoder structure of GANomaly is set to half of the input dimension. We train the GANomaly for 50 epochs with 64 batch size, where the SGD \cite{ruder2016overview} optimizer with 0.01 learning rate and 0.7 momentum is applied for both the generator and the discriminator.
    
    \item \textbf{Deep Semi-supervised Anomaly Detection (DeepSAD)} \cite{DeepSAD}. A deep one-class method that improves the unsupervised DeepSVDD \cite{DeepSVDD} by penalizing the inverse of the distances of anomaly representation such that anomalies must be mapped further away from the hypersphere center. The hyperparameter $\eta$ in the loss function is set to 1.0, where DeepSAD is trained for 50 epochs with 128 batch size. Adam optimizer with 0.001 learning rate and $10^{-6}$ weight decay is applied for updating the network parameters. DeepSAD additionally employs an autoencoder for calculating the initial center of the hypersphere, where the autoencoder is trained for 100 epochs with 128 batch size, and optimized by Adam optimizer with learning rate 0.001 and $10^{-6}$ weight decay.
    
    \item \textbf{REPresentations for a random nEarest Neighbor distance-based method (REPEN)} \cite{REPEN}. A neural network-based model that leverages transformed low-dimensional representation for random distance-based detectors. The hidden size of REPEN is set to 20, and the margin of triplet loss is set to 1000. REPEN is trained for 30 epochs with 256 batch size, where the total number of steps (batches of samples) is set to 50. Adadelta \cite{zeiler2012adadelta} optimizer with 0.001 learning rate and 0.95 $\rho$ is applied to update network parameters.
    
    \item \textbf{Deviation Networks (DevNet)} \cite{devnet}. A neural network-based model uses a prior probability to enforce a statistical deviation score of input instances. The margin hyperparameter $a$ in the deviation loss is set to 5. DevNet is trained for 50 epochs with 512 batch size, where the total number of steps is set to 20. RMSprop \cite{ruder2016overview} optimizer with 0.001 learning rate and 0.95 $\rho$ is applied to update network parameters.
    
    \item \textbf{Pairwise Relation prediction-based ordinal regression Network (PReNet)} \cite{PReNet}. A neural network-based model that defines a two-stream ordinal regression to learn the relation of instance pairs. The score targets of \{unlabeled, unlabeled\}, \{labeled, unlabeled\} and \{labeled, labeled\} sample pairs are set to 0, 4 and 8, respectively. PReNet is trained for 50 epochs with 512 batch size, where the total number of steps is set to 20. RMSprop optimizer with a learning rate of 0.001 and 0.01 weight decay is applied to update network parameters.
    
    \item \textbf{Feature Encoding With Autoencoders for Weakly Supervised Anomaly Detection (FEAWAD)} \cite{FEAWAD}. A neural network-based model that incorporates the network architecture of DAGMM \cite{DAGMM} with the deviation loss of DevNet \cite{devnet}. FEAWAD is trained for 30 epochs with 512 batch size, where the total number of steps is set to 20. Adam optimizer with 0.0001 learning rate is applied to update network parameters.
    
    \item \textbf{Extreme Gradient Boosting Outlier Detection (XGBOD)} \cite{zhao2018xgbod}. XGBOD first uses the passed-in unsupervised outlier detectors to extract richer representations of the data and then concatenates the newly generated features to the original feature for constructing the augmented feature space. An XGBoost classifier is then applied to this augmented feature space. We use the default hyperparameters in PyOD.
    
\end{enumerate}

\textit{(iii) \nsup supervised algorithms}:
\begin{enumerate} [leftmargin=*,noitemsep]
    \item \textbf{Naive Bayes (NB)} \cite{bayes1763lii}. NB methods are based on applying Bayes’ theorem with the “naive” assumption of conditional independence between every pair of features given the value of the class variable. We use the Gaussian NB in \system.
    
    \item  \textbf{Support Vector Machine (SVM)} \cite{cortes1995support}. SVM is effective in high-dimensional spaces and could still be effective in cases where the number of dimensions is greater than the number of samples. We use the default hyperparameters in scikit-learn for SVM (and for the following MLP and RF).
    
    \item \textbf{Multi-layer Perceptron (MLP)} \cite{rosenblatt1958perceptron}. MLP uses the binary cross entropy loss to update network parameters.
    
    \item \textbf{Random Forest (RF)} \cite{liaw2002classification}. RF is a meta estimator that fits several decision tree classifiers on various sub-samples of the dataset and uses averaging to improve the predictive accuracy and control over-fitting.
    
    \item \textbf{eXtreme Gradient Boosting (XGBoost)} \cite{XGBoost}. XGBoost is an optimized distributed gradient boosting method designed to be highly efficient, flexible, and portable. We use the default hyperparameter settings in the XGBoost official repository\footnote{https://xgboost.readthedocs.io/en/stable/parameter.html}.
    
    \item \textbf{Highly Efficient Gradient Boosting Decision Tree (LightGBM)} \cite{ke2017lightgbm}. LightGBM is a gradient boosting framework that uses tree-based learning algorithms with faster training speed, higher efficiency, lower memory usage, and better accuracy. The default hyperparameter settings in the LightGBM official repository\footnote{https://lightgbm.readthedocs.io/en/latest/Parameters.html} are used.
    
    \item \textbf{Categorical Boosting (CatBoost)} \cite{prokhorenkova2018catboost}. CatBoost is a fast, scalable, high-performance gradient boosting on decision trees. CatBoost uses the default hyperparameter settings in its official repository\footnote{https://catboost.ai/en/docs/references/training-parameters/}.
    
    \item \textbf{Residual Nets (ResNet)} \cite{gorishniy2021revisiting}. This method introduces a ResNet-like architecture \cite{resnet} for tabular based data. ResNet is trained for 100 epochs with 64 batch size. AdamW \cite{loshchilov2017decoupled} optimizer with 0.001 learning rate is applied to update network parameters.
    
    \item \textbf{Feature Tokenizer + Transformer (FTTransformer)} \cite{gorishniy2021revisiting}. FTTransformer is an effective adaptation of the Transformer architecture \cite{vaswani2017attention} for tabular data. FTTransformer is trained for 100 epochs with 64 batch size. AdamW optimizer with 0.0001 learning rate and $10^{-5}$ weight decay is applied to update network parameters.
\end{enumerate}

\clearpage
\newpage 

\subsection{\system Dataset List}
\label{appendix:datasets}
\begin{figure*}[!ht]
    \centering
    \includegraphics[width=0.7\linewidth]{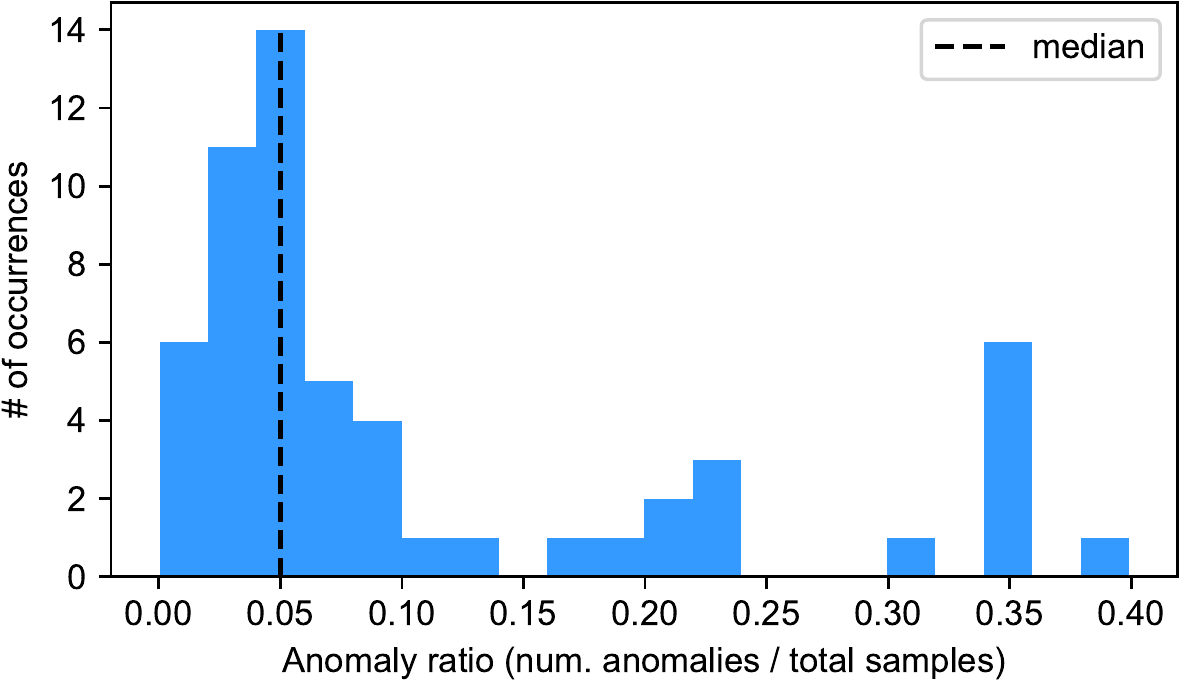}
    \caption{Distribution of anomaly ratios in \ndatasets datasets in \system, where 40 datasets' anomaly ratio is below 10\% (median=5\%).}
    \label{fig:ratio}
\end{figure*}

\textbf{Overview}. As described in \S \ref{subsec:algo_datasets}, \system is the largest AD benchmark with \ndatasets datasets. More specifically, Table \ref{table:datasets} shows the datasets used in \system, covering many application domains, including healthcare (e.g., disease diagnosis), audio and language processing (e.g., speech recognition), image processing (e.g., object identification), finance (e.g., financial fraud detection), and more, where we show this information in the last column. We resample the sample size to 1,000 for those datasets smaller than 1,000, and use the subsets of 10,000 for those datasets greater than 10,000 due to the computational cost. Fig. \ref{fig:ratio} provides the anomaly ratio distribution of the datasets, where the median is equal to 5\%. We release all the datasets and their raw version(s) when possible at \url{https://github.com/Minqi824/ADBench/tree/main/datasets}.

\textbf{Newly-added Datasets in \system}. Since most of the public datasets are relatively small and simple, we introduce \ndatasetscomplex more complex datasets from CV and NLP domains with more samples and richer features in \system (highlighted in Table \ref{table:datasets} in \textcolor{blue}{blue}). 

\textbf{\textit{Reasoning of Using CV/NLP Datasets}}. It is often challenging to directly run large CV and NLP datasets on selected shallow methods, e.g., OCSVM \cite{scholkopf1999support} and kNN \cite{ramaswamy2000efficient} with high time complexity, we follow DeepSAD \cite{DeepSAD}, ADIB \cite{deecke2021transfer}, and DATE \cite{manolache2021date} to extract representations of CV and NLP datasets by neural networks for downstream detection tasks. More specifically, ADIB \cite{deecke2021transfer} shows that ``transferring features from semantic tasks can provide powerful and generic representations for various AD problems'', which is true even when the pre-trained task is only loosely related to downstream AD tasks. Similarly, DeepSAD \cite{DeepSAD} uses pre-trained autoencoder to extract features for training classical AD detectors like OCSVM \cite{scholkopf1999support} and IForest \cite{liu2008isolation}. For NLP datasets, DATE \cite{manolache2021date} uses fastText \cite{joulin2017bag} and Glove \cite{pennington2014glove} embeddings for evaluating classical AD methods (e.g., OCSVM \cite{scholkopf1999support} and IForest \cite{liu2008isolation}) against proposed methods in NLP datasets.

We want to elaborate further on the reasons for adapting CV and NLP datasets for tabular AD. First, some shallow models, such as OCSVM \cite{scholkopf1999support}, cannot directly run on (large, high-dimensional) CV datasets. Second, it is interesting to see whether tabular AD methods can work on CV/NLP data representations, which carry values in real-world applications where deep models are infeasible to run. Moreover, the extracted representations often lead to better downstream detection results \cite{deecke2021transfer}. Thus, we extract features from CV and NLP datasets by deep models to create ``tabular’’ versions of them. Although not perfect, this may provide insights into shallow methods’ performance on (originally infeasible) CV and NLP datasets.

\textit{\textbf{CV Datasets}}: For MNIST-C, we set original MNIST images to be normal and corrupted images in MNIST-C to be abnormal, like in the recent work \cite{lee2022semi}. For MVTec-10, we test different AD algorithms on the 15 image sets, where anomalies correspond to various manufacturing defects. For CIFAR10, FashionMNIST, and SVHN, we follow previous works \cite{DeepSVDD, DeepSAD} and set one of the multi-classes as normal and downsample the remaining classes to $5\%$ of the total instances as anomalies by default, and report the average results over all the respective classes.

\textit{\textbf{NLP Datasets}}: For Amazon and Imdb, we regard the original negative class as the anomaly class. 
For Yelp, we regard the reviews of 0 and 1 stars as the anomaly class, and the reviews of 3 and 4 stars as the normal class.
For 20newsgroups dataset, like in DATE \cite{manolache2021date} and CVDD \cite{ruff2019self}, we only take the articles from the six top-level classes: \textit{computer}, \textit{recreation}, \textit{science}, \textit{miscellaneous}, \textit{politics}, \textit{religion}. Similarly, for the multi-classes datasets 20newsgroups and Agnews, we set one of the classes as normal and downsample the remaining classes to $5\%$ of the total instances as anomalies.

\textbf{\textit{Backbone Choices of Feature Extraction}}. 
Pretrained models are applied to extract data embedding from CV and NLP datasets to access these more complex representations. 
For CV datasets, following \cite{bergmann2019mvtec} and \cite{reiss2021panda}, we use ResNet18\footnote{https://pytorch.org/hub/pytorch\_vision\_resnet/} \cite{resnet} pretrained on the ImageNet \cite{ImageNet} to extract meaningful embedding after the last average pooling layer. We also provide the embedding version that are extracted by the ImageNet-pretrained ViT\footnote{https://github.com/lukemelas/PyTorch-Pretrained-ViT} \cite{dosovitskiy2020image}.
For NLP datasets, instead of using traditional embedding methods like fastText \cite{bojanowski2017enriching, joulin2017bag} or Glove \cite{pennington2014glove}, we apply BERT\footnote{https://huggingface.co/bert-base-uncased} \cite{BERT} pretrained on the BookCorpus and English Wikipedia to extract more enriching embedding of the [CLS] token. In addition, we provide the embedding version that are extracted by the pretrained RoBERTa\footnote{https://huggingface.co/roberta-base} \cite{liu2019roberta} in our codebase\footnote{\url{https://github.com/Minqi824/ADBench/tree/main/datasets}}. Although we release all the generated datasets for completeness, we analyze the results based on the datasets generated by BERT and ResNet18. Future work may consider analyzing the impact of backbones on detection performance.

\input{tables/datasets}
\input{tables/MVTec-AD}
\input{tables/20news}

\clearpage
\newpage

\subsection{Additional Demonstration of Synthetic Anomalies for \S \ref{subsub:types}}

In addition to Fig. \ref{fig:synthetic anomalies} that demonstrates the synthetic anomalies on Lymphography dataset in \S \ref{subsub:types}, we provide another example here for Ionosphere data.

\begin{figure}[h]
    \begin{minipage}[c]{0.8\textwidth}
     \centering
     \begin{subfigure}[b]{0.242\textwidth}
         \centering
         \includegraphics[width=\textwidth]{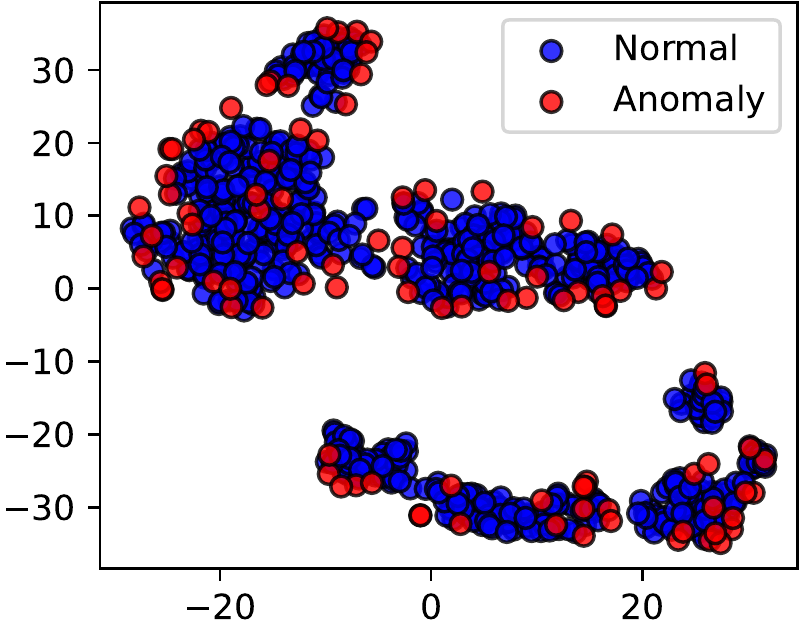}
         \caption{Local}
         \label{fig:demo_add_local}
     \end{subfigure}
     \hfill
     \begin{subfigure}[b]{0.242\textwidth}
         \centering
         \includegraphics[width=\textwidth]{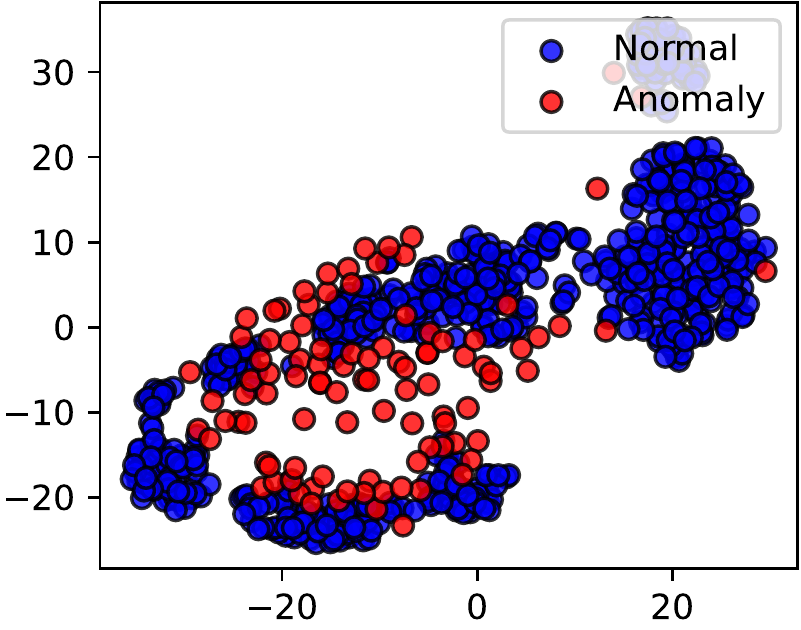}
         \caption{Global}
         \label{fig:demo_add_global}
     \end{subfigure}
     \hfill
     \begin{subfigure}[b]{0.242\textwidth}
         \centering
         \includegraphics[width=\textwidth]{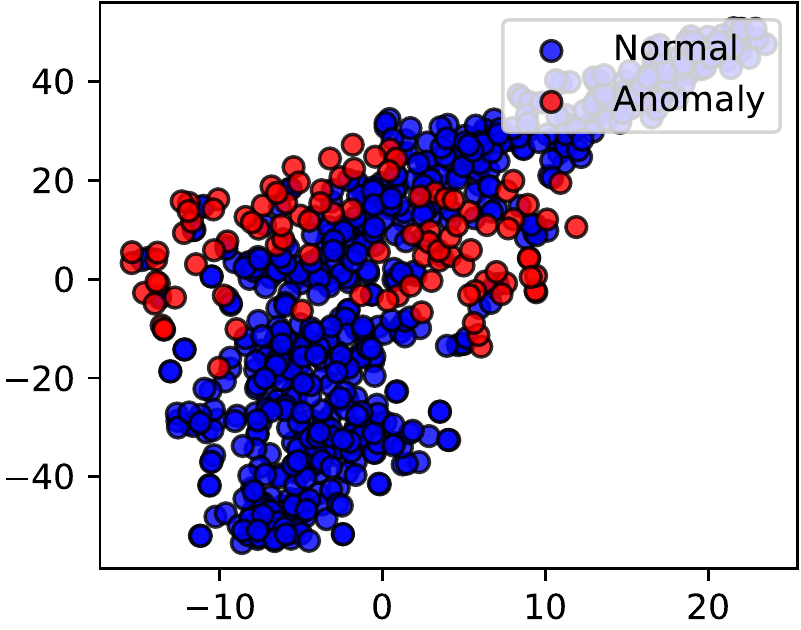}
         \caption{Dependency}
         \label{fig:demo_add_dependency}
     \end{subfigure}
     \hfill
     \begin{subfigure}[b]{0.242\textwidth}
         \centering
         \includegraphics[width=\textwidth]{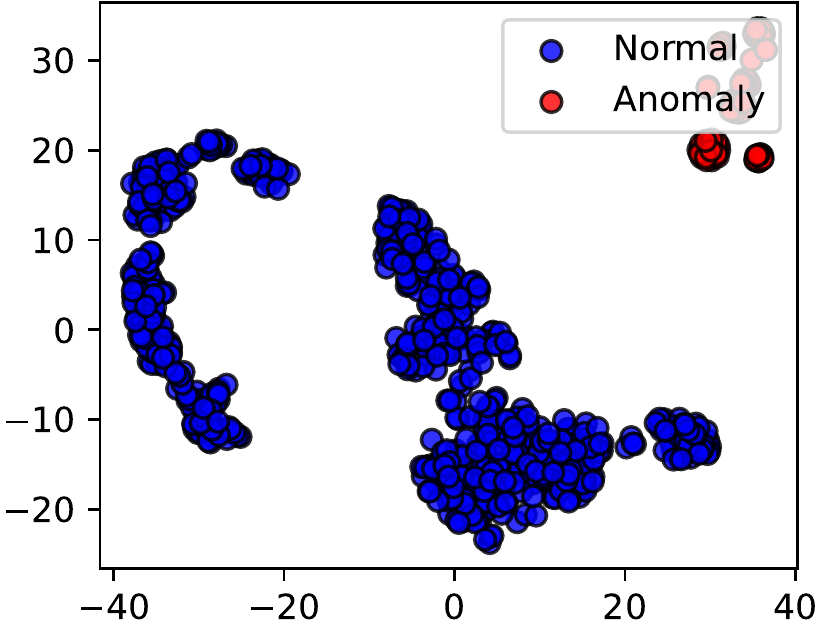}
         \caption{Clustered}
         \label{fig:demo_add_clusdtered}
     \end{subfigure}
    \end{minipage}
    \hfill
    \begin{minipage}[c]{0.19\textwidth}
     \caption{Illustration of four types of synthetic anomalies shown on Ionosphere dataset.}
     \label{fig:synthetic anomalies add}
    \end{minipage}
\end{figure}

\subsection{Open-source Release}
As mentioned before, the full experiment code, datasets, and examples of benchmarking new algorithms are available at \url{https://github.com/Minqi824/ADBench}. We specify the key environment setting of using \system, e.g., \texttt{scikit-learn==0.20.3}, \texttt{pyod==0.9.8}, etc. With our interactive example in Jupyter notebooks, one may compare a newly proposed AD algorithm easily.

\clearpage
\newpage

\section{Details on Experiment Setting}
\label{appendix:exp_setting}

We provide additional details on experiment setting to \S \ref{subsec:exp_setting} in this section.

\textbf{General Experimental Settings}. Although unsupervised AD algorithms are primarily designed for the transductive setting (i.e., outputting the anomaly scores on the input data only other than making predictions on newcoming data), we adapt all the algorithms for the inductive setting to predict the newcoming data, which is helpful in applications and also common in popular AD library PyOD \cite{zhao2019pyod}, TODS \cite{lai2021tods,revisiting_time_series}, and PyGOD \cite{liu2022pygod}. Thus, we use $70\%$ data for training and the remaining $30\%$ as a testing set. We use stratified sampling to keep the anomaly ratio consistent. We repeat each experiment 3 times and report the average.
The \ndatasetscomplex complex CV and NLP datasets are mainly considered for evaluating algorithm performance on the public datasets and are not included in the experiments of different types of anomalies and algorithm robustness, since such high-dimensional data could make it hard to generate synthetic anomalies (e.g., the Vine Copula is computationally expensive for fitting such high-dimensional data), or introduce too much noise in input data (e.g., the noise ratio of irrelevant features $50\%$ would lead to 384 noise features in the 768 input dimensions of NLP data). Future works may resort to the help of the latest generative methods like diffusion models \cite{Yang2022DiffusionMA}.

\textbf{Hyperparameter Settings}. For all the algorithms in  \system, we use their default hyperparameter (HP) settings in the original paper for a fair comparison. Specific values can be found in Appx.\ref{appendix:algorithms} and our codebase\footnote{\system repo: \url{https://github.com/Minqi824/ADBench}}. It is also acknowledged that it is possible to use a small hold-out data for hyperparameter tuning for semi- and fully-supervised methods \cite{soenen2021effect}, while we do not consider this setting in this work.

\textbf{Extensive Experiments}. In total \system conducts \nexps experiments, where each denotes one algorithm's result on a dataset under a specific setting. 
More specifically, we have 27,090 experiments in \S \ref{subsec:overall_performance}.
For \ndatasetssimple classical datasets:
\begin{itemize} [leftmargin=*,noitemsep]
    \item  Unsupervised methods on benchmark \st{real-world} datasets $\{$\nunsup algorithms, \ndatasetssimple datasets, 3 repeat times$\}$ leads to 1,974 experiments.
    \item Semi- and fully-supervised on real-world datasets $\{$16 algorithms, \ndatasetssimple datasets, 3 repeat times, 7 settings of labeled anomalies$\}$ leads to 15,792 experiments.
\end{itemize}

As we described in Appx. \ref{appendix:datasets}, we totally have 74 subclasses for the \ndatasetscomplex CV and NLP datasets, thus generating:
\begin{itemize} [leftmargin=*,noitemsep]
    \item  Unsupervised methods on benchmark datasets $\{$\nunsup algorithms, 74 subclasses, 1 repeat times$\}$ leads to 1,036 experiments.
    \item Semi- and fully-supervised on real-world datasets $\{$16 algorithms, 74 subclasses, 1 repeat times, 7 settings of labeled anomalies$\}$ leads to 8,288 experiments.
\end{itemize}

Additionally, we have 17,766 experiments for understanding the algorithm performances under four types of anomalies in \S \ref{exp:types}:

\begin{itemize} [leftmargin=*,noitemsep]
    \item  Unsupervised methods on benchmark \st{real-world}
 datasets $\{$\nunsup algorithms, \ndatasetssimple datasets, 3 repeat times$\}$ leads to 1,974 experiments.
    \item Semi- and fully-supervised on benchmark datasets $\{$16 algorithms, \ndatasetssimple datasets, 3 repeat times, 7 settings of labeled anomalies$\}$ leads to 15,792 experiments.
\end{itemize}

Finally, we have 53,580 experiments for evaluating the algorithm robustness under three settings of data noises and corruptions in \S \ref{exp:robustness}:

\begin{itemize} [leftmargin=*,noitemsep]
    \item For duplicated anomalies and irrelevant features $\{$\nmodels algorithms, \ndatasetssimple datasets, 3 repeat times, 5 settings of data noises, 2 scenarios$\}$ leads to 42,300 experiments.
    \item For annotation errors $\{$16 algorithms, \ndatasetssimple datasets, 3 repeat times, 5 settings of data noises$\}$ leads to 11,280 experiments.
\end{itemize}

\textbf{Computational Resources}. Classical anomaly detection models are run on an Intel i7-8700 @3.20 GHz, 16GB RAM, 12-core workstation. For deep learning models (especially for ResNet and FTTransformer), we run experiments on an NVIDIA Tesla V100 GPU accelerator. The model runtime on benchmark datasets is reported in Appx.~\S \ref{appendix:exp_realworld_results}.

\clearpage
\newpage

\section{Additional Experiment Results}
\label{appendix:exp_results}

\subsection{Additional Results for Overall Model Performance on Benchmark Datasets in \S \ref{subsec:overall_performance}} 
\label{appendix:exp_realworld_results}

\begin{figure}[h!]
     \centering
     \begin{subfigure}[b]{0.37\textwidth}
         \centering
         \includegraphics[width=\textwidth]{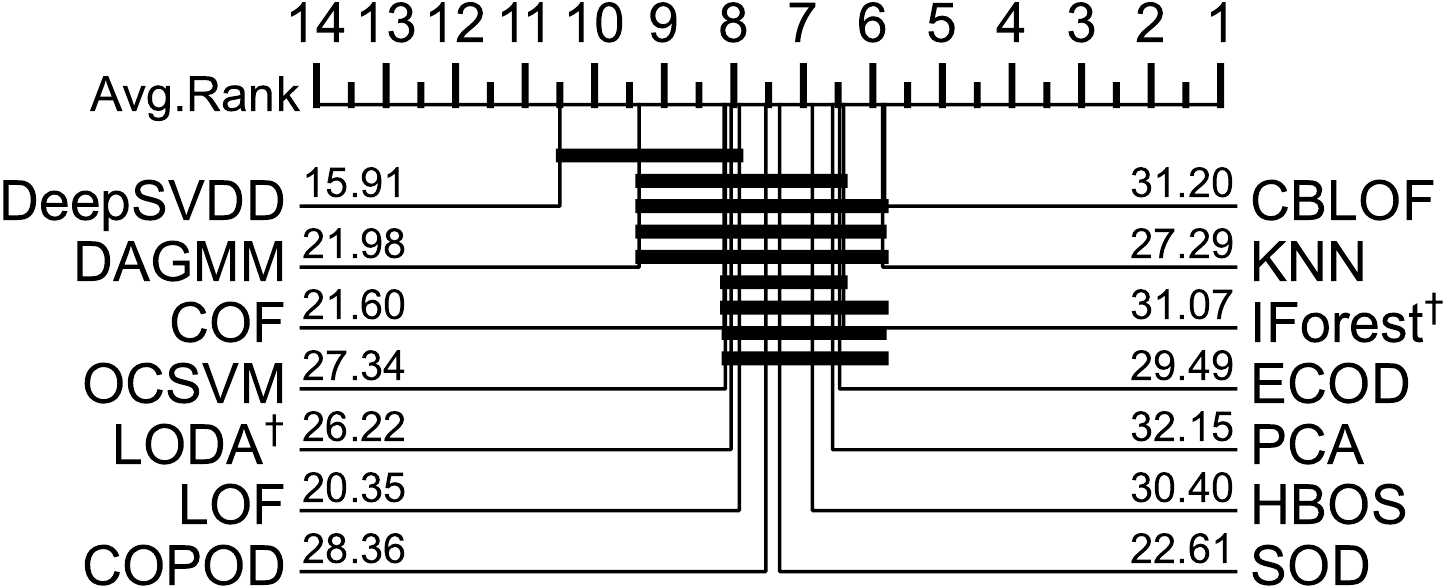}
         \caption{Avg. rank (lower the better) and avg. AUCPR (on each line) of unsupervised methods;
         groups of algorithms not statistically different are connected horizontally.}
         \label{fig:overall_unsupervised_pr}
     \end{subfigure}
     \hfill
     \begin{subfigure}[b]{0.61\textwidth}
         \centering
         \hfill
         \includegraphics[width=1\textwidth]{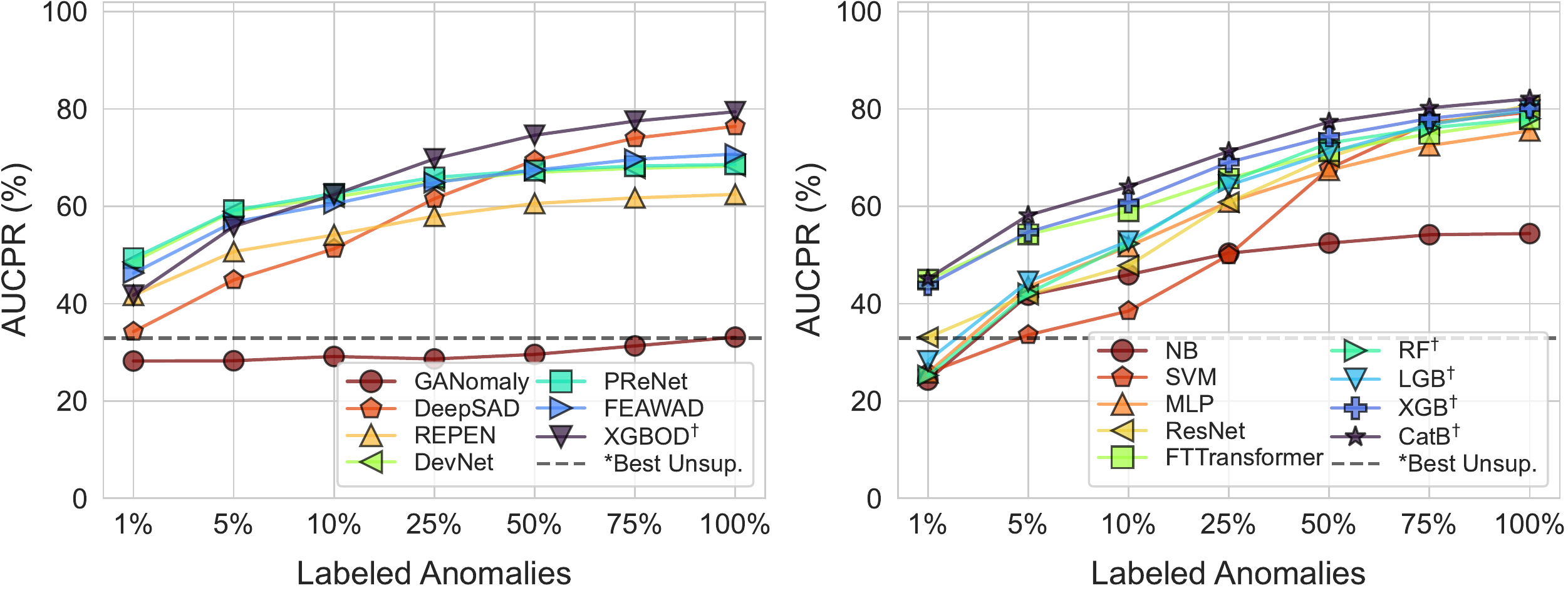}
         \caption{Avg. AUCPR (on \ndatasets datasets) vs. \% of labeled anomalies (x-axis); semi-supervised (left) and fully-supervised (right). Most label-informed algorithms outperform the best \textit{unsupervised} algorithm CBLOF (denoted as the dashed line) with 10\% labeled anomalies.}
         \label{fig:overall_supervised_pr}
     \end{subfigure}
     \caption{AD model's AUCPR on \ndatasets benchmark datasets. Generally, the AUCPR results are consistent with the AUCROC results in \S \ref{subsec:overall_performance}. (a) shows that no unsupervised algorithm can statistically outperform. (b) shows the AUCPR of semi- and supervised methods under varying ratios of labeled anomalies $\gamma_{l}$. The semi-supervised methods leverage the labels more efficiently w/ small $\gamma_{l}$. }
     \label{fig:overall model performance aucpr}
\end{figure}

In addition to the AUCROC results presented in \S \ref{subsec:overall_performance}, we also show the AUCPR results of model performance on \ndatasets benchmark datasets in Fig.~\ref{fig:overall model performance aucpr}, where the corresponding conclusions are similar to that of AUCROC results. There is still no statistically superior solution for unsupervised methods regarding AUCPR. Semi-supervised methods perform better than supervised methods when only limited label data is available, say the labeled anomalies $\gamma_{l}$ is less than $5\%$. Besides, we show that the semi-supervised GANomaly, which learns an intermediate representation of the normal data, performs worse than those anomaly-informed models leveraging labeled anomalies (see Fig.~\ref{fig:overall model performance aucpr}(b)). This conclusion verifies that merely capturing the normal behaviors is not enough for detecting the underlying anomalies, where the lack of knowledge about the true anomalies would lead to high false positives/negatives \cite{devnet_explainable, PReNet,devnet}.

\begin{figure}[h]
     \centering
     \begin{subfigure}[[b]{0.74\textwidth}
         \centering
         \includegraphics[width=\textwidth]{images_revision/Boxplot_AUCROC_0.01.pdf}
         \caption{AUCROC, $\gamma_{l}=1\%$}
     \end{subfigure}
     \hfill
    \begin{subfigure}[[b]{0.74\textwidth}
         \centering
         \includegraphics[width=\textwidth]{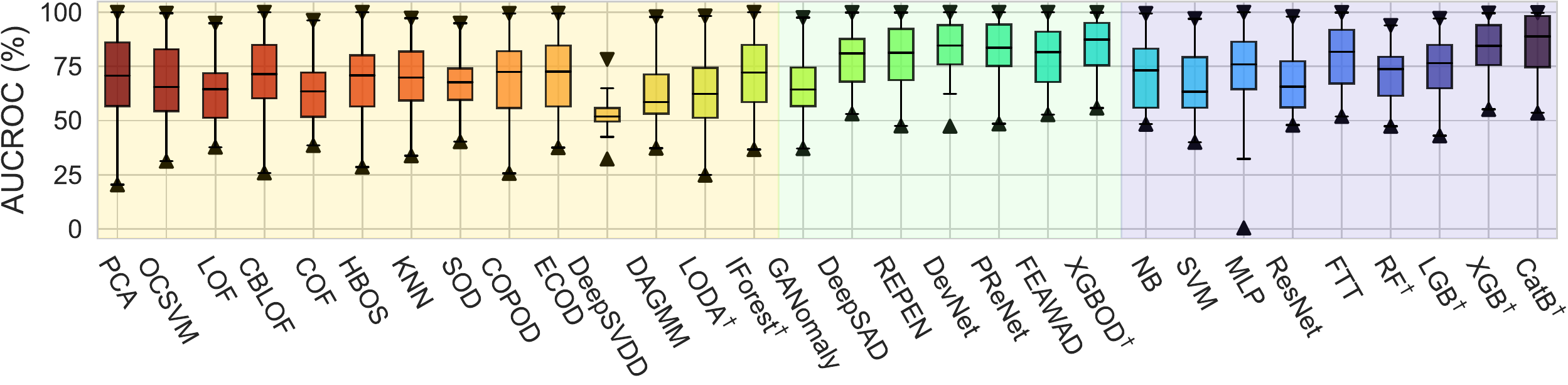}
         \caption{AUCROC, $\gamma_{l}=5\%$}
     \end{subfigure}
     \hfill
     \begin{subfigure}[[b]{0.74\textwidth}
         \centering
         \includegraphics[width=\textwidth]{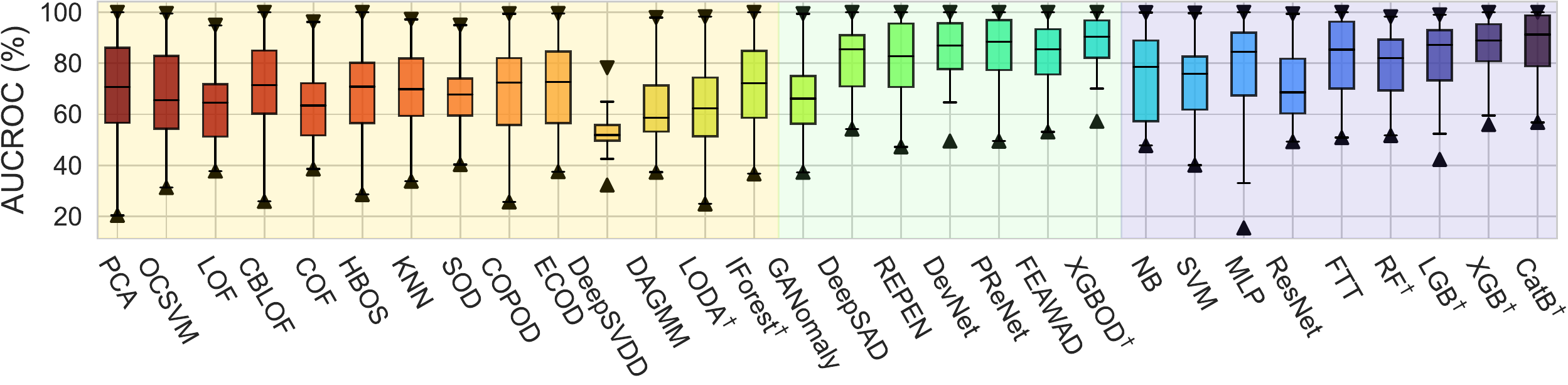}
         \caption{AUCROC, $\gamma_{l}=10\%$}
     \end{subfigure}
     \hfill
     \begin{subfigure}[[b]{0.74\textwidth}
         \centering
         \includegraphics[width=\textwidth]{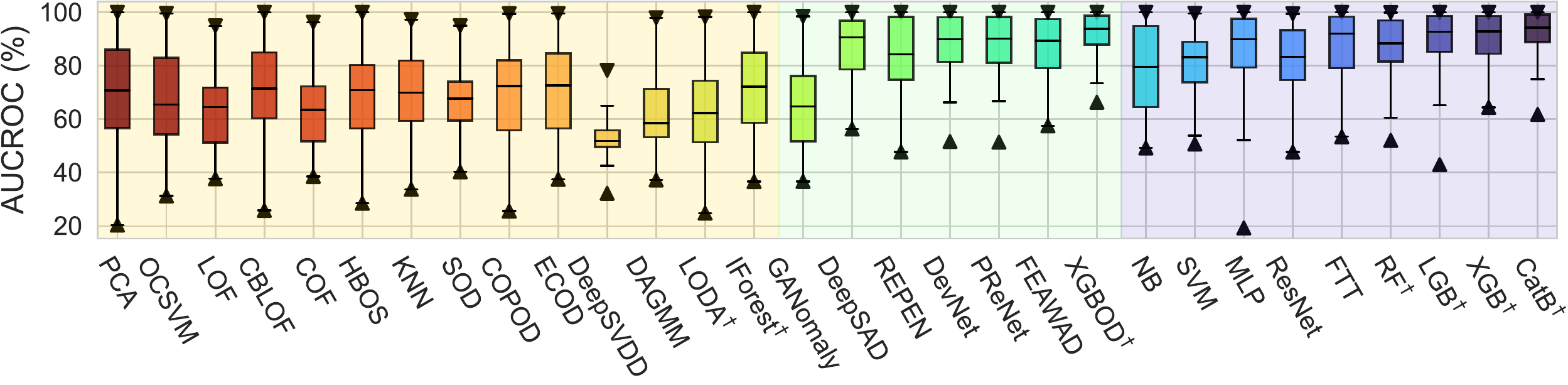}
         \caption{AUCROC, $\gamma_{l}=25\%$}
     \end{subfigure}
     \hfill
     \begin{subfigure}[[b]{0.74\textwidth}
         \centering
         \includegraphics[width=\textwidth]{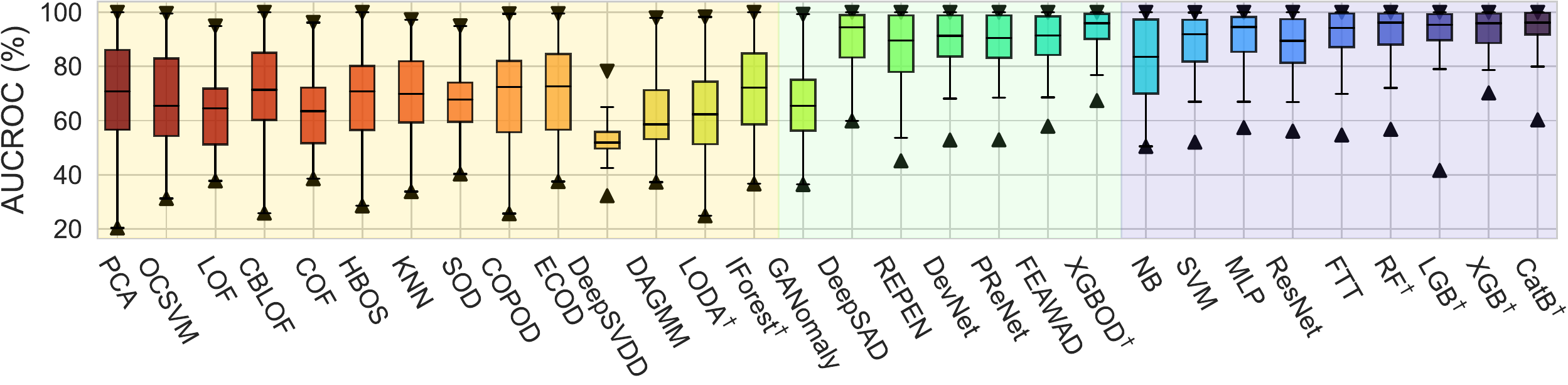}
         \caption{AUCROC, $\gamma_{l}=50\%$}
     \end{subfigure}
      \begin{subfigure}[[b]{0.74\textwidth}
         \centering
         \includegraphics[width=\textwidth]{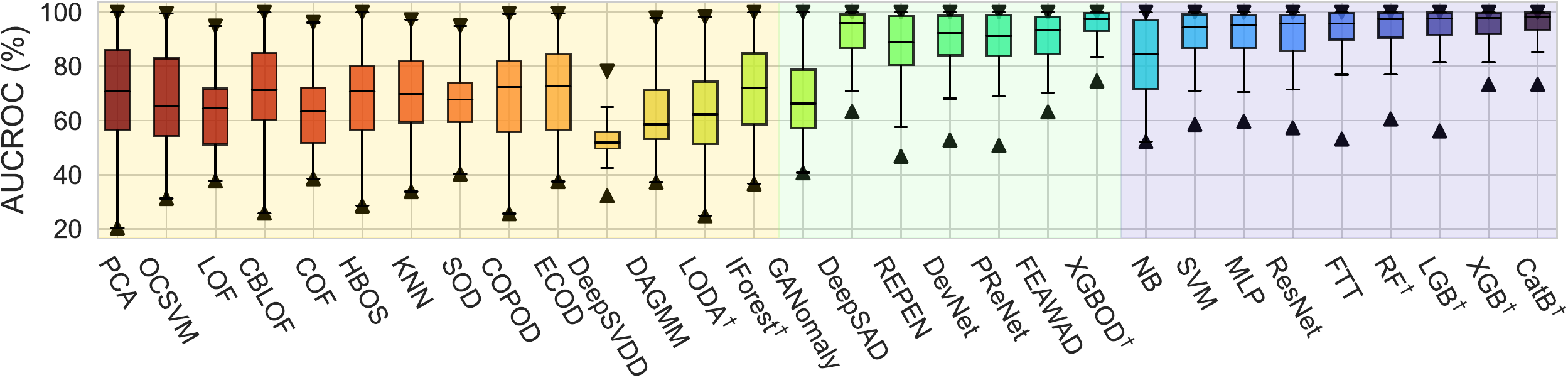}
         \caption{AUCROC, $\gamma_{l}=75\%$}
     \end{subfigure}
     \hfill
     \begin{subfigure}[[b]{0.74\textwidth}
         \centering
         \includegraphics[width=\textwidth]{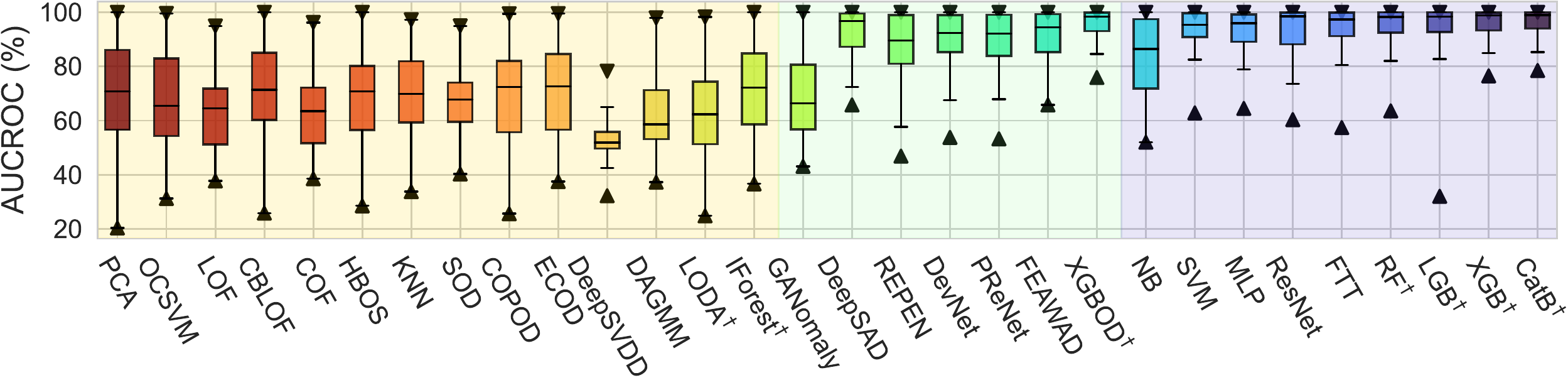}
         \caption{AUCROC, $\gamma_{l}=100\%$}
     \end{subfigure}
     \caption{Boxplot of AUCROC. We denote unsupervised methods in \crule[aliceyellow]{0.2cm}{0.2cm} (light yellow), semi-supervised methods in \crule[alicegreen]{0.2cm}{0.2cm} (light green), and supervised methods in \crule[aliceblue]{0.2cm}{0.2cm} (light purple). Consistent with the CD diagrams, we notice that none of the unsupervised methods visually outperform. 
     } 
     \label{fig:boxplot_aucroc}
\end{figure}

\begin{figure}[h]
     \centering
     \begin{subfigure}[[b]{0.74\textwidth}
         \centering
         \includegraphics[width=\textwidth]{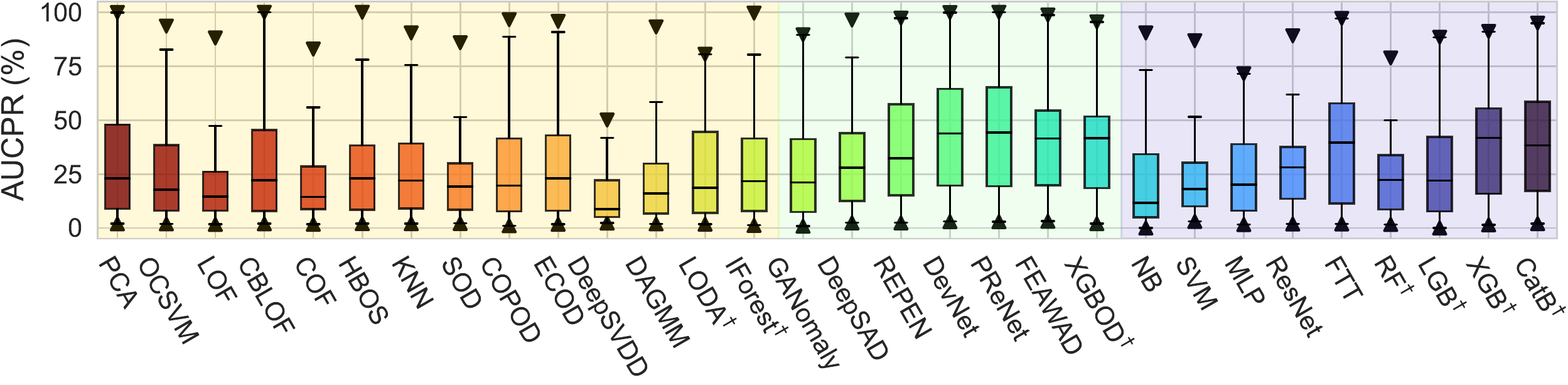}
         \caption{AUCPR, $\gamma_{l}=1\%$}
     \end{subfigure}
     \hfill
    \begin{subfigure}[[b]{0.74\textwidth}
         \centering
         \includegraphics[width=\textwidth]{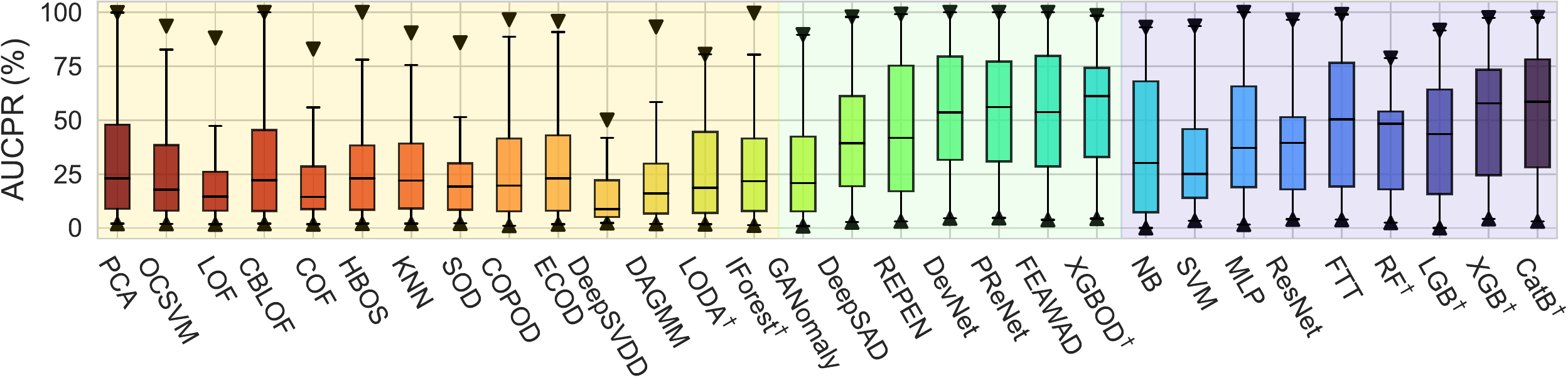}
         \caption{AUCPR, $\gamma_{l}=5\%$}
     \end{subfigure}
     \hfill
     \begin{subfigure}[[b]{0.74\textwidth}
         \centering
         \includegraphics[width=\textwidth]{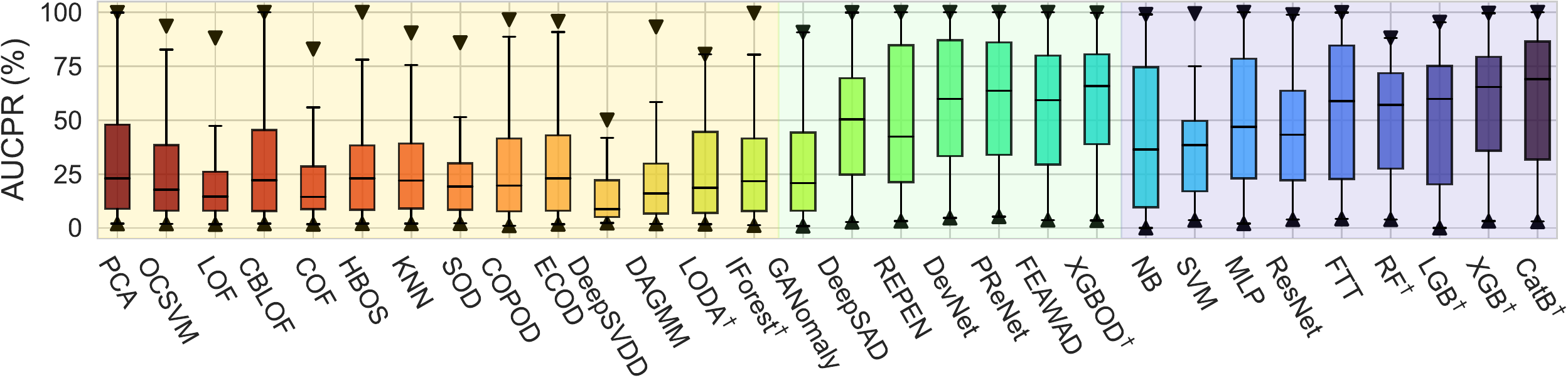}
         \caption{AUCPR, $\gamma_{l}=10\%$}
     \end{subfigure}
     \hfill
     \begin{subfigure}[[b]{0.74\textwidth}
         \centering
         \includegraphics[width=\textwidth]{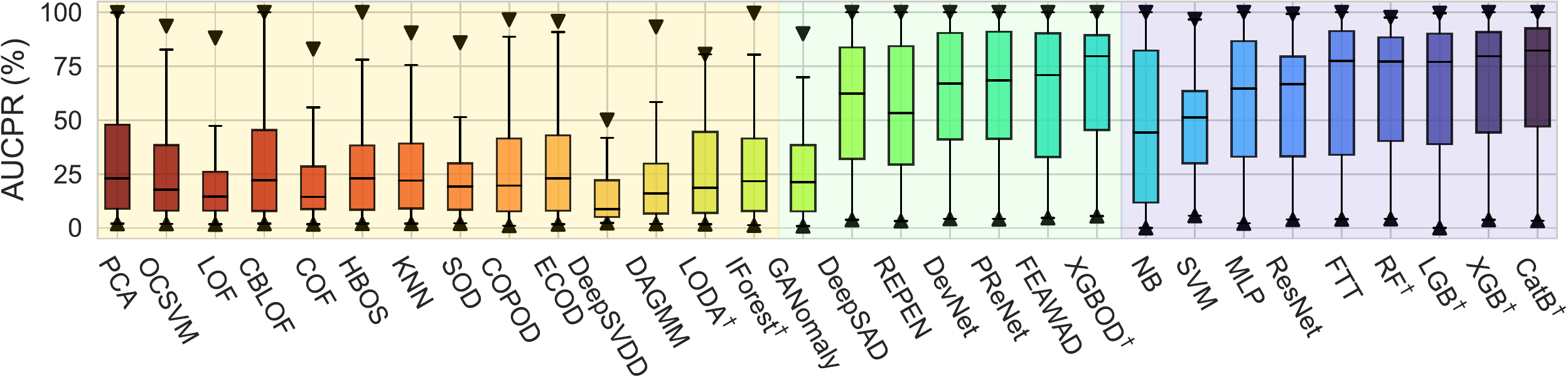}
         \caption{AUCPR, $\gamma_{l}=25\%$}
     \end{subfigure}
     \hfill
     \begin{subfigure}[[b]{0.74\textwidth}
         \centering
         \includegraphics[width=\textwidth]{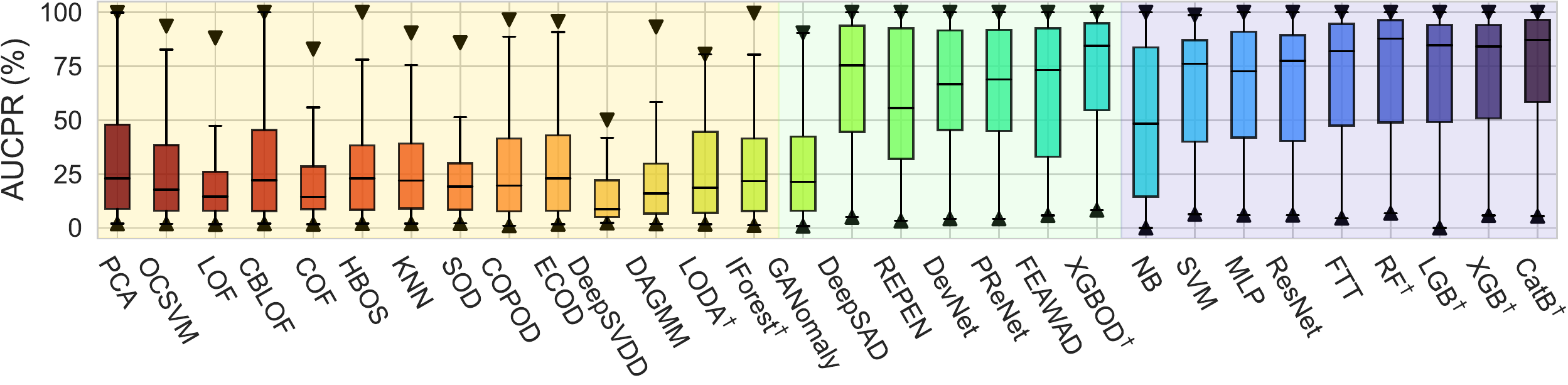}
         \caption{AUCPR, $\gamma_{l}=50\%$}
     \end{subfigure}
      \begin{subfigure}[[b]{0.74\textwidth}
         \centering
         \includegraphics[width=\textwidth]{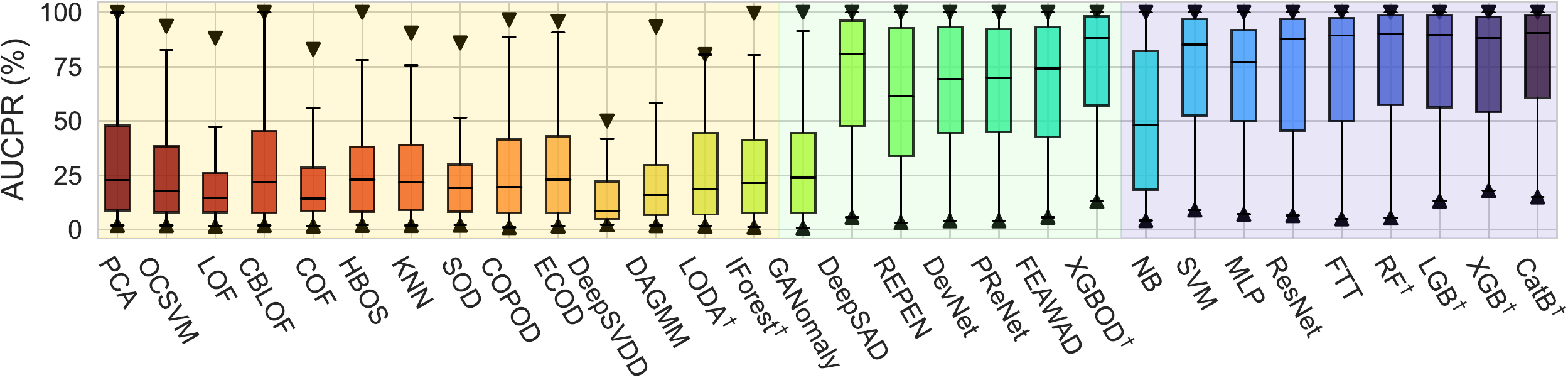}
         \caption{AUCPR, $\gamma_{l}=75\%$}
     \end{subfigure}
     \hfill
     \begin{subfigure}[[b]{0.74\textwidth}
         \centering
         \includegraphics[width=\textwidth]{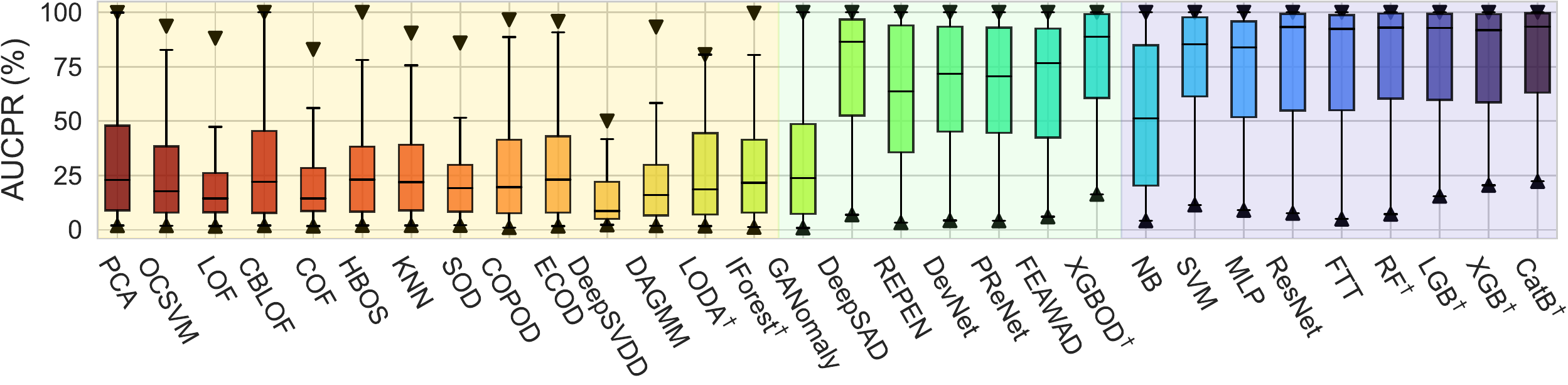}
         \caption{AUCPR, $\gamma_{l}=100\%$}
     \end{subfigure}
     \caption{Boxplot of AUCPR. We denote unsupervised methods in \crule[aliceyellow]{0.2cm}{0.2cm} (light yellow), semi-supervised methods in \crule[alicegreen]{0.2cm}{0.2cm} (light green), and supervised methods in \crule[aliceblue]{0.2cm}{0.2cm} (light purple). Consistent with the CD diagrams, we notice that none of the unsupervised methods visually outperform. 
     } 
     \label{fig:boxplot_aucpr}
\end{figure}

Fig.~\ref{fig:boxplot_aucroc} and \ref{fig:boxplot_aucpr} show the boxplots of AUCROC and AUCPR of \nmodels algorithms on the \ndatasets benchmark datasets. These results validate the no-free-lunch theorem, where no model is both the best and the most stable performer. For example, DeepSVDD and RF are the most stable detectors among un- and fully-supervised methods, respectively, but they are inferior to most of the other algorithms. Besides, IForest and CatB(oost) can be regarded as two very competitive methods among un- and fully-supervised methods, respectively, but their variances of model performance are relatively large compared to the other methods.

Additionally, we also present the full results in tables in \S \ref{appx:all_tables}.

\begin{figure*}[!h]
    \centering
    \includegraphics[width=1\linewidth]{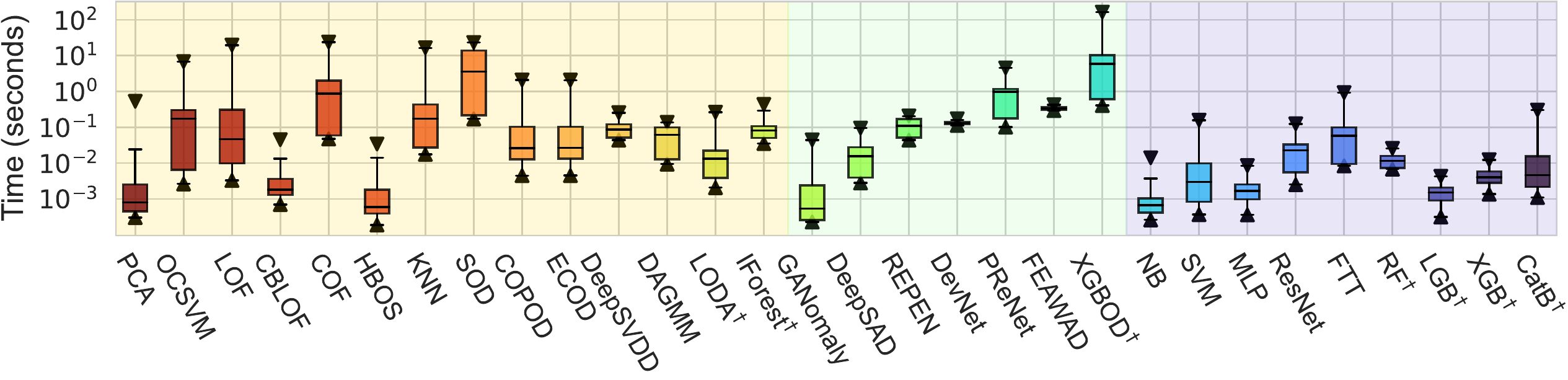}
    \caption{Inference time of included algorithms. We denote unsupervised methods in \crule[aliceyellow]{0.2cm}{0.2cm} (light yellow), semi-supervised methods in \crule[alicegreen]{0.2cm}{0.2cm} (light green), and supervised methods in \crule[aliceblue]{0.2cm}{0.2cm} (light purple). Consistent with the train time in Fig. \ref{sfig:runtime}, 
     PCA, HBOS, GANomaly and NB take the least inference time on test datasets, while more complex feature representation methods like SOD and XGBOD spend more time due to the search of the feature subspace.
    }
    \vspace{-0.25in}
    \label{fig:boxplot_infertime}
\end{figure*}

\clearpage
\subsection{Additional Results for Different Types of Anomalies \S \ref{exp:types}}
\begin{figure}[h!]
     \centering
     \begin{subfigure}[b]{0.45\textwidth}
         \centering
         \includegraphics[width=1\textwidth]{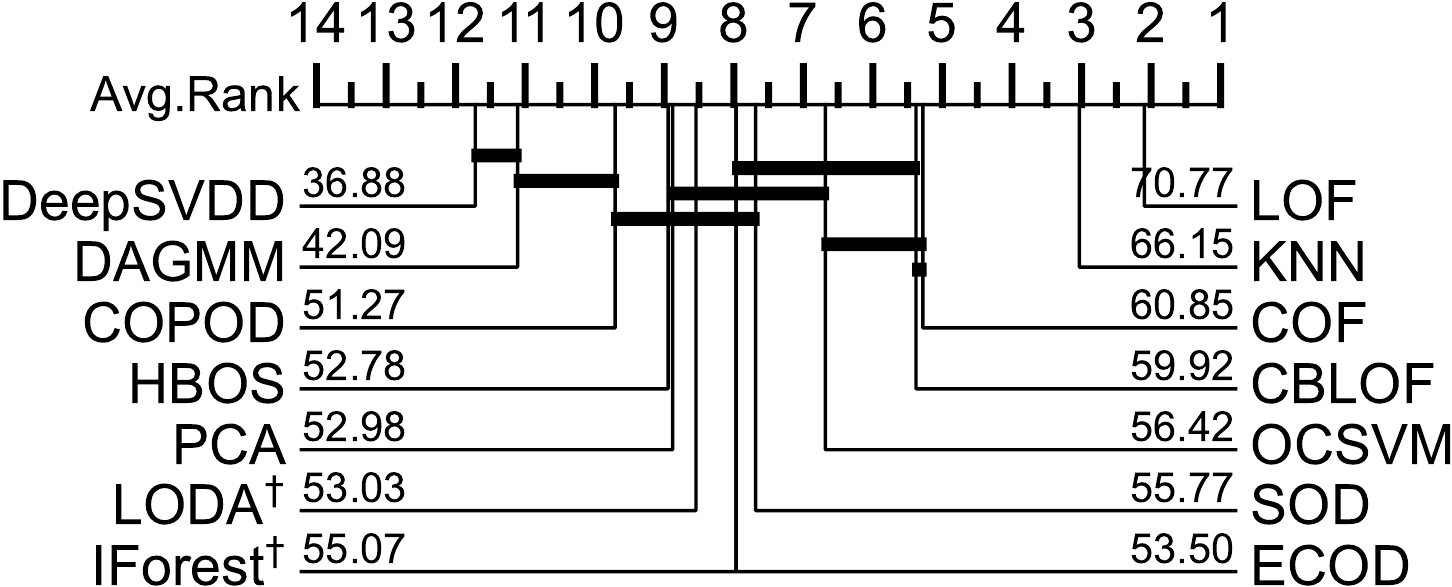}
         \caption{Local anomalies}
     \end{subfigure}
     \hfill
     \begin{subfigure}[b]{0.45\textwidth}
         \centering
         \includegraphics[width=1\textwidth]{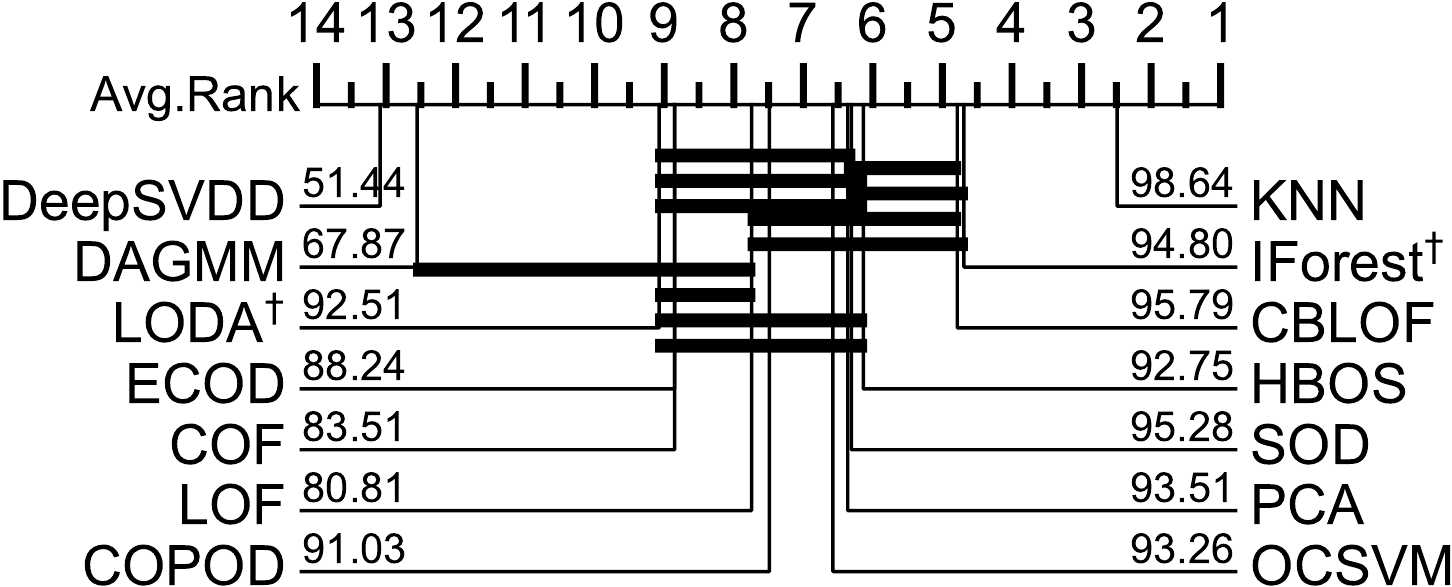}
         \caption{Global anomalies}
     \end{subfigure}
     \hfill
     \begin{subfigure}[b]{0.45\textwidth}
         \centering
         \includegraphics[width=1\textwidth]{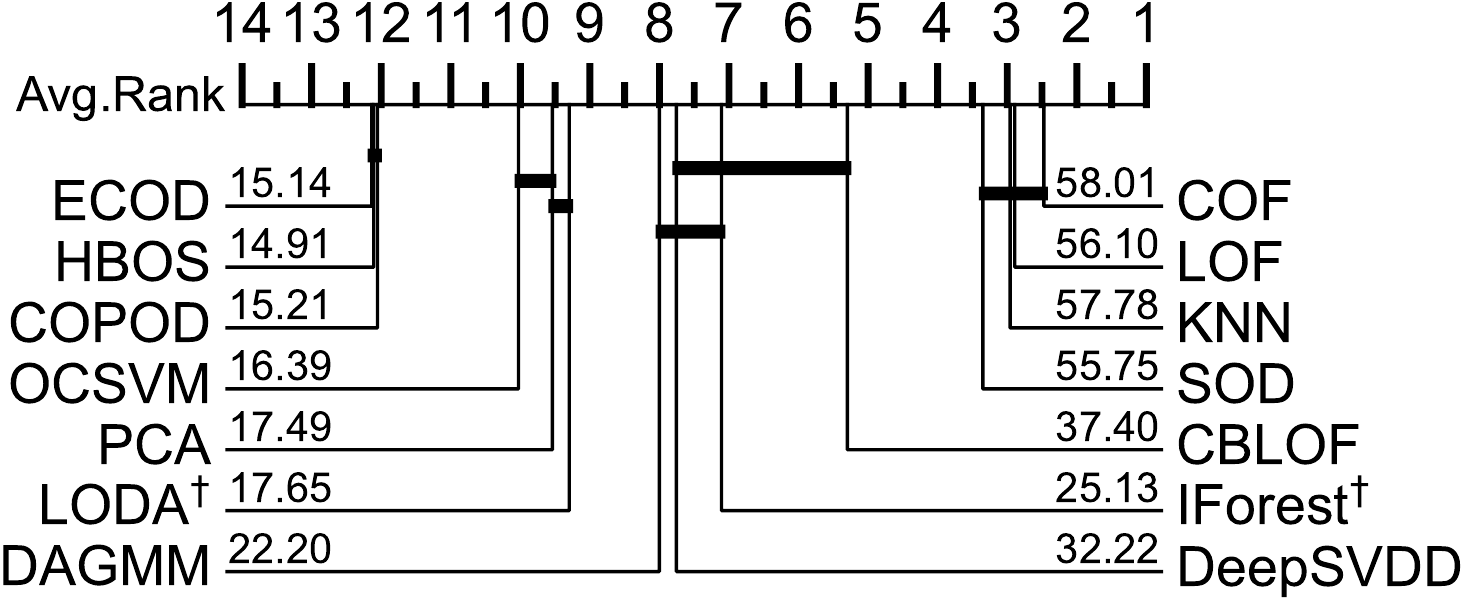}
         \caption{Dependency anomalies}
     \end{subfigure}
     \hfill
     \begin{subfigure}[b]{0.45\textwidth}
         \centering
         \includegraphics[width=1\textwidth]{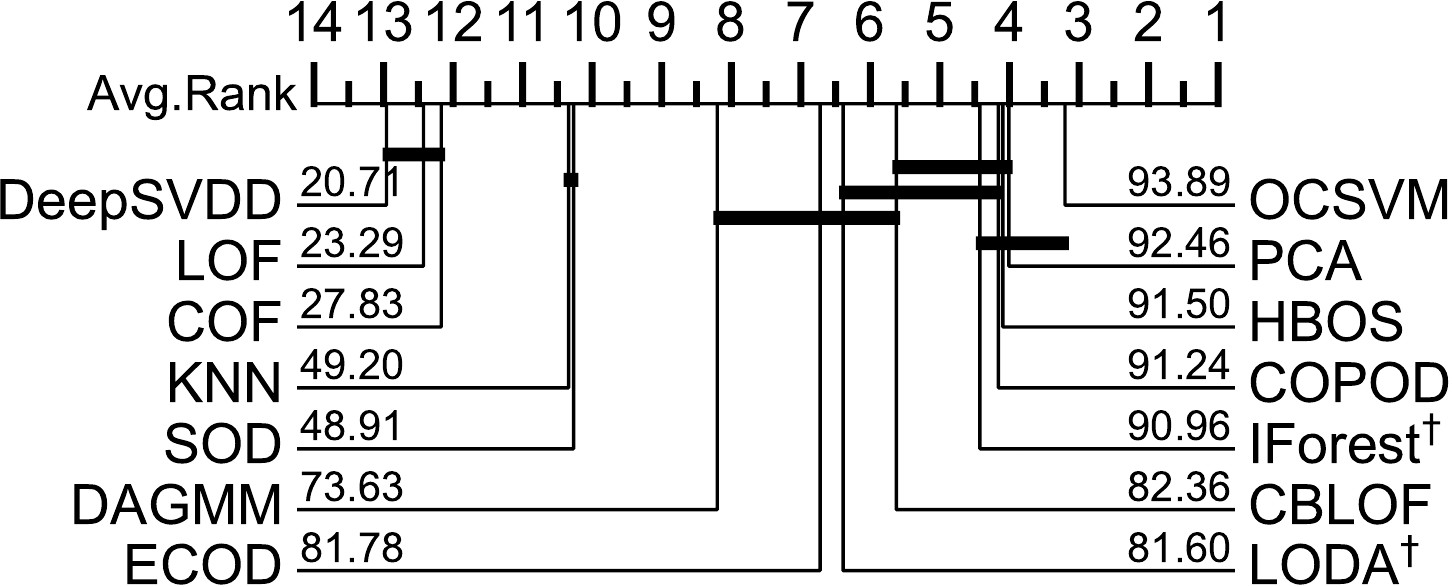}
         \caption{Clustered anomalies}
     \end{subfigure}
     \caption{AUCPR CD Diagram of unsupervised methods on different types of anomalies. The unsupervised methods perform well when their assumptions conform to the anomaly types.}
     \label{fig:type_unsupervised aucpr}
\end{figure}

\begin{figure}[!ht]
     \centering
     \begin{subfigure}[b]{0.48\textwidth}
         \centering
         \includegraphics[width=\textwidth]{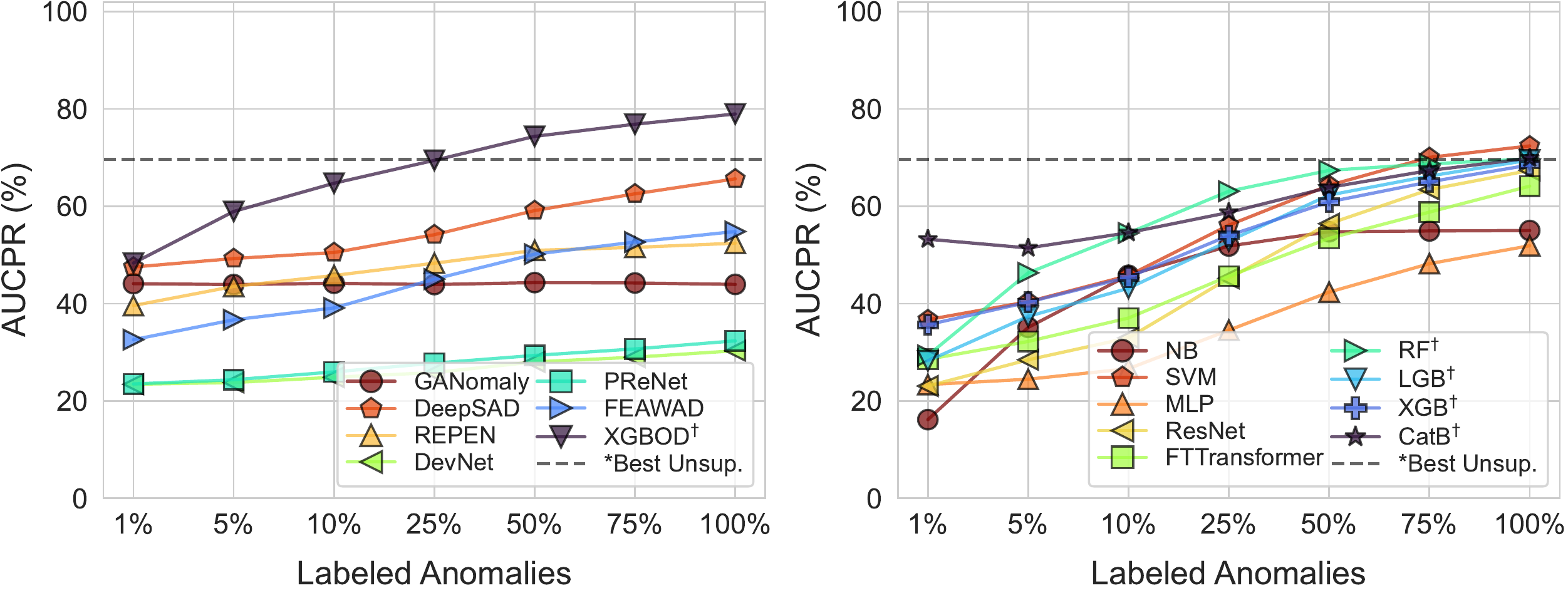}
         \caption{Local anomalies}
     \end{subfigure}
     \hfill
     \begin{subfigure}[b]{0.48\textwidth}
         \centering
         \includegraphics[width=\textwidth]{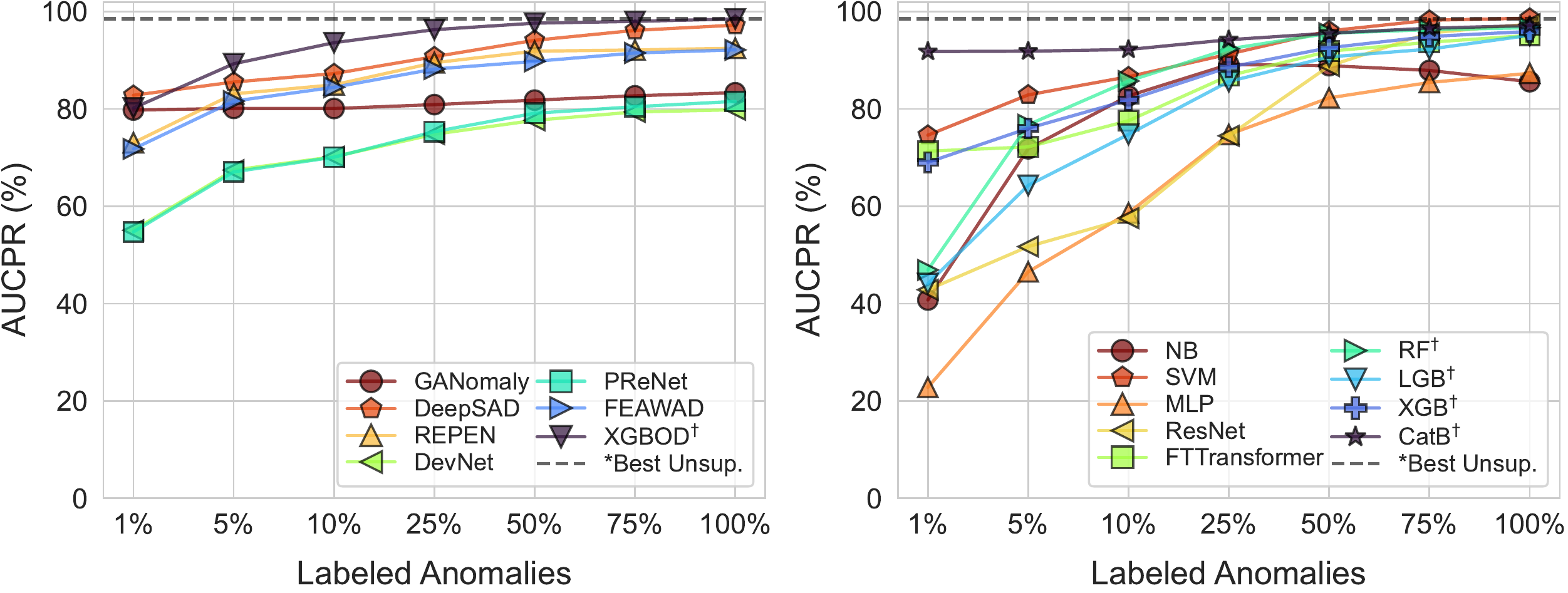}
         \caption{Global anomalies}
     \end{subfigure}
     \hfill
     \begin{subfigure}[b]{0.48\textwidth}
         \centering
         \includegraphics[width=\textwidth]{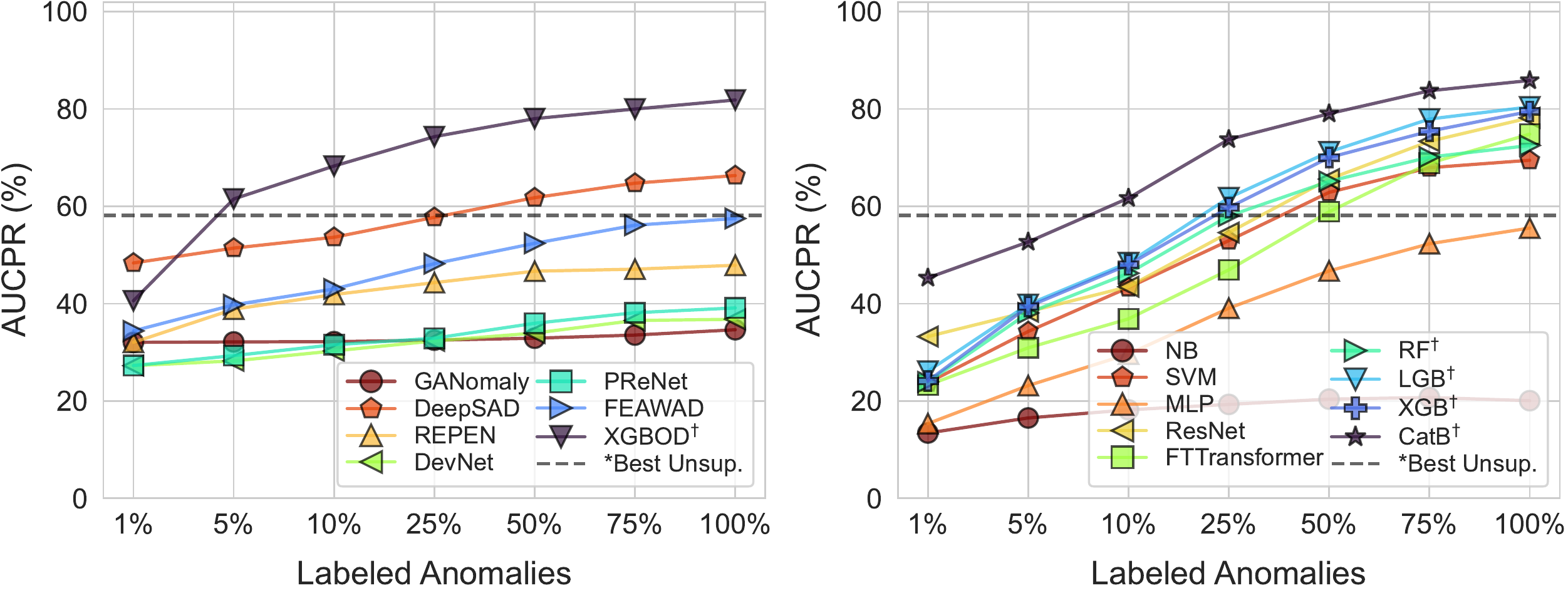}
         \caption{Dependency anomalies}
     \end{subfigure}
     \hfill
     \begin{subfigure}[b]{0.48\textwidth}
         \centering
         \includegraphics[width=\textwidth]{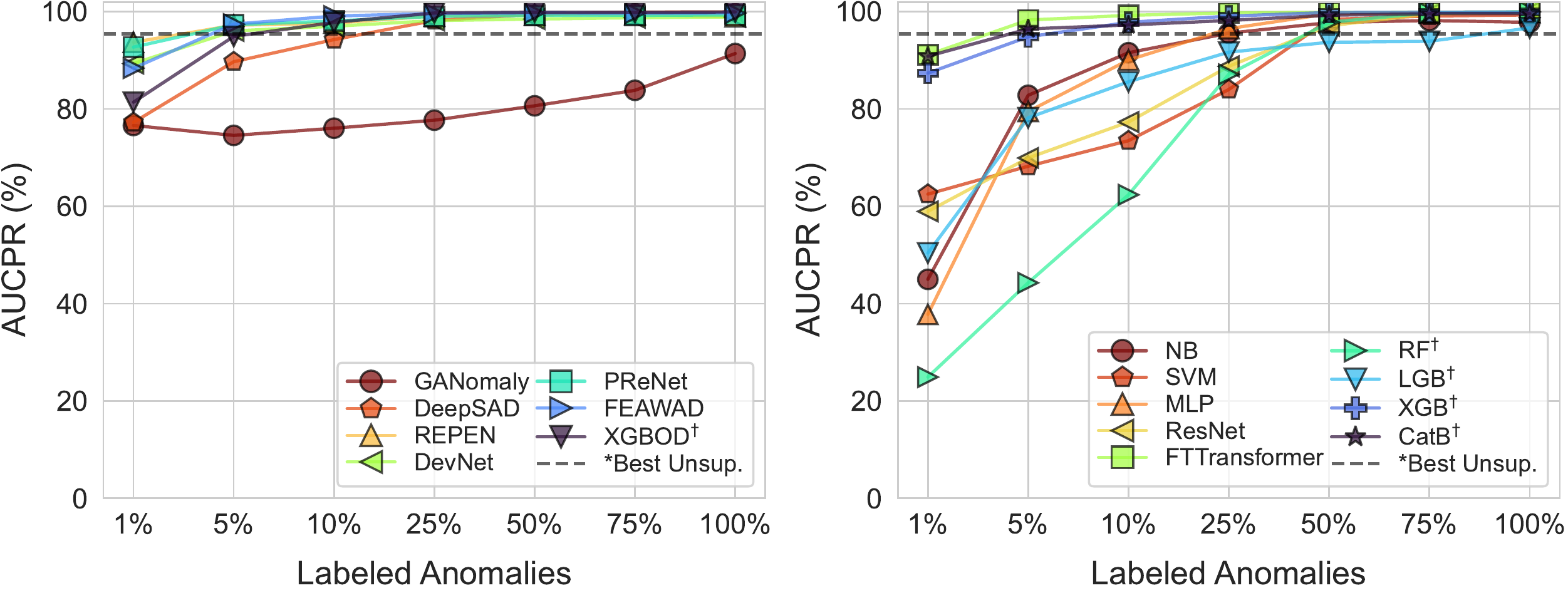}
         \caption{Clustered anomalies}
     \end{subfigure}
     \caption{Semi- (left of each subfigure) and supervised (right) algorithms' performance on different types of anomalies with varying levels of labeled anomalies for AUCPR performance. Surprisingly, these label-informed algorithms are \textit{inferior} to the best unsupervised method except for the clustered anomalies.}
     \label{fig:type_supervised aucpr}
\end{figure}

We additionally show the AUCPR results for model performance on different types of anomalies in Fig.~\ref{fig:type_unsupervised aucpr} and Fig.~\ref{fig:type_supervised aucpr}, which are consistent with the conclusions drawn in \S \ref{exp:types}, i.e., the unsupervised methods are significantly better if their model assumptions conform to the underlying anomaly types. Moreover, the prior knowledge of anomaly types can be more important than that of label information, where those label-informed algorithms generally underperform the best unsupervised methods for local, global, and dependency anomalies.

We want to note that XGBOD can be regarded as an exception to the above observations, which is comparable to or even outperforms the best unsupervised model when more labeled anomalies are available. Recall that XGBOD employs the stacking ensemble method \cite{wolpert1992stacked}, where heterogeneous unsupervised methods are integrated with the supervised model XGBoost, therefore XGBOD is more adaptable to different data assumptions while effectively leveraging the label information. This validates the conclusion that such ensemble learning techniques should be considered in future research directions.

\clearpage
\subsection{Additional Results for Algorithm Robustness in \S \ref{exp:robustness}}

\begin{figure}[h!]
     \centering
     \begin{subfigure}[b]{0.245\textwidth}
         \centering
         \includegraphics[width=\textwidth]{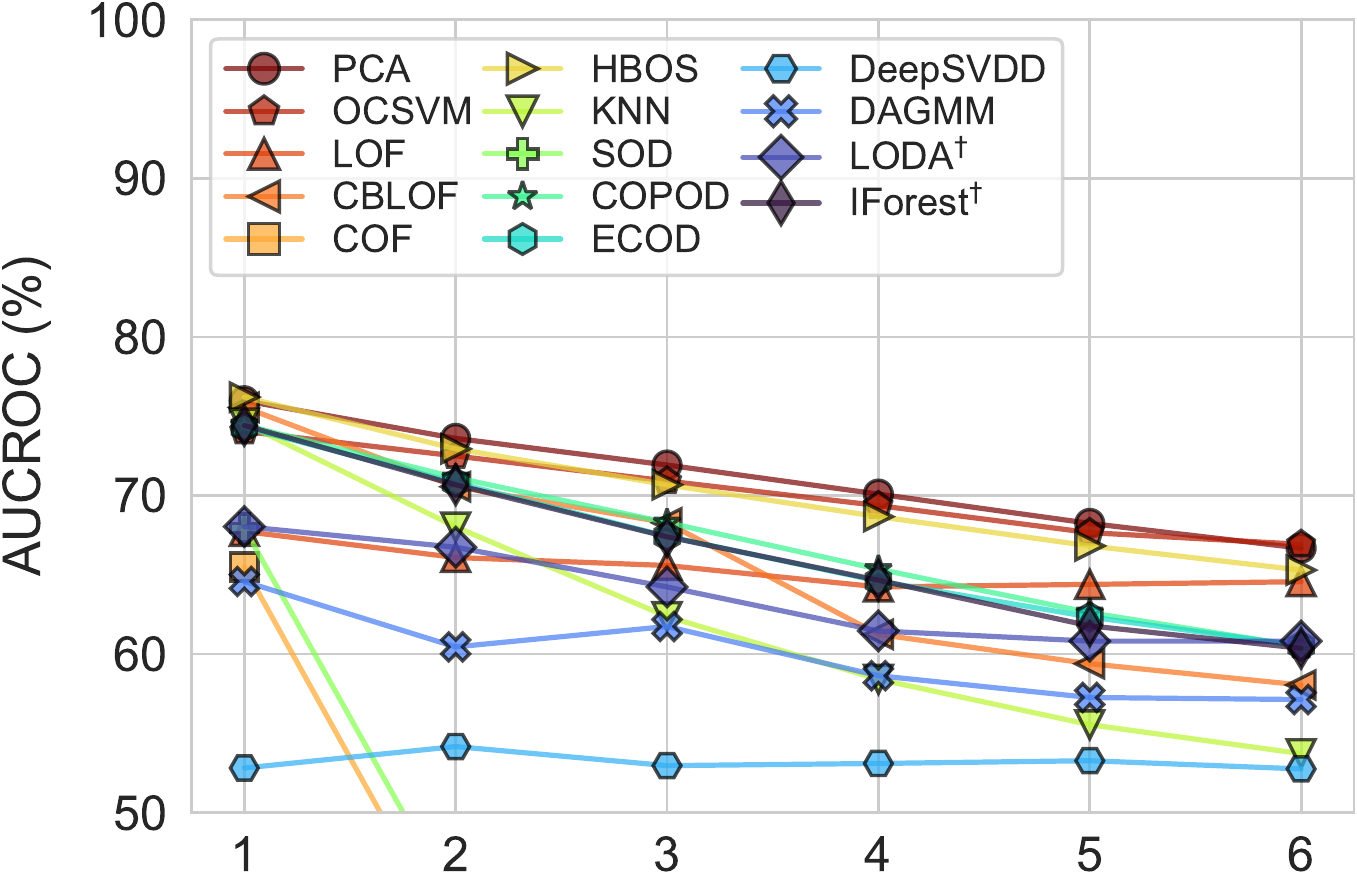}
         \caption{\scriptsize Duplicated Anomalies, unsup}
         \label{fig:dp_unsup_aucroc}
     \end{subfigure}
     \begin{subfigure}[b]{0.245\textwidth}
         \centering
         \includegraphics[width=\textwidth]{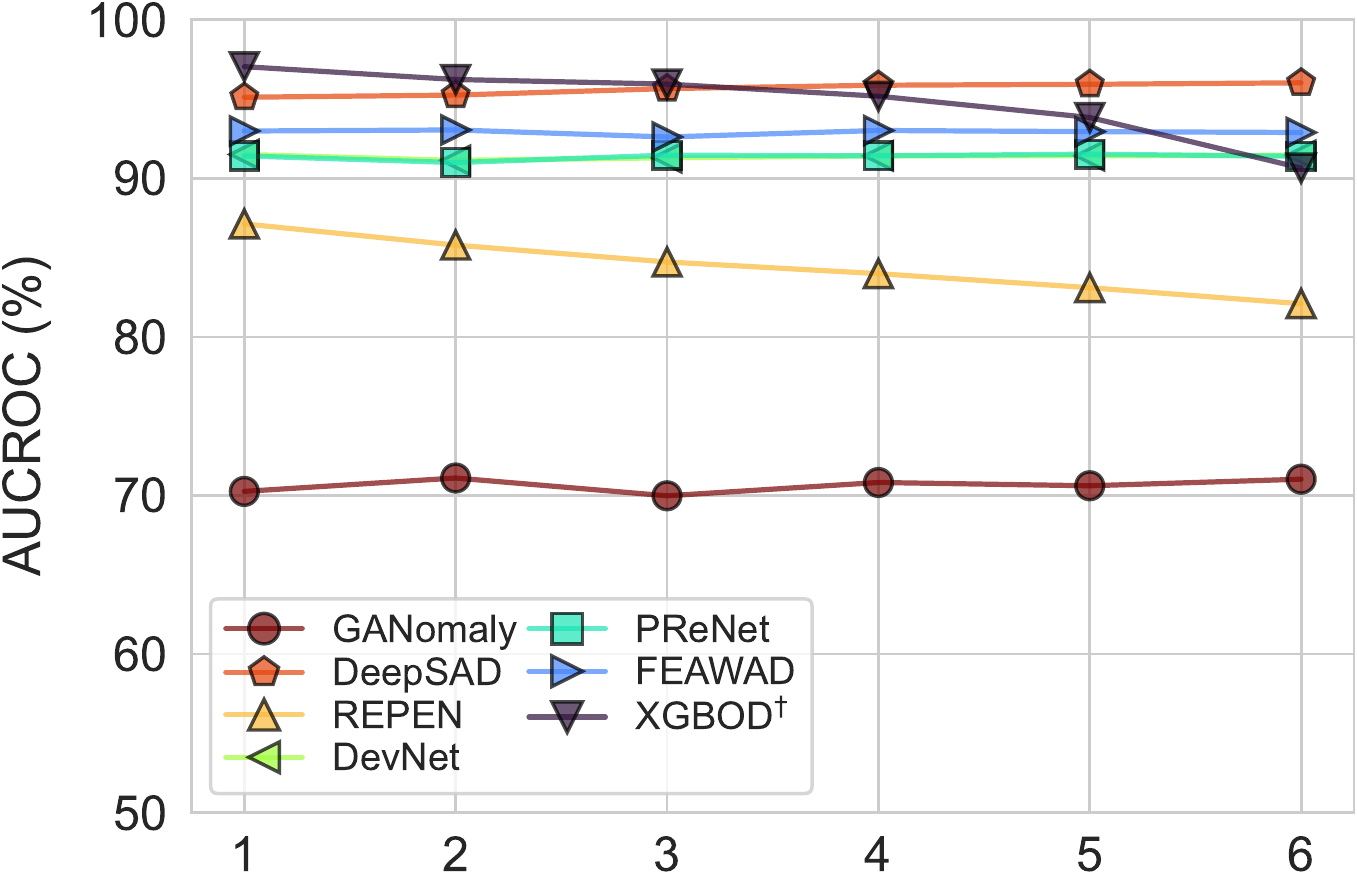}
         \caption{\scriptsize Duplicated Anomalies, semi}
         \label{fig:dp_semi_aucroc}
     \end{subfigure}
     \begin{subfigure}[b]{0.245\textwidth}
         \centering
         \includegraphics[width=\textwidth]{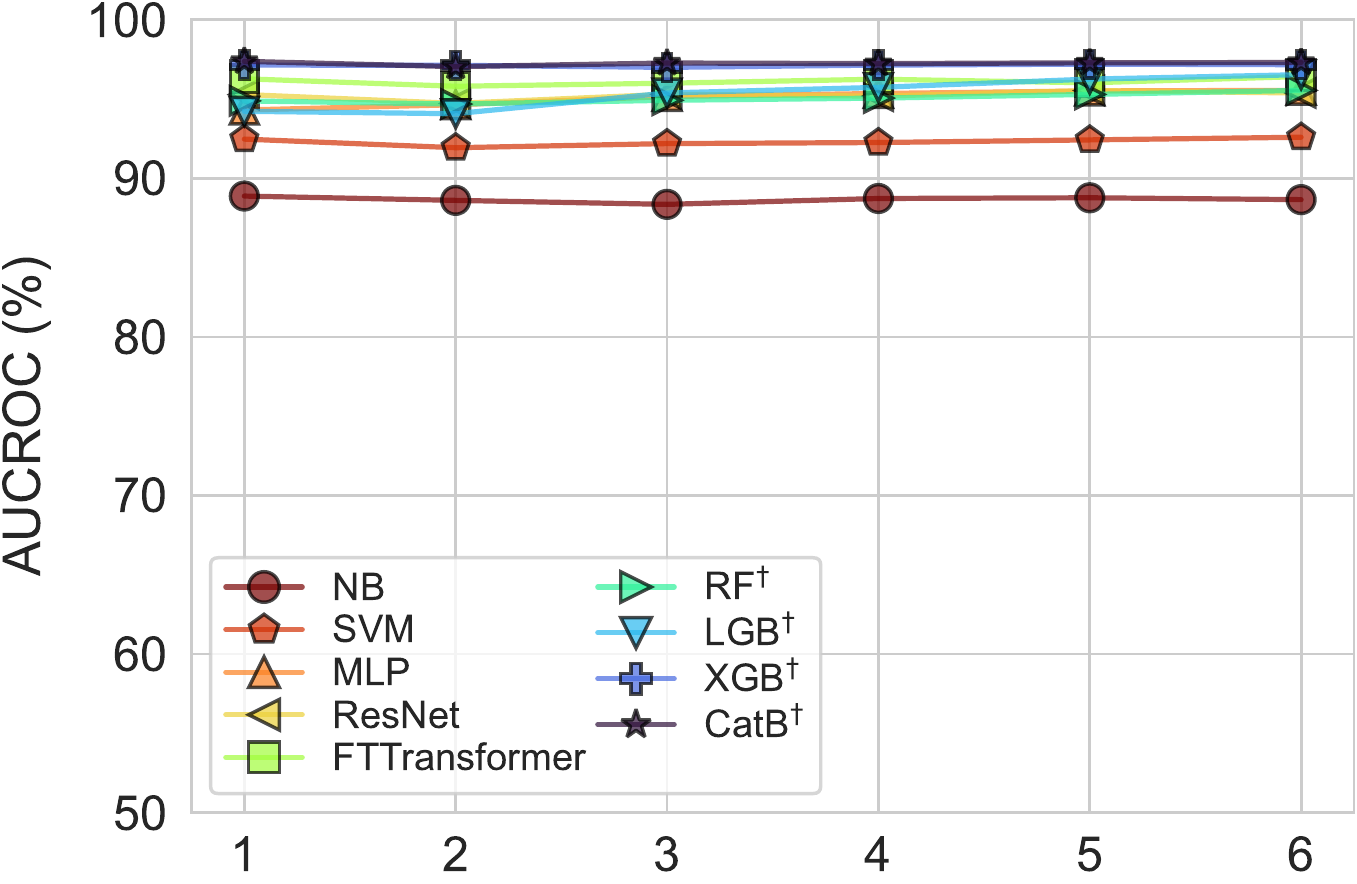}
         \caption{\scriptsize Duplicated Anomalies, sup}
         \label{fig:dp_sup_roc}
     \end{subfigure}
     \begin{subfigure}[b]{0.245\textwidth}
         \centering
         \includegraphics[width=\textwidth]{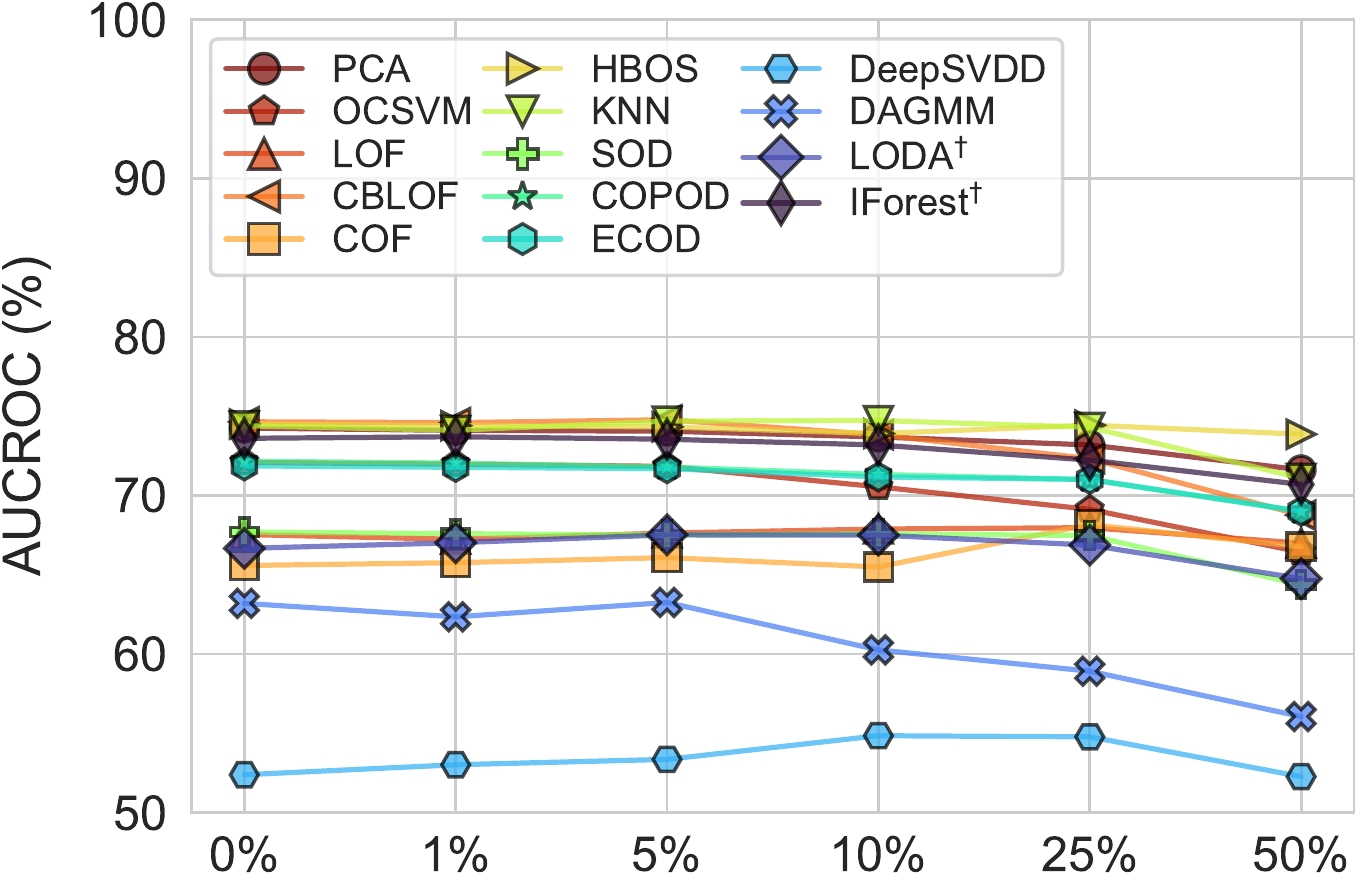}
         \caption{\scriptsize Irrelevant Features, unsup}
         \label{fig:ir_unsup_auc_roc}
     \end{subfigure}
     \begin{subfigure}[b]{0.245\textwidth}
         \centering
         \includegraphics[width=\textwidth]{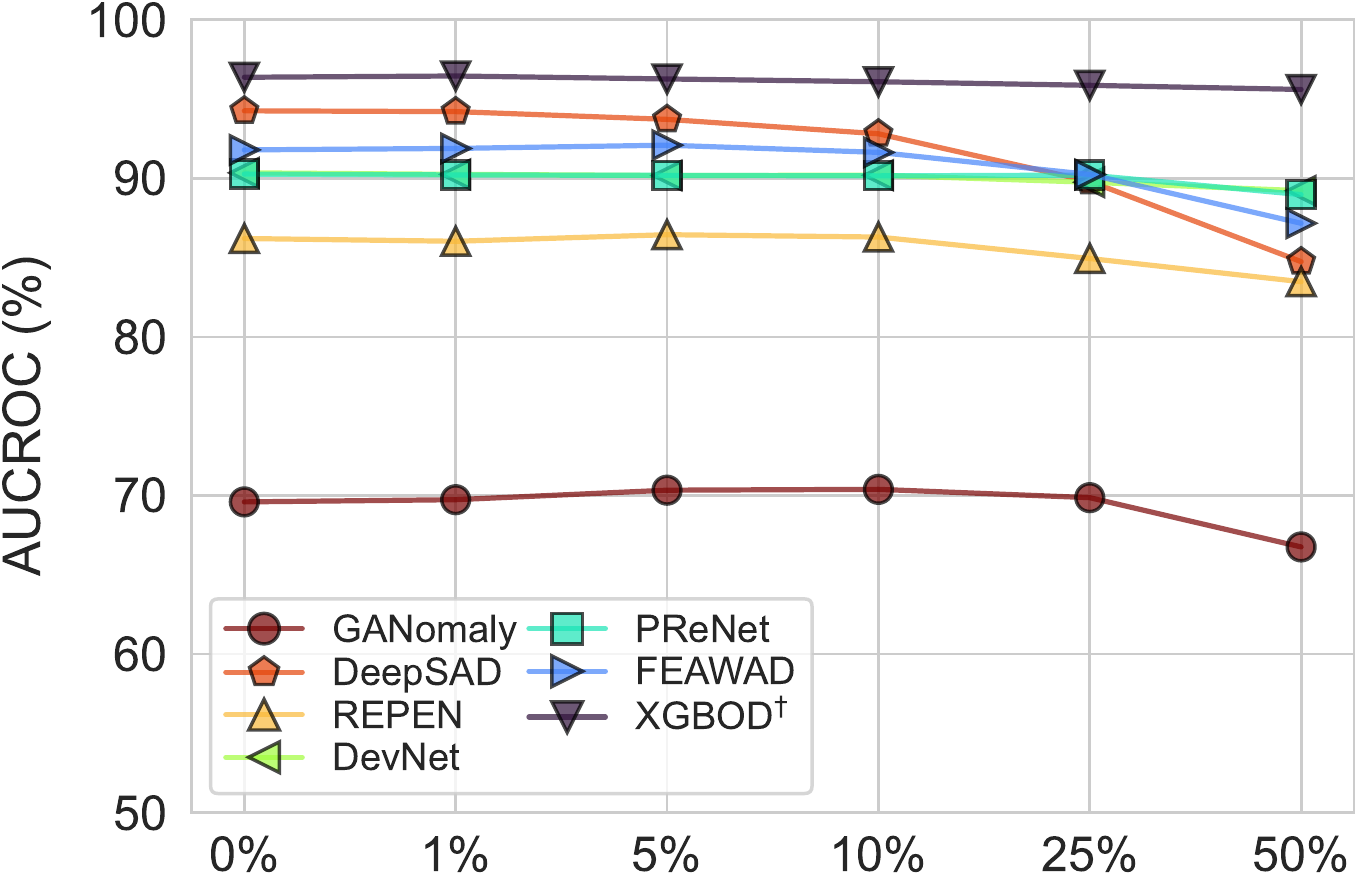}
         \caption{\scriptsize Irrelevant Features, semi}
         \label{fig:ir_semi_roc}
     \end{subfigure}
     \begin{subfigure}[b]{0.245\textwidth}
         \centering
         \includegraphics[width=\textwidth]{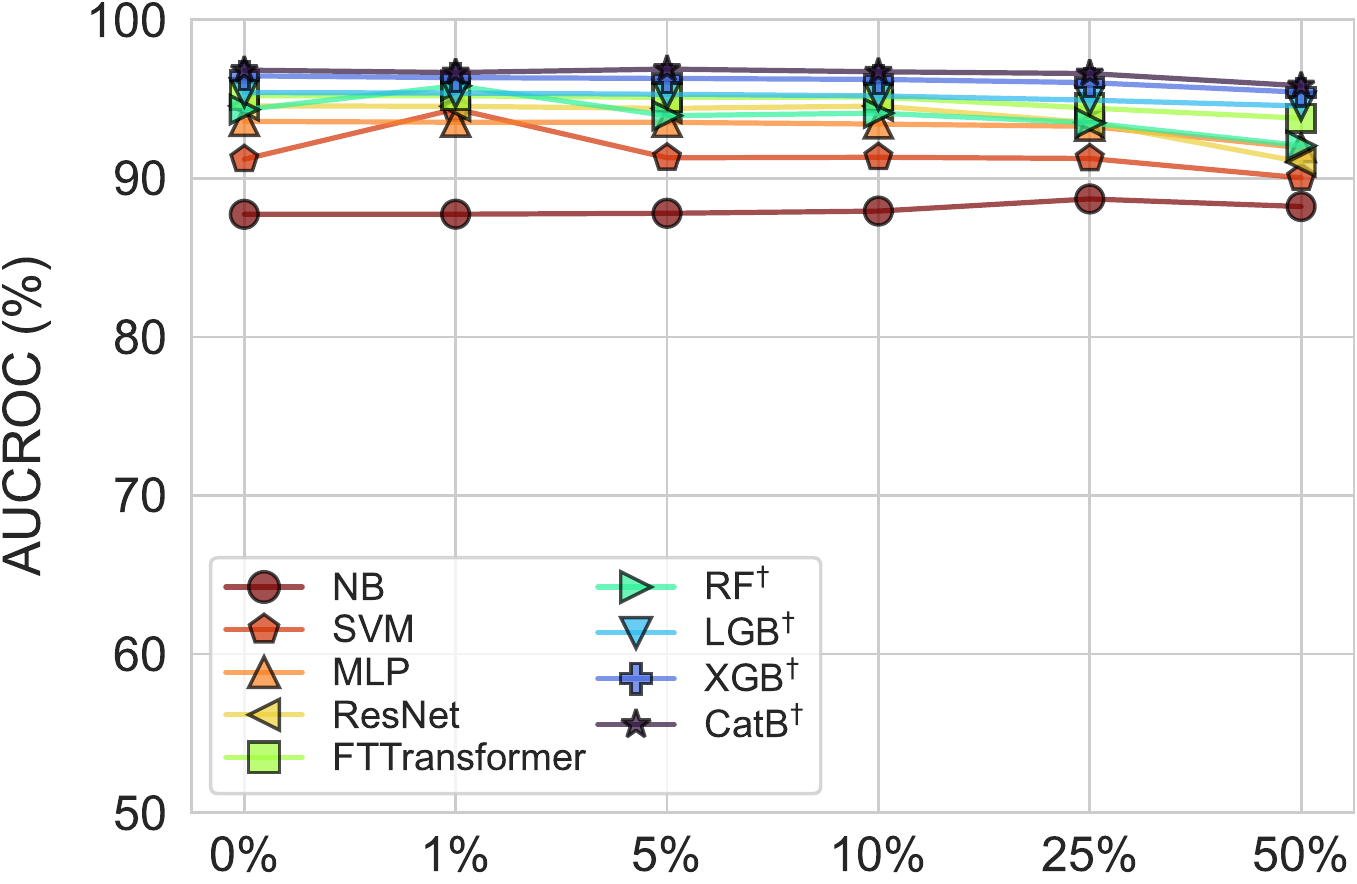}
         \caption{\scriptsize Irrelevant Features, sup}
         \label{fig:ir_sup_roc}
     \end{subfigure}
     \begin{subfigure}[b]{0.245\textwidth}
         \centering
         \includegraphics[width=\textwidth]{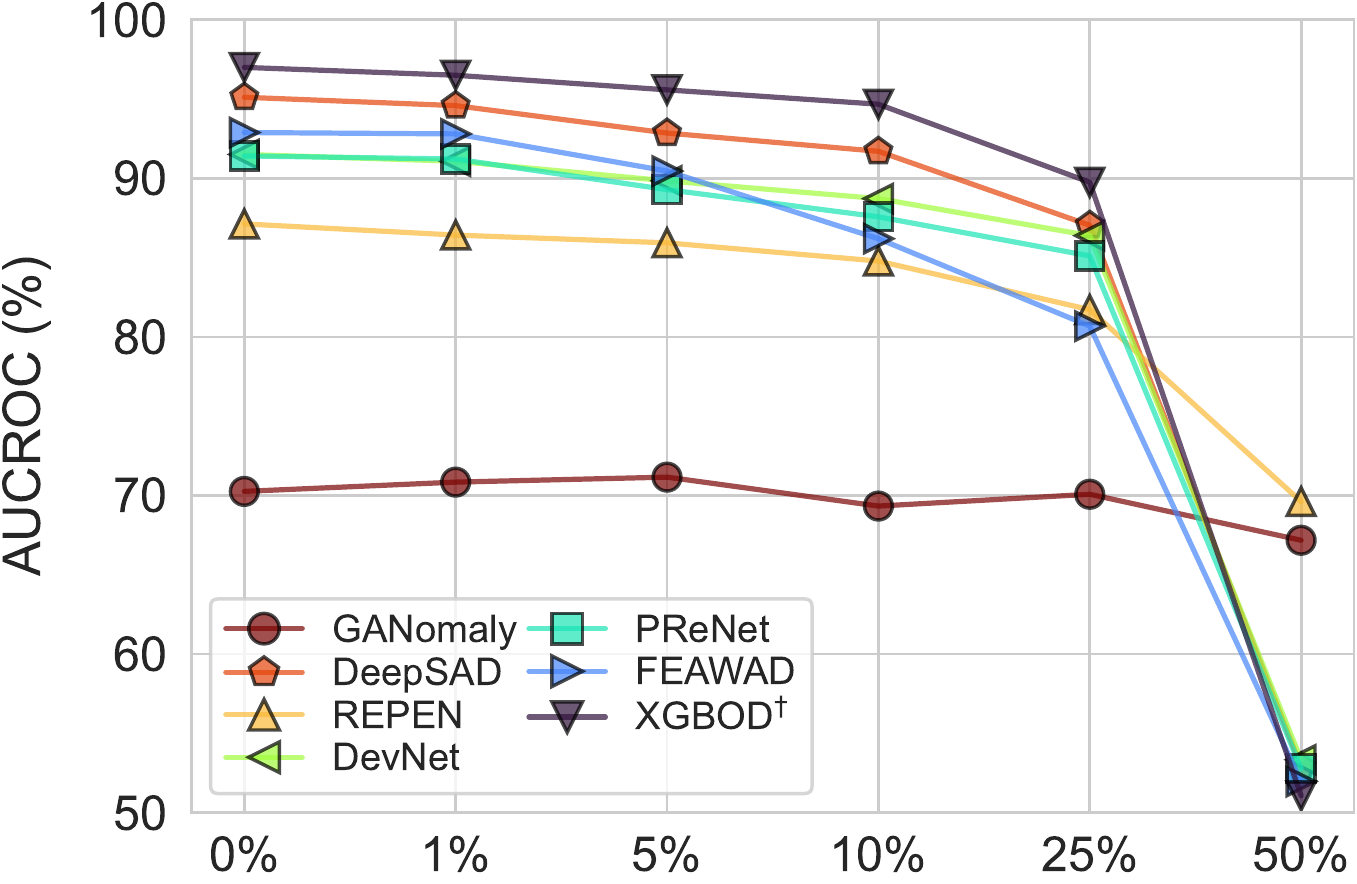}
         \caption{\scriptsize Annotation Errors, semi}
         \label{fig:ae_semi_roc}
     \end{subfigure}
     \begin{subfigure}[b]{0.245\textwidth}
         \centering
         \includegraphics[width=\textwidth]{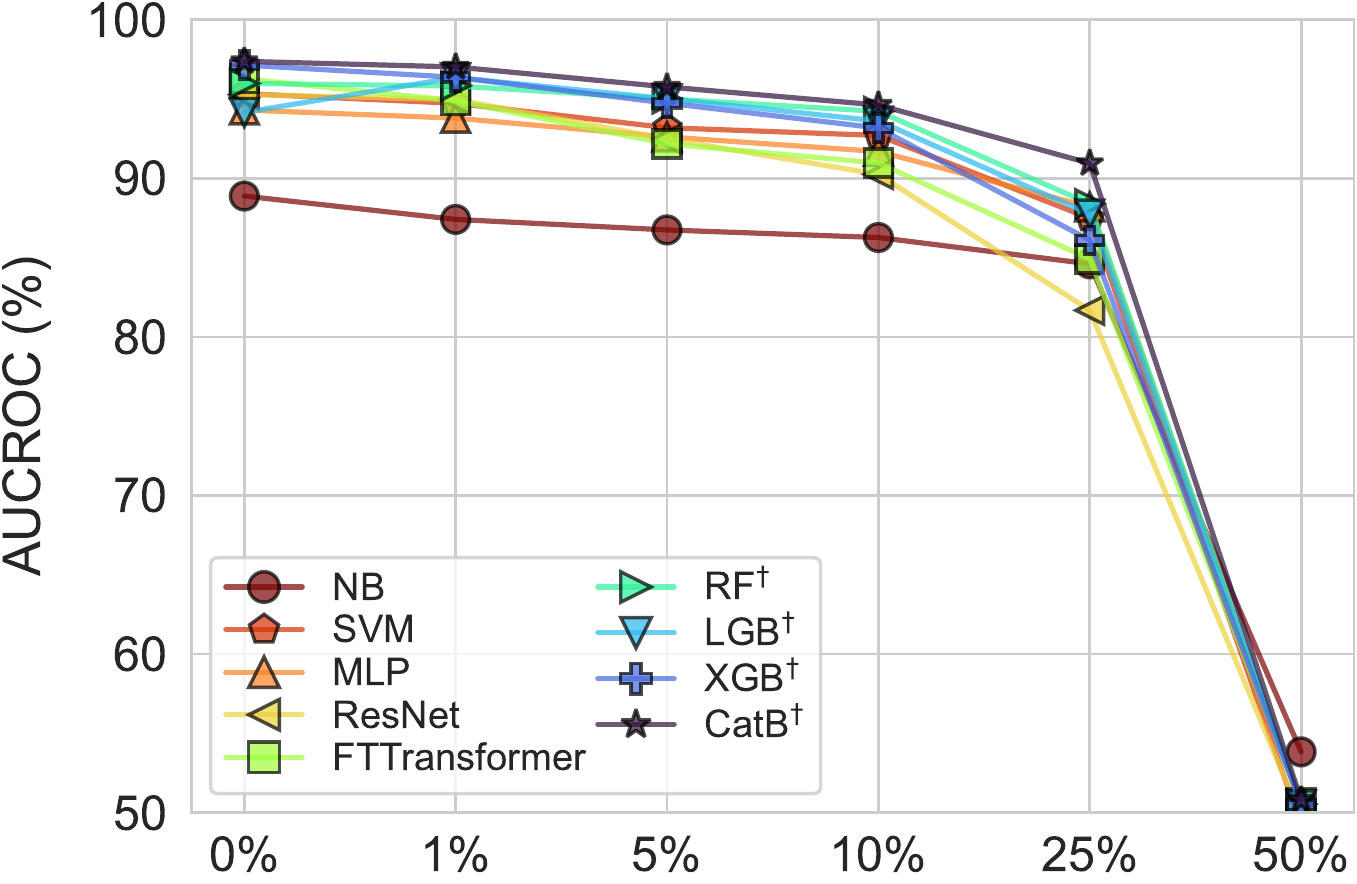}
         \caption{\scriptsize Annotation Errors, sup}
         \label{fig:ae_sup_aucroc}
     \end{subfigure}
     \vspace{-0.15in}
    \caption{Algorithm performance under noisy and corrupted data (i.e., duplicated anomalies for (a)-(c), irrelevant features for (d)-(f), and annotation errors for (g) and (h)). X-axis denotes either the duplicated times or the noise ratio. Y-axis denotes the $\text{AUCROC}$ performance and its range remains consistent across different algorithms. The results reveal unsupervised methods' susceptibility to duplicated anomalies and the usage of label information in defending irrelevant features. Un-, semi-, and fully-supervised methods are denoted as \textit{unsup}, \textit{semi}, and \textit{sup}, respectively. The results are mostly consistent with the observations in Fig. \ref{fig:model_robustness} (\S \ref{exp:robustness}) showing the relative performance change.
    }
    \vspace{-0.1in}
    \label{fig:robustness:absolute}
\end{figure}

\normalsize 
In Fig. \ref{fig:robustness:absolute}, we provide the performance of the AD algorithms under noisy and corrupted data. Along with the relative performance changes shown in Fig. \ref{fig:model_robustness}, the analysis in \ref{exp:robustness} still stands.

\begin{figure}[h!]
     \centering
     \begin{subfigure}[b]{0.245\textwidth}
         \centering
         \includegraphics[width=\textwidth]{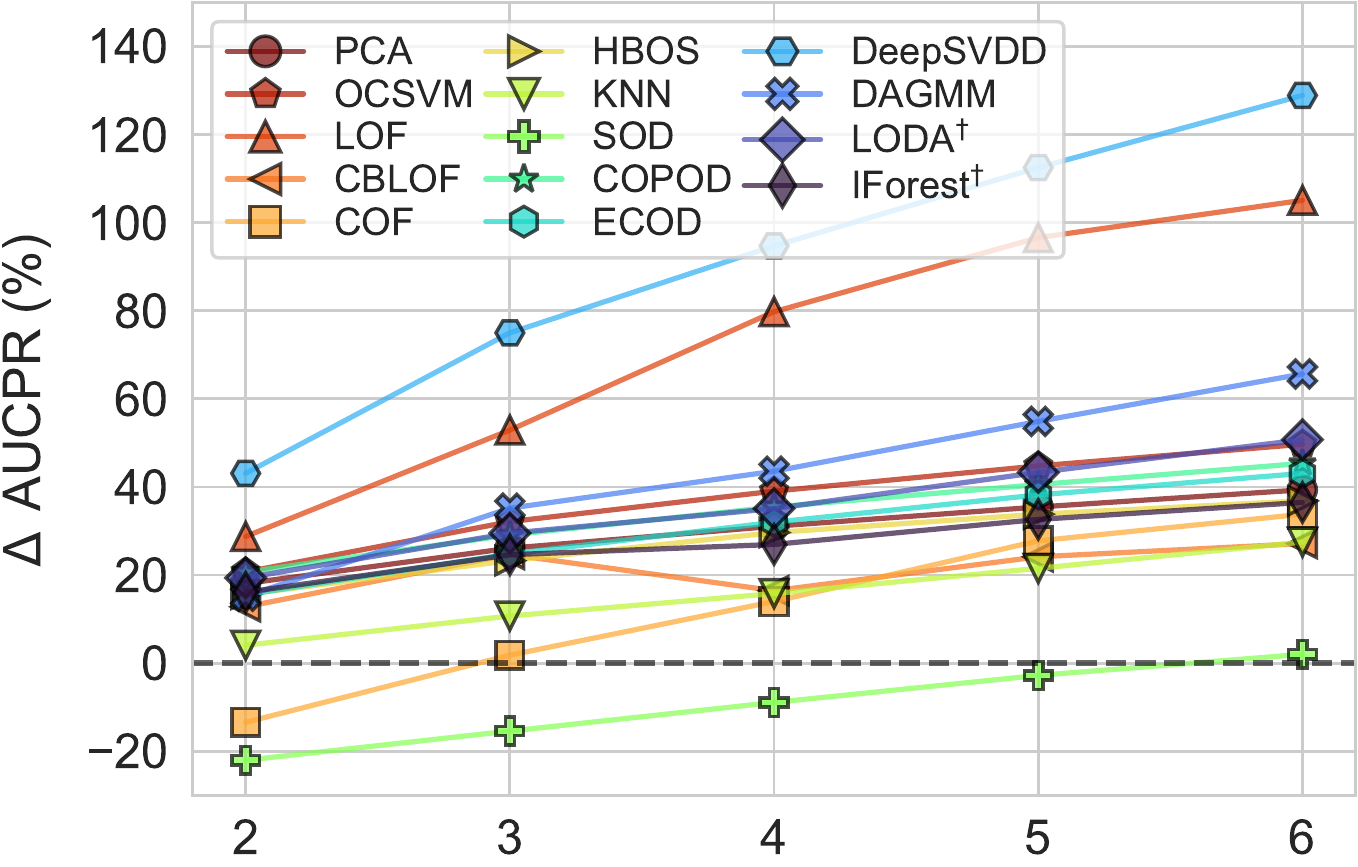}
         \caption{\scriptsize Duplicated Anomalies, unsup}
     \end{subfigure}
     \begin{subfigure}[b]{0.245\textwidth}
         \centering
         \includegraphics[width=\textwidth]{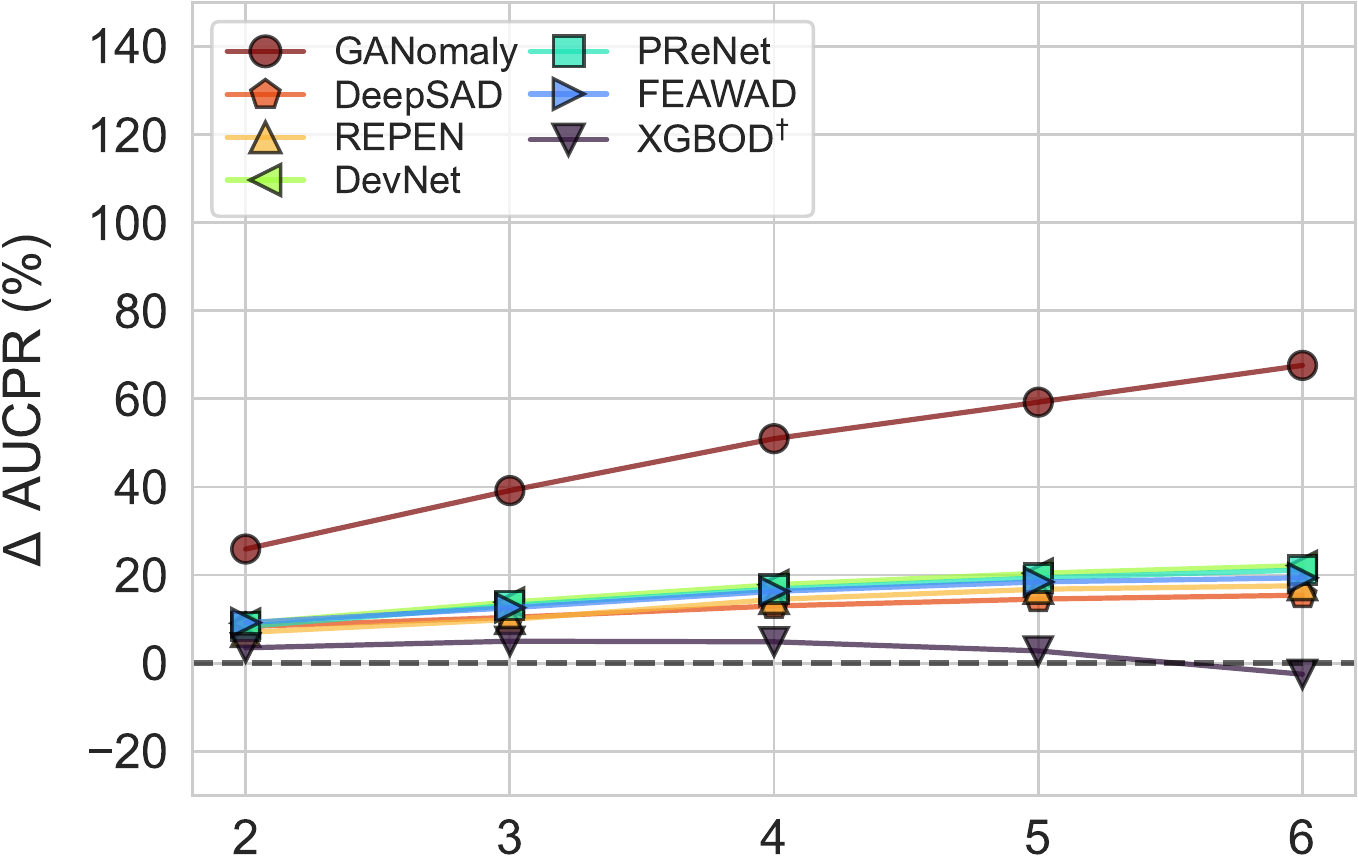}
         \caption{\scriptsize Duplicated Anomalies, semi}
     \end{subfigure}
     \begin{subfigure}[b]{0.245\textwidth}
         \centering
         \includegraphics[width=\textwidth]{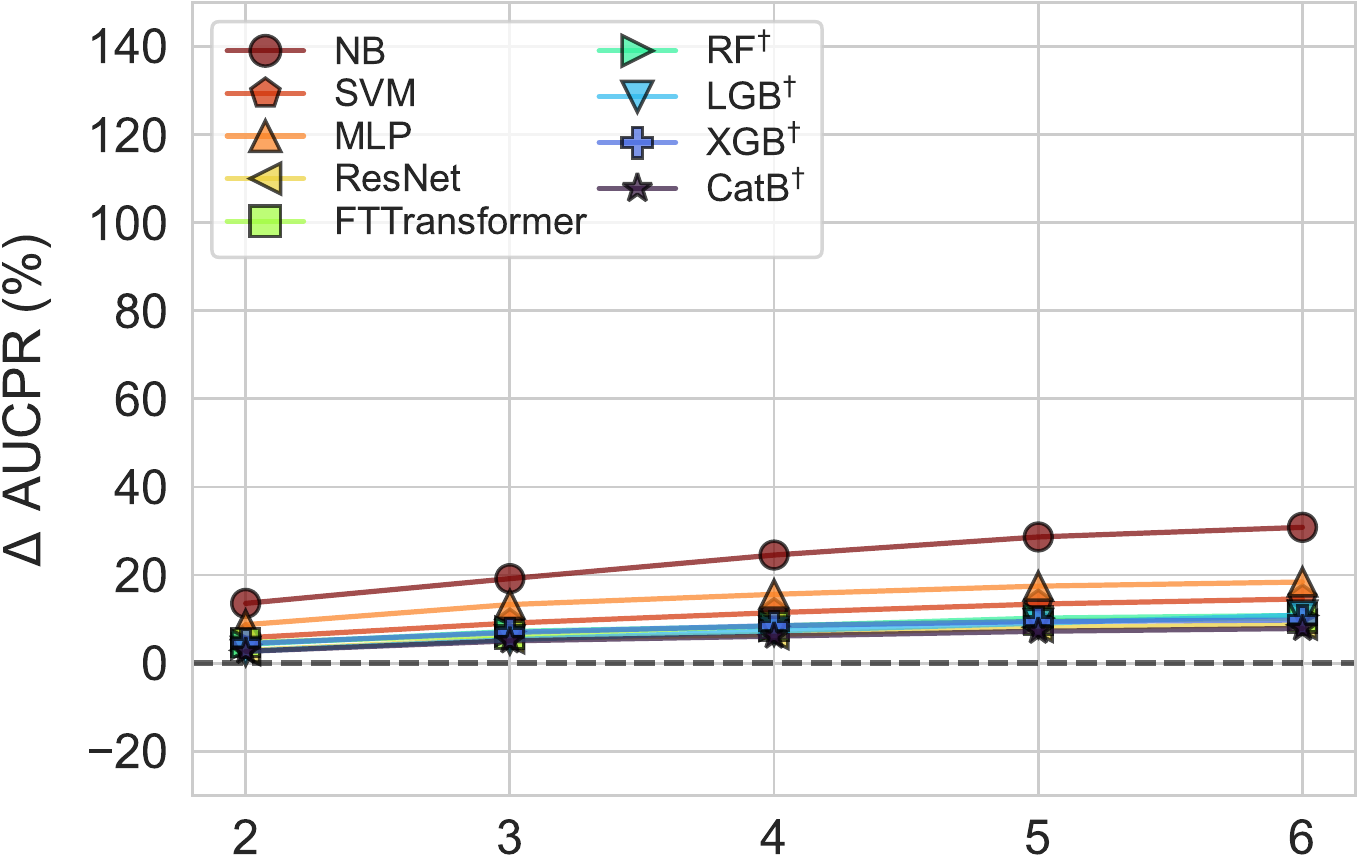}
         \caption{\scriptsize Duplicated Anomalies, sup}
     \end{subfigure}
     \vspace{0.1in}
     \begin{subfigure}[b]{0.245\textwidth}
         \centering
         \includegraphics[width=\textwidth]{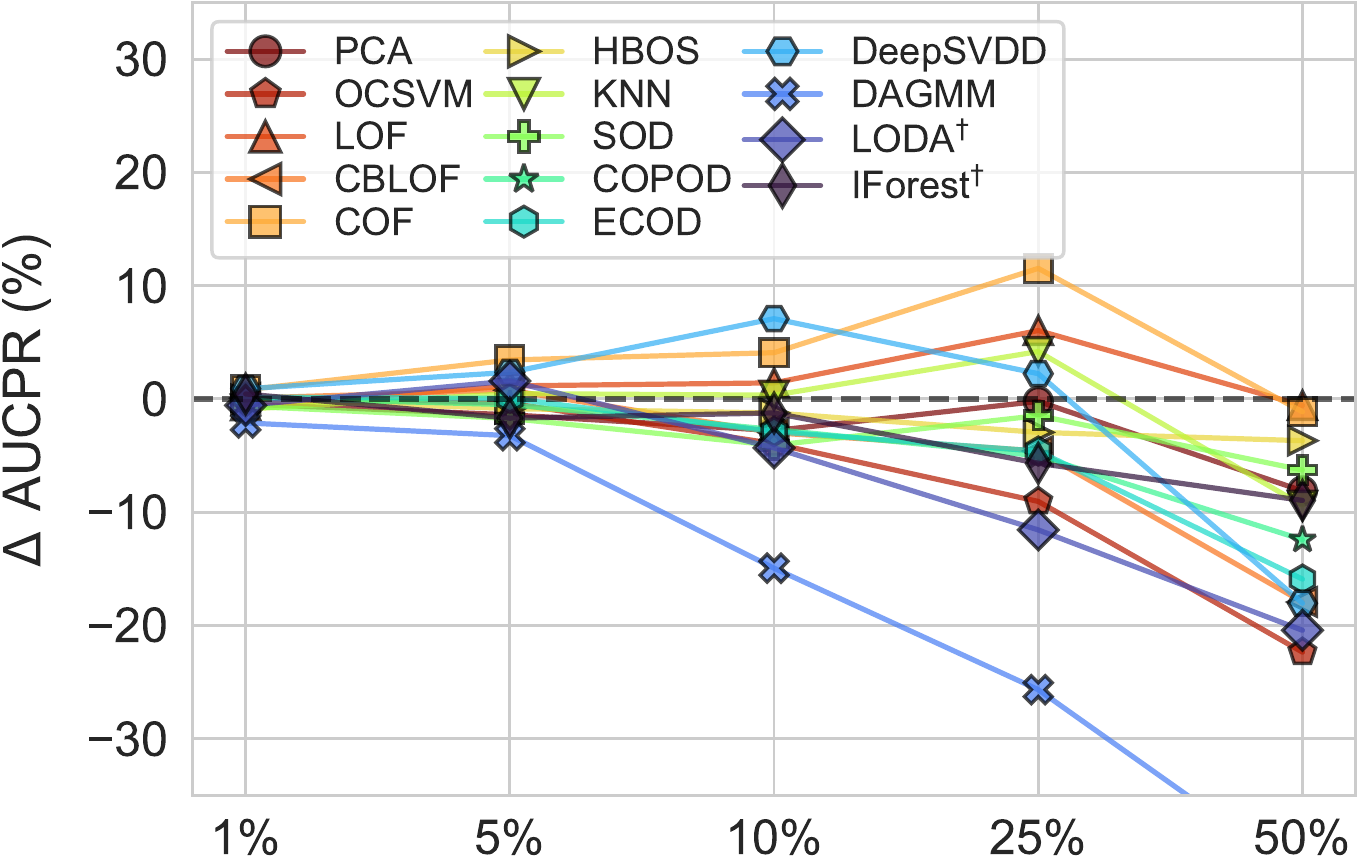}
         \caption{\scriptsize Irrelevant Features, unsup}
     \end{subfigure}
     \begin{subfigure}[b]{0.245\textwidth}
         \centering
         \includegraphics[width=\textwidth]{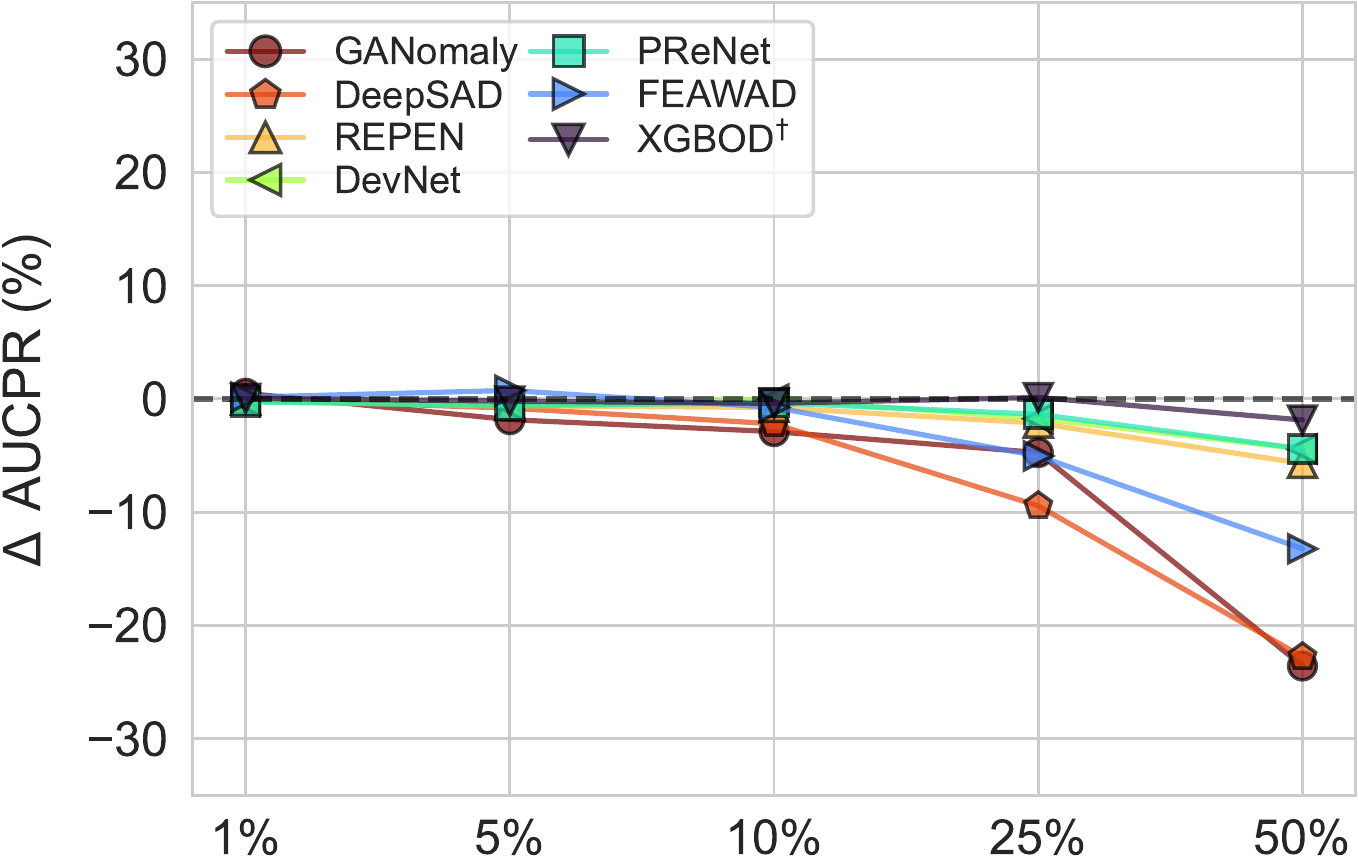}
         \caption{\scriptsize Irrelevant Features, semi}
     \end{subfigure}
     \begin{subfigure}[b]{0.245\textwidth}
         \centering
         \includegraphics[width=\textwidth]{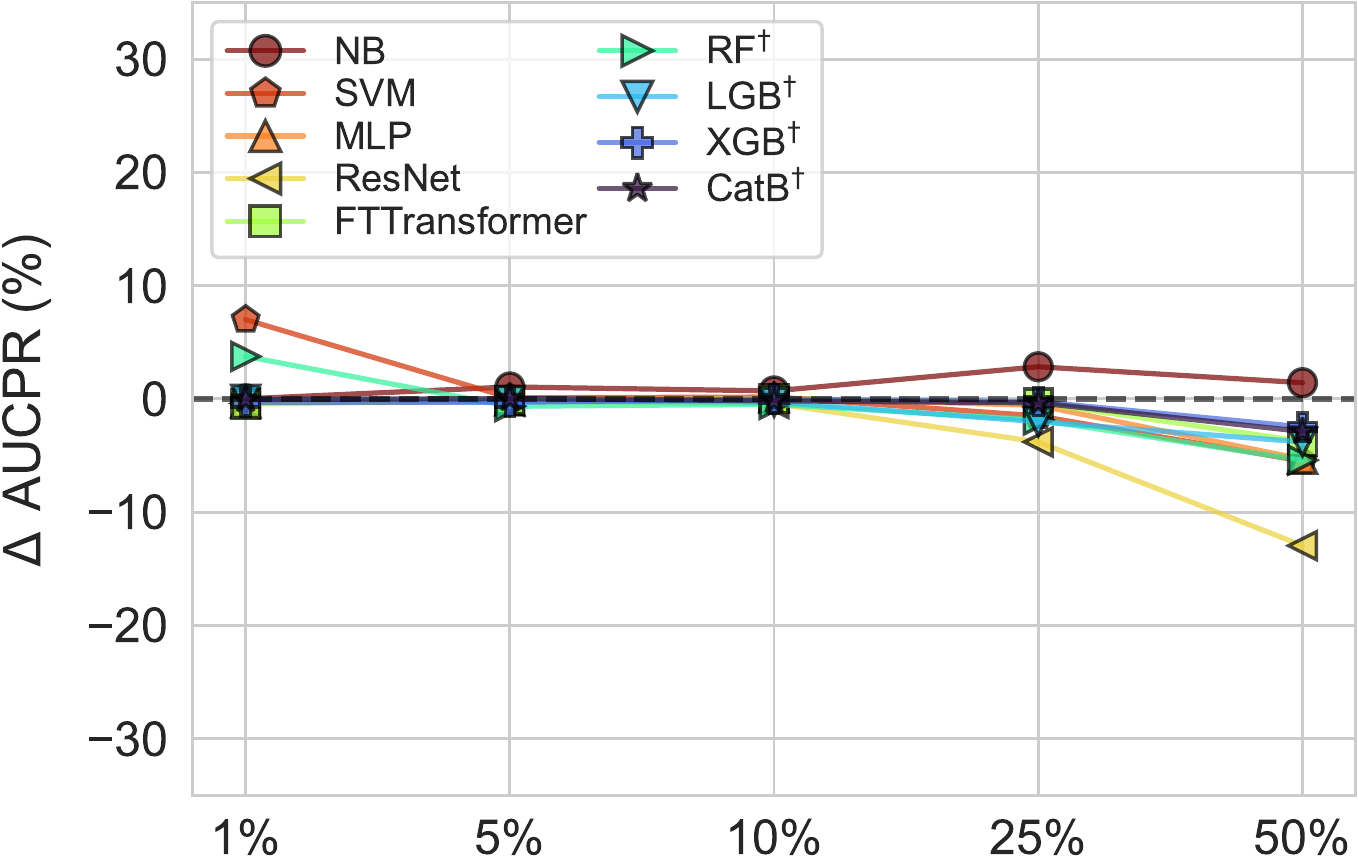}
         \caption{\scriptsize Irrelevant Features, sup}
     \end{subfigure}
          \vspace{0.1in}
     \begin{subfigure}[b]{0.245\textwidth}
         \centering
         \includegraphics[width=\textwidth]{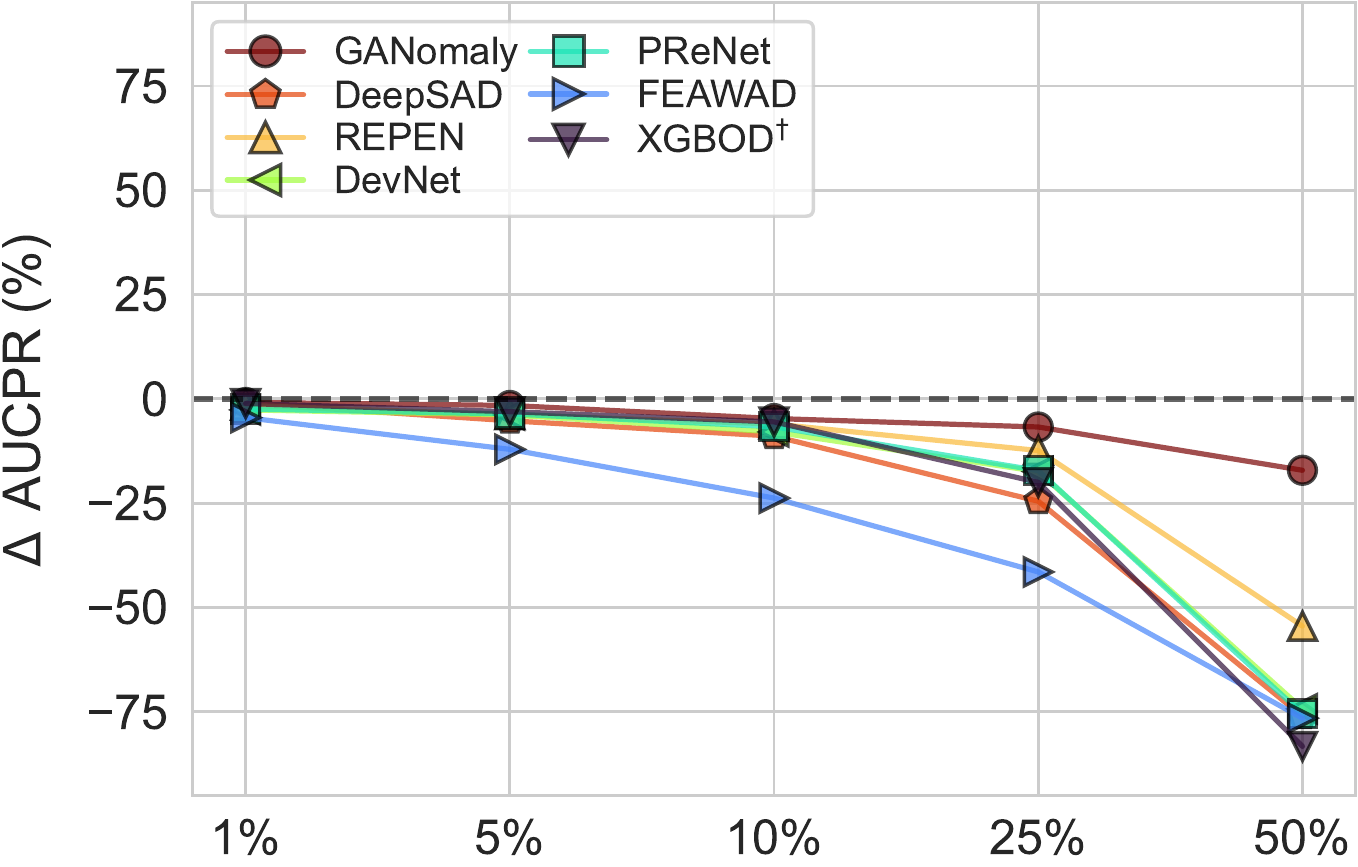}
         \caption{\scriptsize Annotation Errors, semi}
     \end{subfigure}
     \begin{subfigure}[b]{0.245\textwidth}
         \centering
         \includegraphics[width=\textwidth]{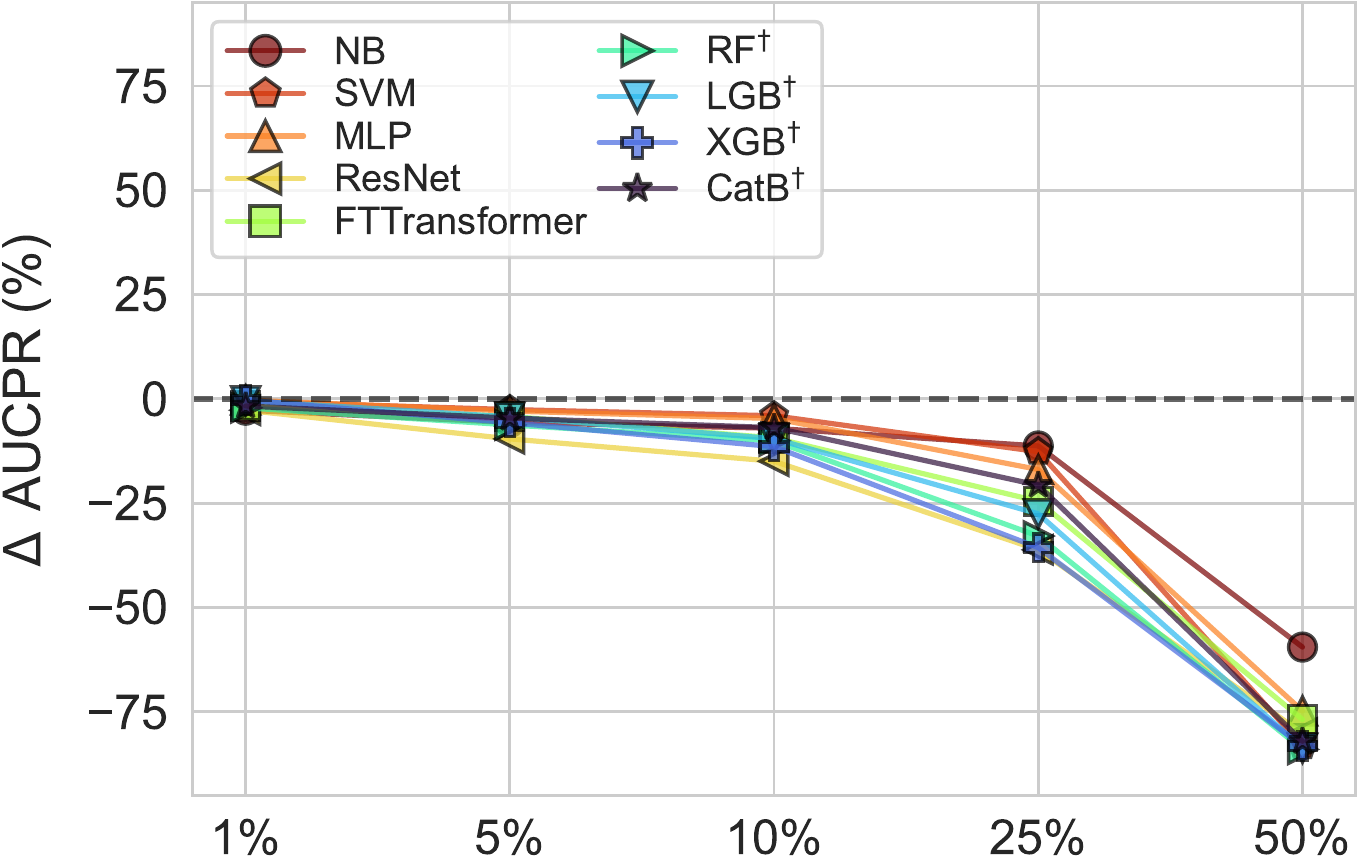}
         \caption{\scriptsize Annotation Errors, sup}
     \end{subfigure}
    \caption{Algorithm performance change under noisy and corrupted data (i.e., duplicated anomalies for (a)-(c), irrelevant features for (d)-(f), and annotation errors for (g) and (h)). y-axis denotes the \% of performance change ($\Delta \text{AUCPR}$) and its range remains consistent across different algorithms. The results reveal the usage of label information in defending irrelevant features, and the robustness of label-informed methods to the minor annotation errors. Un-, semi-, and fully-supervised methods are denoted as \textit{unsup}, \textit{semi}, and \textit{sup}, respectively. The results are mostly consistent with the observations in Fig. \ref{fig:model_robustness} (\S \ref{exp:robustness}) showing the AUCROC.
    }
    \label{fig:model robustness aucpr}
\end{figure}

\begin{figure}[h!]
     \centering
     \begin{subfigure}[b]{0.245\textwidth}
         \centering
         \includegraphics[width=\textwidth]{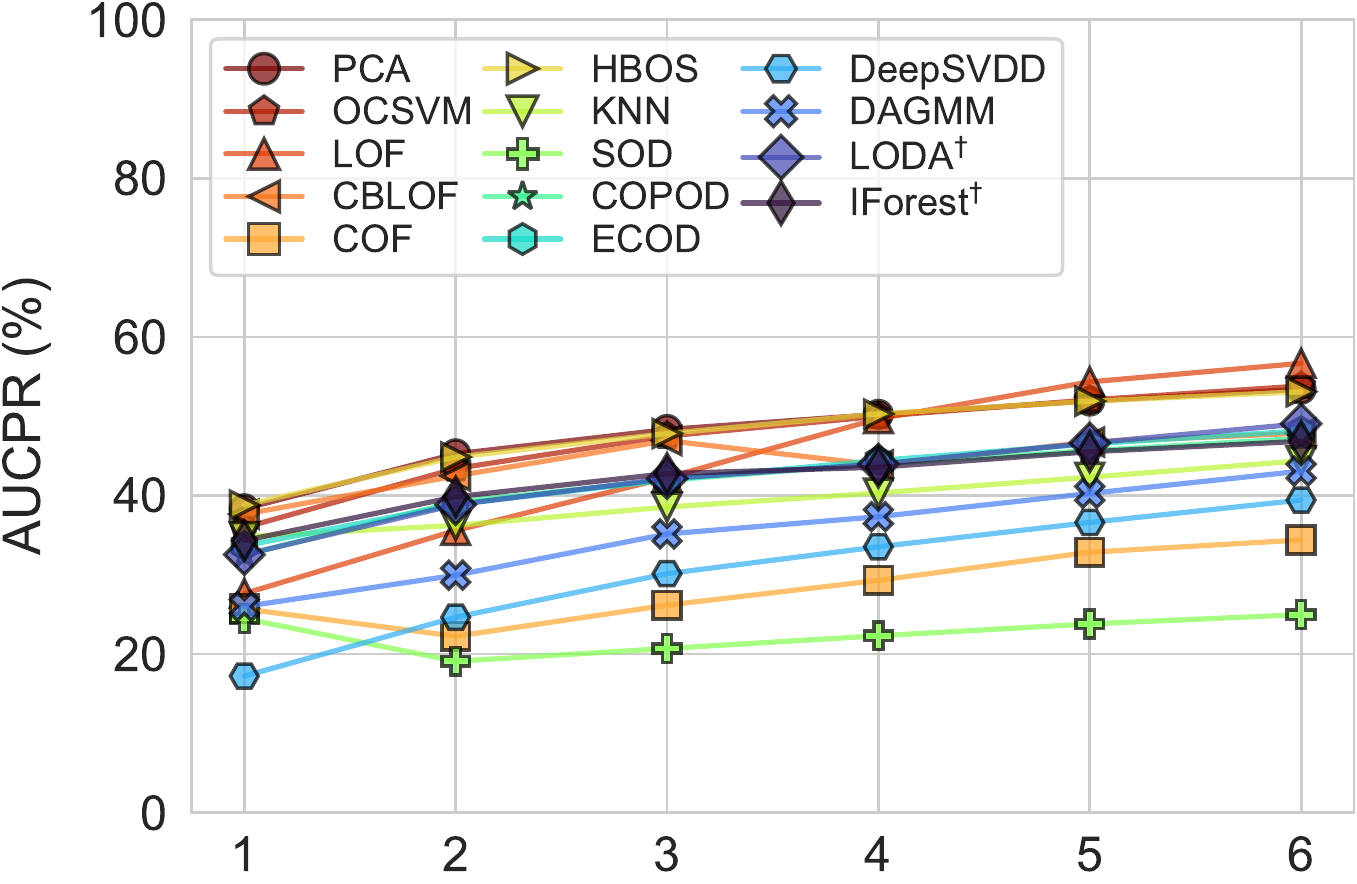}
         \caption{\scriptsize Duplicated Anomalies, unsup}
         \label{fig:dp_unsup_aucpr}
     \end{subfigure}
     \begin{subfigure}[b]{0.245\textwidth}
         \centering
         \includegraphics[width=\textwidth]{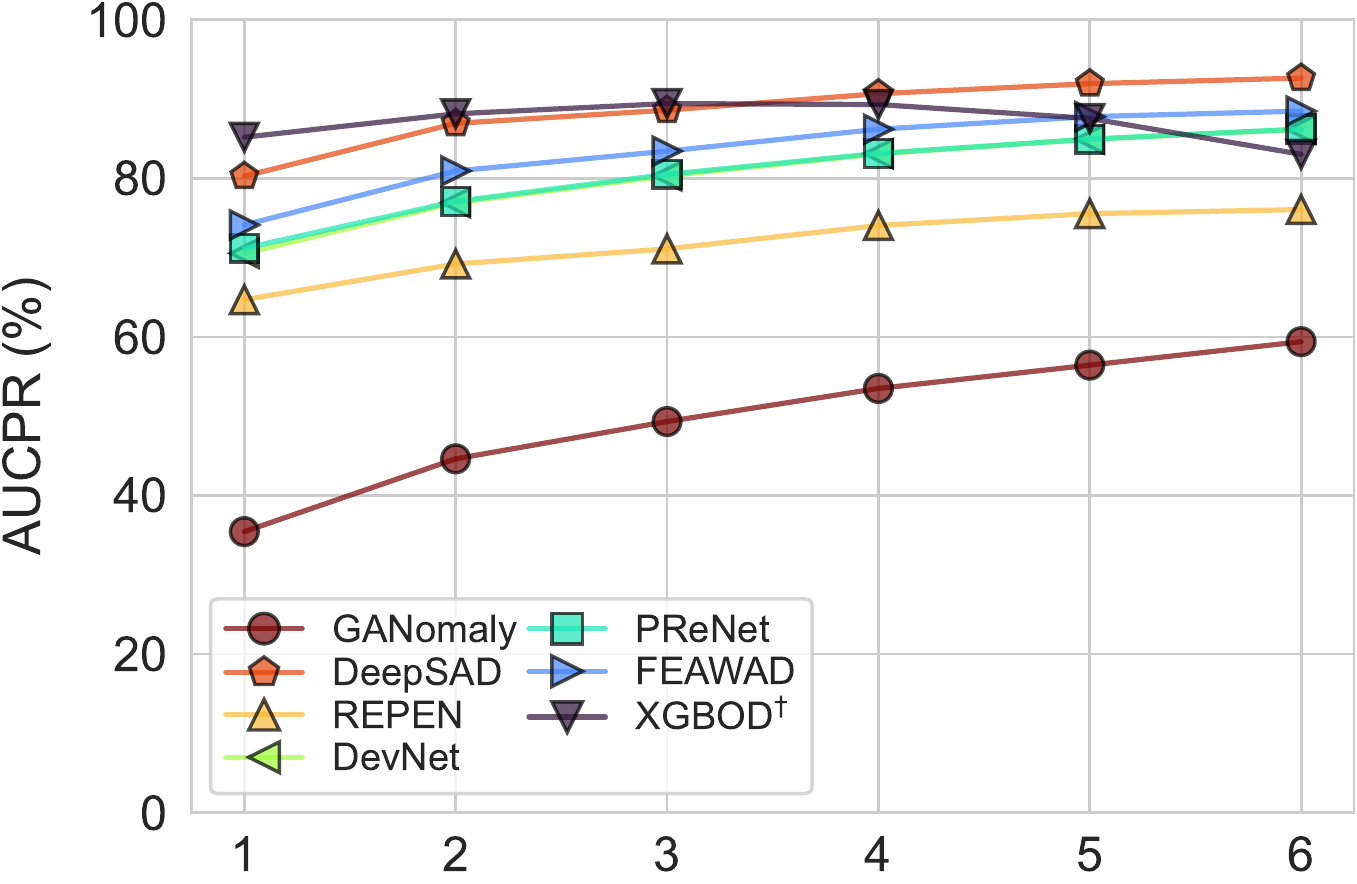}
         \caption{\scriptsize Duplicated Anomalies, semi}
         \label{fig:dp_semi_aucpr}
     \end{subfigure}
     \begin{subfigure}[b]{0.245\textwidth}
         \centering
         \includegraphics[width=\textwidth]{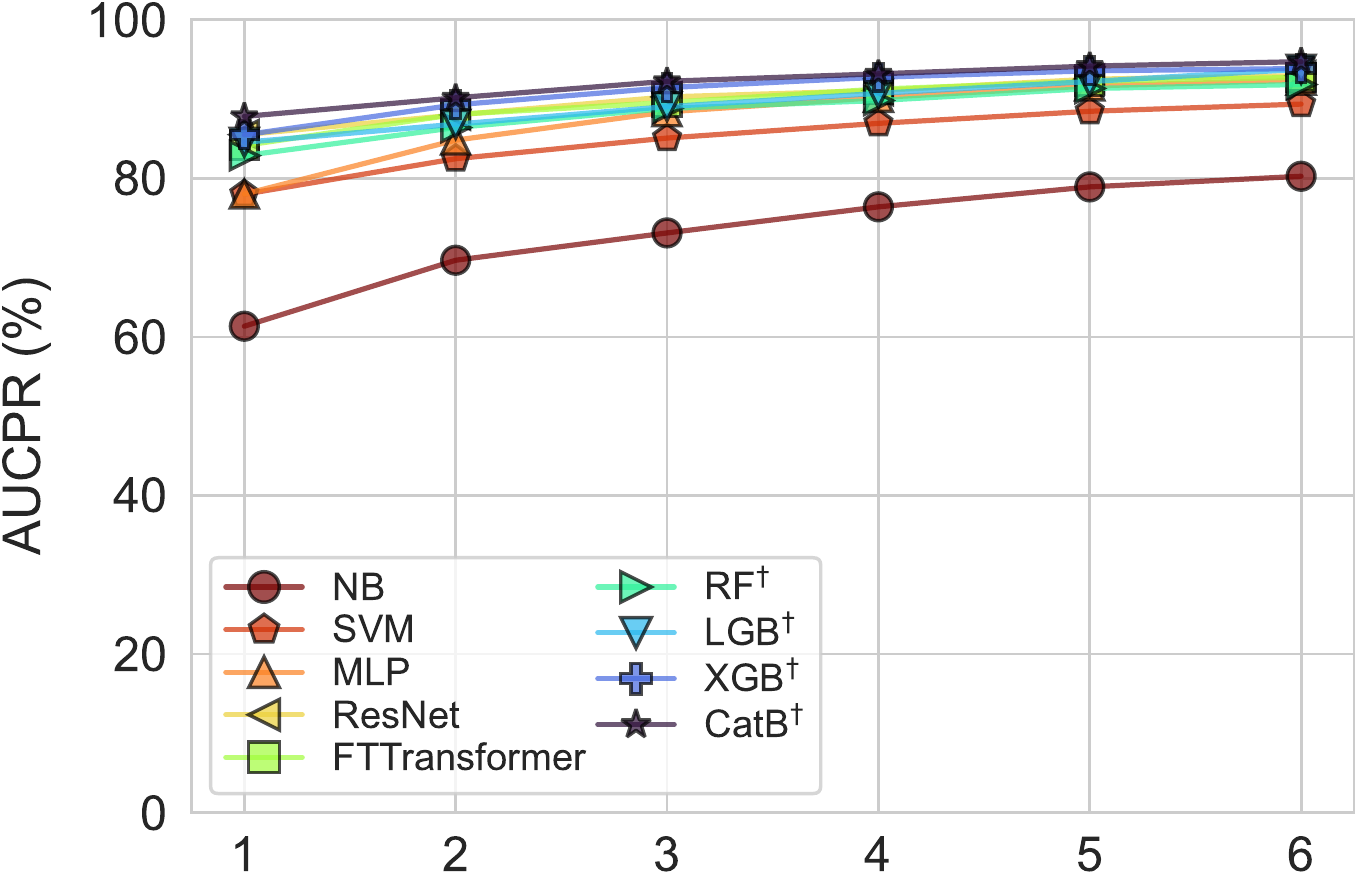}
         \caption{\scriptsize Duplicated Anomalies, sup}
         \label{fig:dp_sup_aucpr}
     \end{subfigure}
     \begin{subfigure}[b]{0.245\textwidth}
         \centering
         \includegraphics[width=\textwidth]{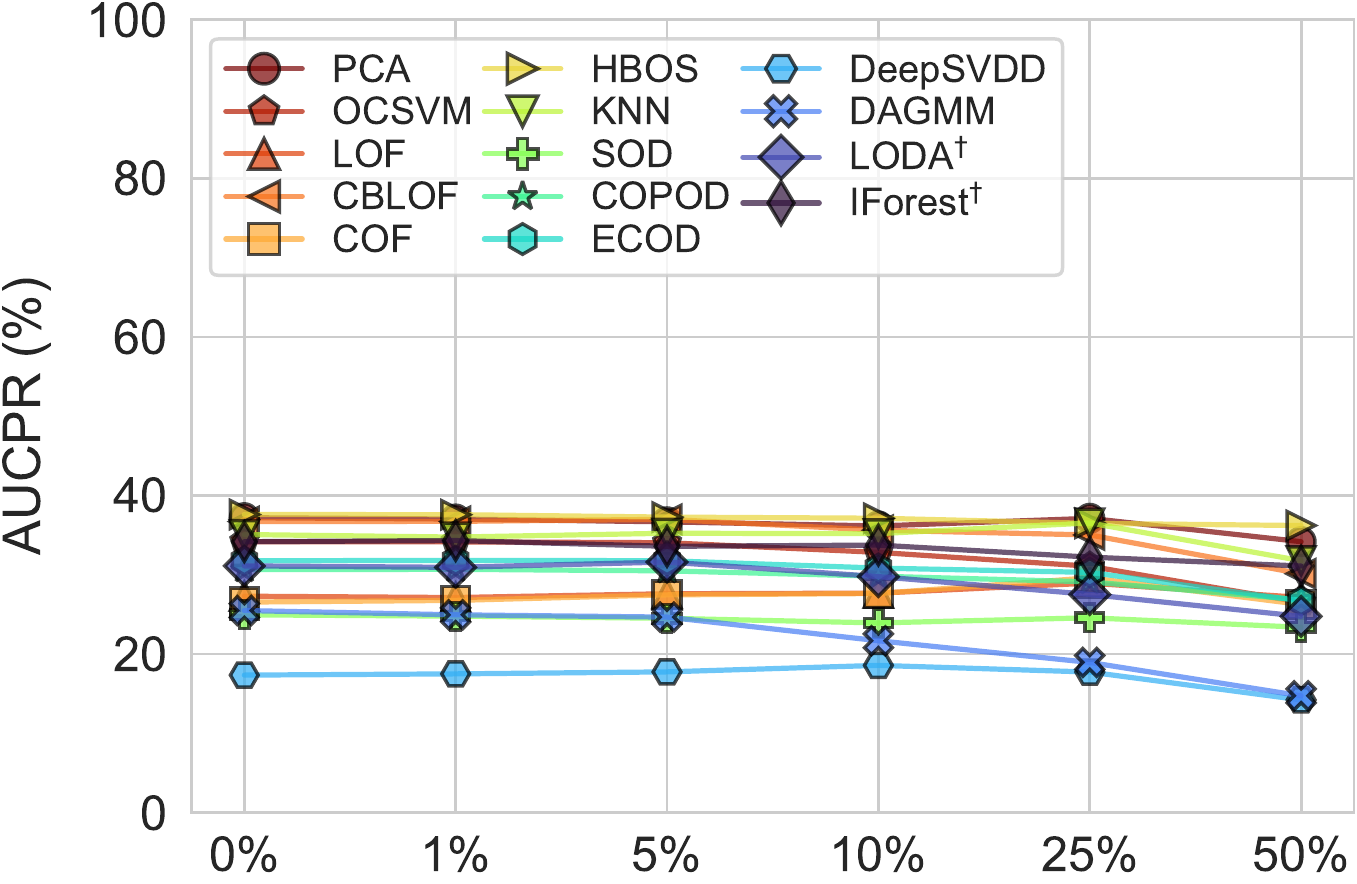}
         \caption{\scriptsize Irrelevant Features, unsup}
         \label{fig:ir_unsup_aucpr}
     \end{subfigure}
     \begin{subfigure}[b]{0.245\textwidth}
         \centering
         \includegraphics[width=\textwidth]{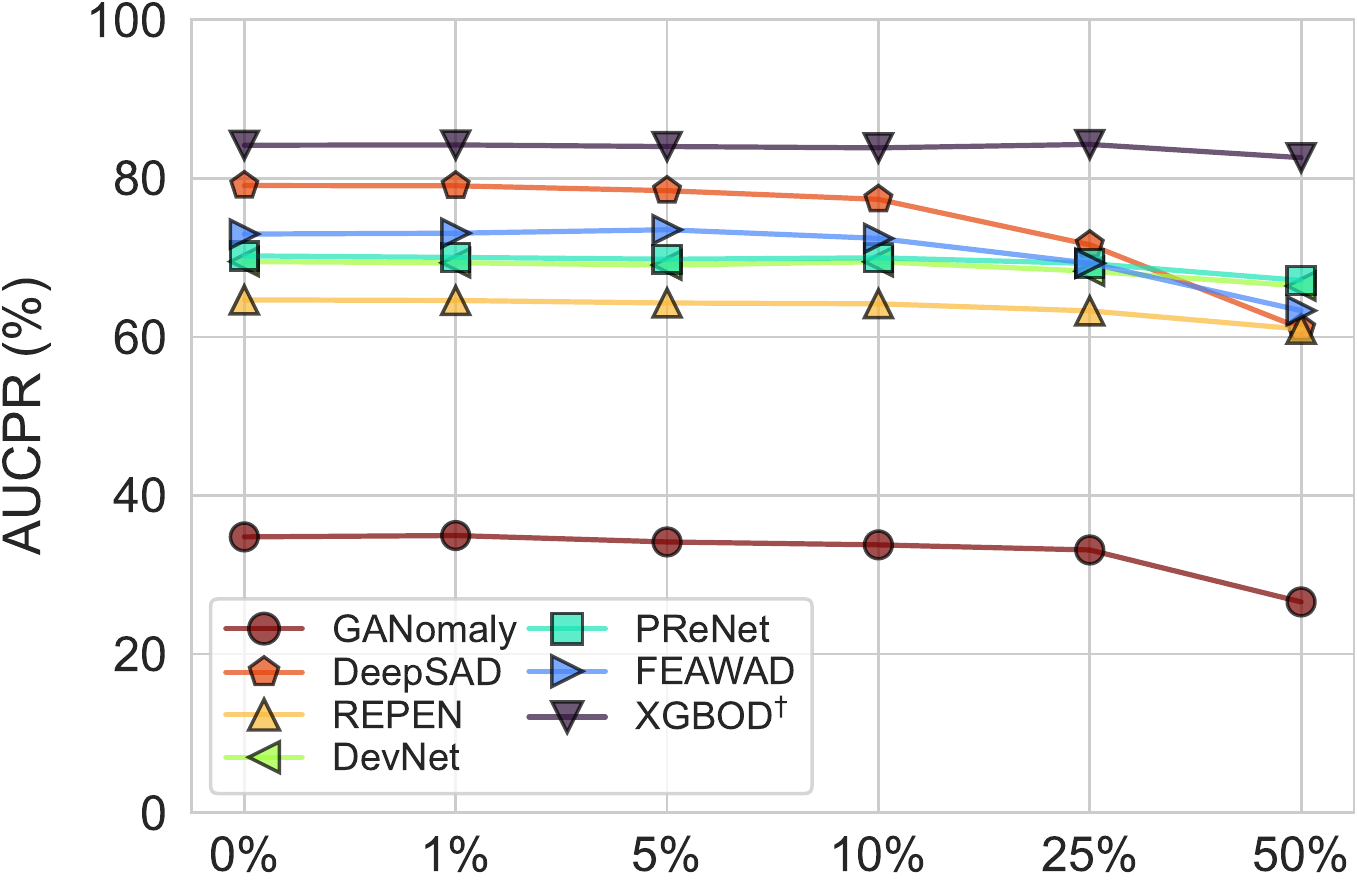}
         \caption{\scriptsize Irrelevant Features, semi}
         \label{fig:ir_semi_aucpr}
     \end{subfigure}
     \begin{subfigure}[b]{0.245\textwidth}
         \centering
         \includegraphics[width=\textwidth]{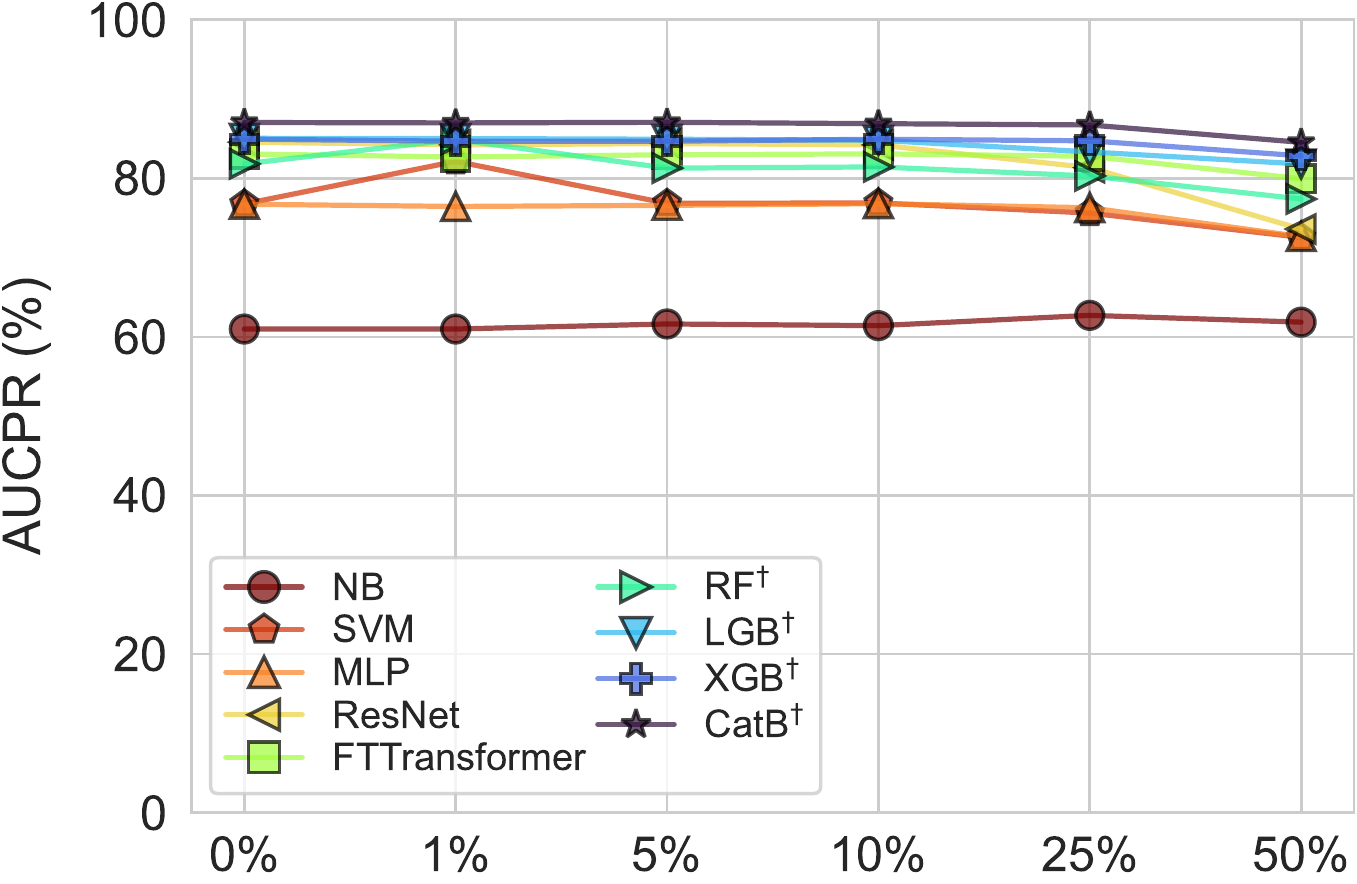}
         \caption{\scriptsize Irrelevant Features, sup}
         \label{fig:ir_sup_aucpr}
     \end{subfigure}
     \begin{subfigure}[b]{0.245\textwidth}
         \centering
         \includegraphics[width=\textwidth]{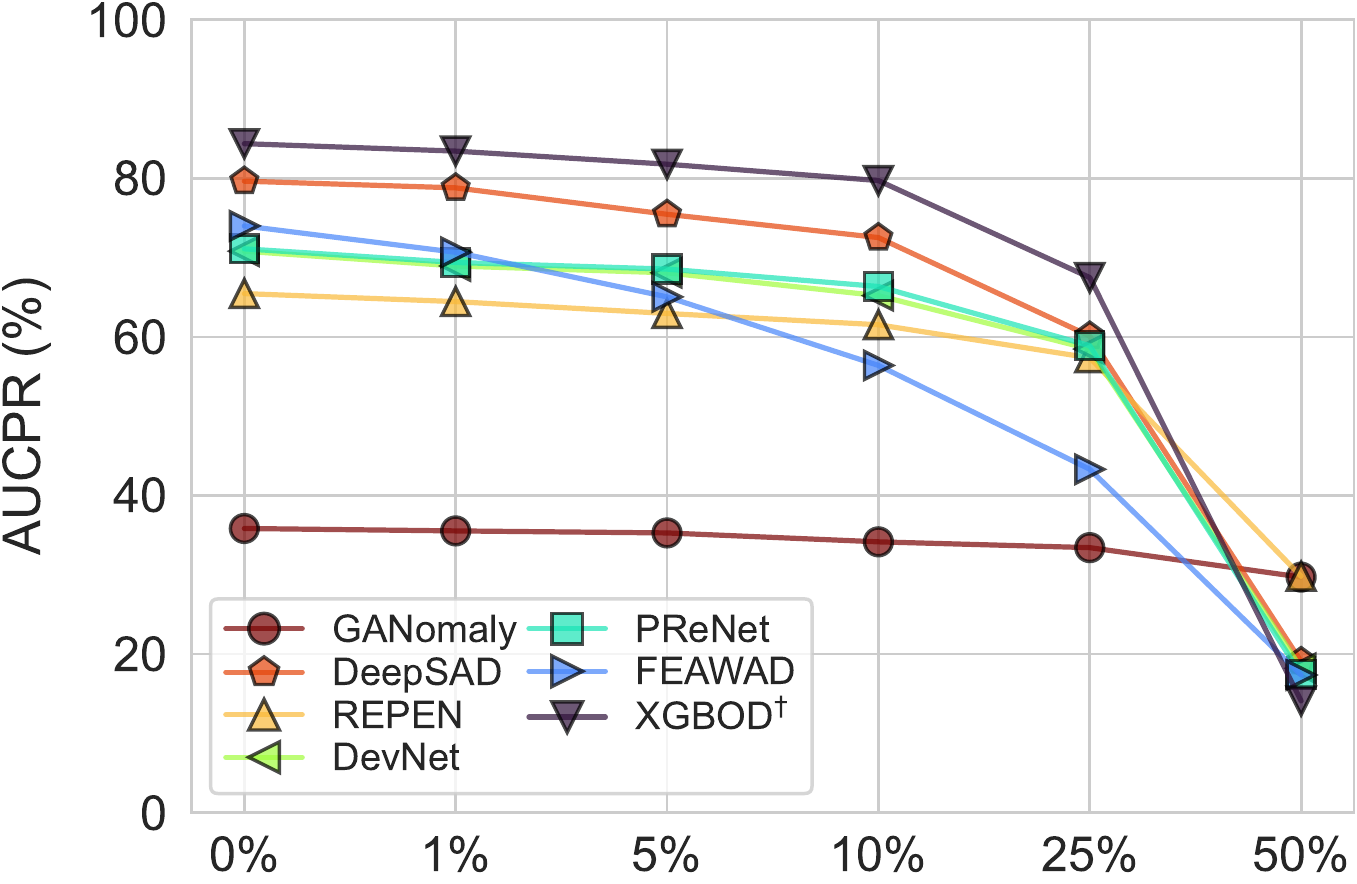}
         \caption{\scriptsize Annotation Errors, semi}
         \label{fig:ae_semi_aucpr}
     \end{subfigure}
     \begin{subfigure}[b]{0.245\textwidth}
         \centering
         \includegraphics[width=\textwidth]{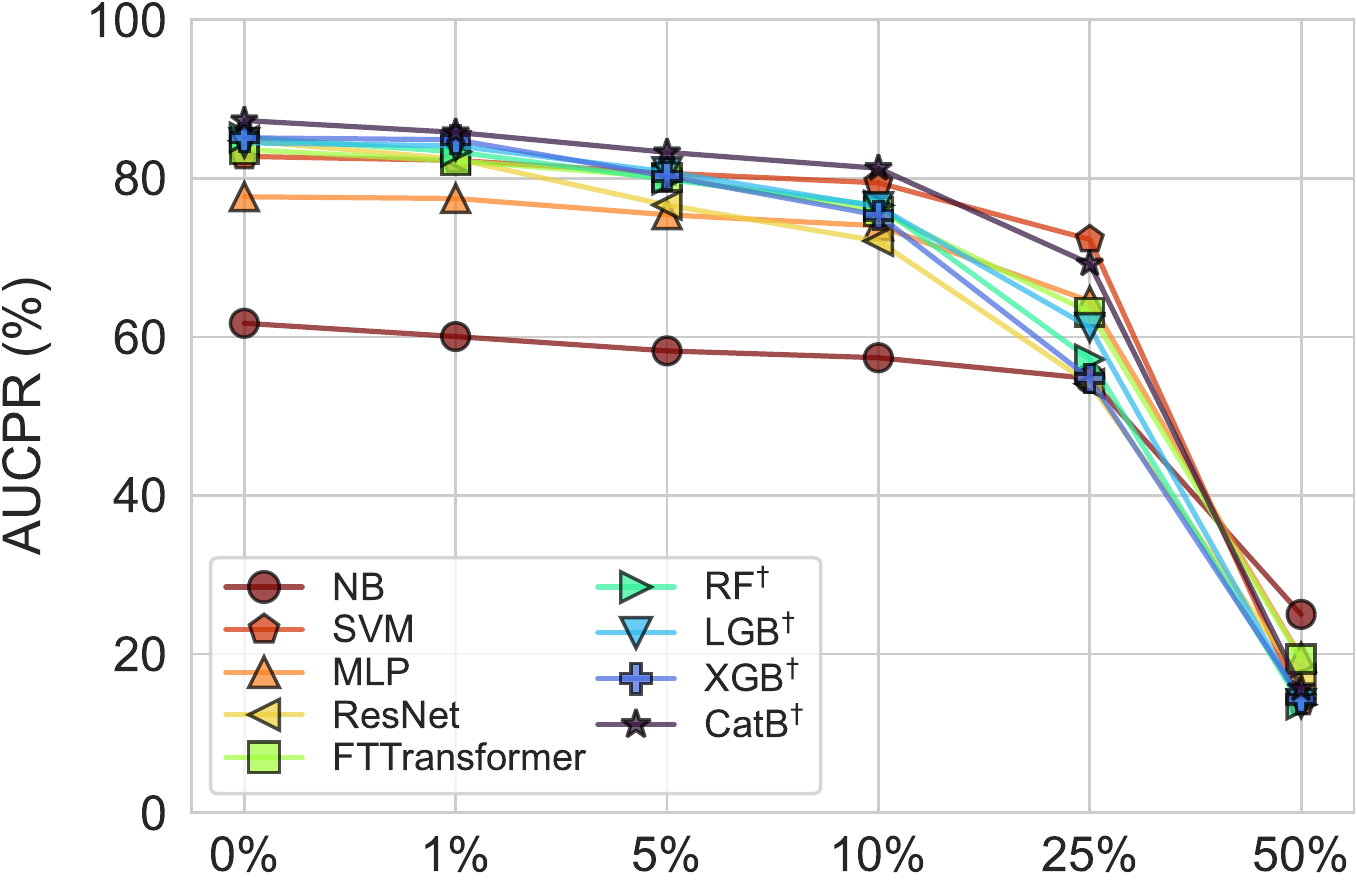}
         \caption{\scriptsize Annotation Errors, sup}
         \label{fig:ae_sup_aucpr}
     \end{subfigure}
     \vspace{-0.15in}
    \caption{Algorithm performance under noisy and corrupted data (i.e., duplicated anomalies for (a)-(c), irrelevant features for (d)-(f), and annotation errors for (g) and (h)). X-axis denotes either the duplicated times or the noise ratio. Y-axis denotes the $\text{AUCPR}$ performance and its range remains consistent across different algorithms. The results reveal unsupervised methods' susceptibility to duplicated anomalies and the usage of label information in defending irrelevant features. Un-, semi-, and fully-supervised methods are denoted as \textit{unsup}, \textit{semi}, and \textit{sup}, respectively.
    }
    \vspace{-0.1in}
    \label{fig:robustness:absolute:aucpr}
\end{figure}

In addition to the primary results shown in \S \ref{exp:robustness}, we provide the AUCPR results for algorithm robustness in Fig.~\ref{fig:model robustness aucpr} and \ref{fig:robustness:absolute:aucpr}. The AUCPR results confirm the robustness of supervised methods for irrelevant features. Besides, both semi- and fully-supervised methods are robust to minor annotation errors, say the annotation errors are less than $10\%$.

One thing to note is we observe AUCPR performance improves under the setting of duplicated anomalies (see Fig.~\ref{fig:model robustness aucpr} (a)-(c)). This is expected as AUCPR emphasizes the positive classes (i.e., anomalies), and more duplicated anomalies favor this metric. Since this observation is consistently true for both unsupervised and label-informed methods, it would not largely impact our selection of algorithms. However, if we care about both anomaly and normal classes equally, the results on AUCROC in \S \ref{exp:robustness} still stand ---unsupervised methods are more susceptible to duplicate anomalies.

\clearpage
\newpage

\subsection{Full Performance Tables on Benchmark Datasets (in addition to \S \ref{subsec:overall_performance} and Appendix \ref{appendix:exp_realworld_results})}
\label{appx:all_tables}

In the following tables, we first present the AUCROC and AUCPR for all unsupervised methods, and then show the label-informed methods' performance at different levels of labeled anomaly ratio (i.e., $\gamma_{l}=\{1\%, ..., 100\%\}$). We would expect these results are useful in constructing unsupervised anomaly detection model selection methods like MetaOD \cite{MetaOD}, where the historical algorithm performance table serves as a great source for building strong meta-learning methods.
\newpage

\input{tables/AUCROC_unsup}
\input{tables/AUCPR_unsup}

\input{tables/AUCROC_label_0.01}
\input{tables/AUCPR_label_0.01}

\input{tables/AUCROC_label_0.05}
\input{tables/AUCPR_label_0.05}

\input{tables/AUCROC_label_0.1}
\input{tables/AUCPR_label_0.1}

\input{tables/AUCROC_label_0.25}
\input{tables/AUCPR_label_0.25}

\input{tables/AUCROC_label_0.5}
\input{tables/AUCPR_label_0.5}

\input{tables/AUCROC_label_0.75}
\input{tables/AUCPR_label_0.75}

\input{tables/AUCROC_label_1}
\input{tables/AUCPR_label_1}

%% file: tables/datasets.tex
\begin{table}[!h]
\centering
\footnotesize
	\caption{Data description of the \ndatasets datasets included in \system; \ndatasetscomplex newly added datasets from CV and NLP domain are highlighted in \textcolor{blue}{blue} at the bottom of the table.
 } 
	\footnotesize
	\scalebox{0.92}{
	\begin{tabular}{l | r  r  r r r r } 
	\toprule
	 \textbf{Data}                 & \textbf{\# Samples} & \textbf{\# Features} & \textbf{\# Anomaly} & \textbf{\% Anomaly} & \textbf{Category} & \textbf{Reference}\\
	\midrule
ALOI                    & 49534   & 27       & 1508      & 3.04                &     Image     & \cite{meta_analysis}\\
annthyroid   & 7200    & 6        & 534       & 7.42                &      Healthcare   & \cite{quinlan1986induction} \\
backdoor   & 95329    & 196        & 2329        & 2.44                & Network        & \cite{moustafa2015unsw}  \\
breastw                              & 683     & 9        & 239       & 34.99               & Healthcare     & \cite{wolberg1990multisurface}    \\
campaign & 41188 & 62 & 4640 & 11.27 & Finance & \cite{devnet} \\
cardio                               & 1831    & 21       & 176       & 9.61                & Healthcare     & \cite{ayres2000sisporto}    \\
Cardiotocography    & 2114    & 21       & 466       & 22.04               & Healthcare     & \cite{ayres2000sisporto}    \\
celeba    & 202599    & 39       & 4547       & 2.24               & Image     & \cite{devnet}    \\
census & 299285 & 500 & 18568 & 6.20 & Sociology & \cite{devnet} \\
cover                                & 286048  & 10       & 2747      & 0.96                & Botany        & \cite{blackard1999comparative} \\
donors & 619326 & 10 & 36710 & 5.93 & Sociology & \cite{devnet}  \\
fault                      & 1941    & 27       & 673       & 34.67               & Physical      &  \cite{meta_analysis}   \\
fraud & 284807 & 29 & 492 & 0.17 & Finance & \cite{devnet} \\
glass & 214     & 7        & 9         & 4.21                & Forensic          & \cite{evett1989rule}\\
Hepatitis           & 80      & 19       & 13        & 16.25               & Healthcare   & \cite{diaconis1983computer}      \\
http                                 & 567498  & 3        & 2211      & 0.39                & Web        & \cite{Rayana2016} \\
InternetAds   & 1966    & 1555     & 368       & 18.72               & Image       & \cite{campos2016evaluation}  \\
Ionosphere        & 351     & 33       & 126       & 35.90                & Oryctognosy     & \cite{sigillito1989classification}    \\
landsat                         & 6435    & 36       & 1333      & 20.71               & Astronautics   & \cite{meta_analysis}      \\
letter                               & 1600    & 32       & 100       & 6.25                & Image   &\cite{frey1991letter}      \\
Lymphography       & 148     & 18       & 6         & 4.05                & Healthcare     &\cite{cestnik1987knowledge}    \\
magic.gamma                     & 19020   & 10       & 6688      & 35.16               & Physical        &\cite{meta_analysis} \\
mammography                          & 11183   & 6        & 260       & 2.32                & Healthcare      &\cite{woods1994comparative}   \\
mnist                                & 7603    & 100      & 700       & 9.21                & Image     & \cite{lecun1998gradient}    \\
musk                                 & 3062    & 166      & 97        & 3.17                & Chemistry      & \cite{dietterich1993comparison}   \\
optdigits                            & 5216    & 64       & 150       & 2.88                & Image      & \cite{alpaydin1998cascading}   \\
PageBlocks         & 5393    & 10       & 510       & 9.46                & Document     & \cite{malerba1996further}     \\
pendigits                            & 6870    & 16       & 156       & 2.27                & Image      & \cite{alimoglu1996methods}   \\
Pima                & 768     & 8        & 268       & 34.90                & Healthcare       & \cite{Rayana2016}  \\
satellite                            & 6435    & 36       & 2036      & 31.64               & Astronautics      & \cite{Rayana2016}   \\
satimage-2                           & 5803    & 36       & 71        & 1.22                & Astronautics     & \cite{Rayana2016}    \\
shuttle                              & 49097   & 9        & 3511      & 7.15                & Astronautics     & \cite{Rayana2016}     \\
skin                            & 245057  & 3        & 50859     & 20.75               &    Image    & \cite{meta_analysis} \\
smtp                                 & 95156   & 3        & 30        & 0.03                & Web        & \cite{Rayana2016} \\
SpamBase            & 4207    & 57       & 1679      & 39.91               & Document      & \cite{campos2016evaluation}   \\
speech                               & 3686    & 400      & 61        & 1.65                & Linguistics     & \cite{brummer2012description}    \\
Stamps              & 340     & 9        & 31        & 9.12                & Document       & \cite{campos2016evaluation}  \\
thyroid                              & 3772    & 6        & 93        & 2.47                & Healthcare       & \cite{quinlan1987inductive}  \\
vertebral                            & 240     & 6        & 30        & 12.50                & Biology    & \cite{berthonnaud2005analysis}     \\
vowels                               & 1456    & 12       & 50        & 3.43                & Linguistics   & \cite{kudo1999multidimensional}       \\
Waveform           & 3443    & 21       & 100       & 2.90                 & Physics       & \cite{loh2011classification}  \\
WBC                & 223     & 9        & 10        & 4.48                & Healthcare      & \cite{mangasarian1995breast}   \\
WDBC               & 367     & 30       & 10        & 2.72                & Healthcare     & \cite{mangasarian1995breast}     \\
Wilt                & 4819    & 5        & 257       & 5.33                & Botany        & \cite{campos2016evaluation} \\
wine                                 & 129     & 13       & 10        & 7.75                & Chemistry     & \cite{aeberhard1992classification}    \\
WPBC             & 198     & 33       & 47        & 23.74               & Healthcare      & \cite{mangasarian1995breast}    \\
yeast                           & 1484    & 8        & 507       & 34.16               & Biology      & \cite{horton1996probabilistic}  \\
    \midrule
\textcolor{blue}{CIFAR10}                           & \textcolor{blue}{5263}    & \textcolor{blue}{512}       & \textcolor{blue}{263}       & \textcolor{blue}{5.00}               & \textcolor{blue}{Image} & \cite{krizhevsky2009learning} \\
\textcolor{blue}{FashionMNIST}                           & \textcolor{blue}{6315}   & \textcolor{blue}{512}        & \textcolor{blue}{315}       & \textcolor{blue}{5.00}               & \textcolor{blue}{Image} & \cite{xiao2017fashion}\\
\textcolor{blue}{MNIST-C}                           & \textcolor{blue}{10000}    & \textcolor{blue}{512}       & \textcolor{blue}{500}       & \textcolor{blue}{5.00}               & \textcolor{blue}{Image} & \cite{mu2019mnist} \\
\textcolor{blue}{MVTec-AD} & \multicolumn{4}{c}{\textcolor{blue}{See Table \ref{tab:mvtec}.} } & \textcolor{blue}{Image} & \cite{bergmann2019mvtec}\\
\textcolor{blue}{SVHN}                           & \textcolor{blue}{5208}    & \textcolor{blue}{512}       & \textcolor{blue}{260}       & \textcolor{blue}{5.00}               & \textcolor{blue}{Image} & \cite{netzer2011reading}\\
    \midrule
\textcolor{blue}{Agnews}                           & \textcolor{blue}{10000}    & \textcolor{blue}{768}       & \textcolor{blue}{500}       & \textcolor{blue}{5.00}               & \textcolor{blue}{NLP} & \cite{zhang2015character}\\
\textcolor{blue}{Amazon}                           & \textcolor{blue}{10000}    & \textcolor{blue}{768}       & \textcolor{blue}{500}       & \textcolor{blue}{5.00}               & \textcolor{blue}{NLP} & \cite{he2016ups}\\
\textcolor{blue}{Imdb}                           & \textcolor{blue}{10000}    & \textcolor{blue}{768}       & \textcolor{blue}{500}       & \textcolor{blue}{5.00}               & \textcolor{blue}{NLP} & \cite{maas-EtAl:2011:ACL-HLT2011}\\
\textcolor{blue}{Yelp}                          & \textcolor{blue}{10000}    & \textcolor{blue}{768}       & \textcolor{blue}{500}       & \textcolor{blue}{5.00}               & \textcolor{blue}{NLP} & \cite{zhang2015character}\\
\textcolor{blue}{20newsgroups} & \multicolumn{4}{c}{\textcolor{blue}{See Table \ref{tab:20news}.}} & \textcolor{blue}{NLP} & \cite{Lang95}\\
    \bottomrule
	\end{tabular}
	}
	\label{table:datasets} 
\end{table}

%% file: tables/MVTec-AD.tex
\begin{table}[h]
  \centering
  \caption{Detailed description of the MVTec-AD dataset; see the full dataset list in Table \ref{table:datasets}. For MVTec-AD dataset, we evaluate \nmodels algorithms on each class and report the average performance of all classes.}
  \scalebox{0.9}{
    \begin{tabular}{l|rrrr}
    \toprule
    \textbf{Class} & \multicolumn{1}{l}{\textbf{\# Samples}} & \multicolumn{1}{l}{\textbf{\# Features}} & \multicolumn{1}{l}{\textbf{\# Anomaly}} & \multicolumn{1}{l}{\textbf{\% Anomaly}} \\
    \midrule
    Carpet & 397   & 512   & 89    & 22.42 \\
    Grid  & 342   & 512   & 57    & 16.67 \\
    Leather & 369   & 512   & 92    & 24.93 \\
    Tile  & 347   & 512   & 84    & 24.21 \\
    Wood  & 326   & 512   & 60    & 18.40 \\
    Bottle & 292   & 512   & 63    & 21.58 \\
    Cable & 374   & 512   & 92    & 24.60 \\
    Capsule & 351   & 512   & 109   & 31.05 \\
    Hazelnut & 501   & 512   & 70    & 13.97 \\
    Metal Nut & 335   & 512   & 93    & 27.76 \\
    Pill  & 434   & 512   & 141   & 32.49 \\
    Screw & 480   & 512   & 119   & 24.79 \\
    Toothbrush & 102   & 512   & 30    & 29.41 \\
    Transistor & 313   & 512   & 40    & 12.78 \\
    Zipper & 391   & 512   & 119   & 30.43 \\
    \midrule
    Total & 5354  & 512   & 1258  & 23.50 \\
    \bottomrule
    \end{tabular}%
  }
  \label{tab:mvtec}%
\end{table}%

%% file: tables/20news.tex
\begin{table}[h]
  \centering
  \caption{Detailed description of the 20newsgroups dataset; see the full dataset list in Table \ref{table:datasets}. For 20newsgroups dataset, we evaluate \nmodels algorithms on each class and report the average performance of all classes.}
  \scalebox{0.88}{
    \begin{tabular}{l|rrrr}
    \toprule
    \textbf{Class} & \multicolumn{1}{l}{\textbf{\# Samples}} & \multicolumn{1}{l}{\textbf{\# Features}} & \multicolumn{1}{l}{\textbf{\# Anomaly}} & \multicolumn{1}{l}{\textbf{\% Anomaly}} \\
    \midrule
    Computer & 3090  & 768   & 154   & 4.98 \\
    Recreation & 2514  & 768   & 125   & 4.97 \\
    Science & 2497  & 768   & 124   & 4.97 \\
    Miscellaneous & 615   & 768   & 30    & 4.88 \\
    Politics & 1657  & 768   & 82    & 4.95 \\
    Religion & 1532  & 768   & 76    & 4.96 \\
    \midrule
    Total & 11905 & 768   & 591   & 4.96 \\
    \bottomrule
    \end{tabular}%
  }
  \label{tab:20news}%
\end{table}%

%% file: tables/AUCROC_unsup.tex
\begin{table}[t]
\scriptsize
  \centering
  \caption{AUCROC of \nunsup unsupervised algorithms on \ndatasets benchmark datasets. We show the performance rank in parenthesis (the lower, the better), and mark the best performing method(s) in \textbf{bold}.
  }
\scalebox{0.65}{

\begin{tabular}{l|llllllllllllll}
    \toprule
    \textbf{Datasets}         & \textbf{PCA} & \textbf{OCSVM} & \textbf{LOF} & \textbf{CBLOF} & \textbf{COF} & \textbf{HBOS} & \textbf{KNN} & \textbf{SOD} & \textbf{COPOD} & \textbf{ECOD} & \textbf{\makecell{Deep\\SVDD}} & \textbf{\makecell{DA\\GMM}} & \textbf{LODA} & \textbf{IForest} \\
    \midrule
    ALOI  & 56.65(6) & 55.85(8) & \textbf{66.63(1)} & 55.22(9) & 64.68(2) & 52.63(11) & 61.47(3) & 61.09(4) & 53.75(10) & 56.60(7) & 50.29(14) & 51.96(12) & 51.33(13) & 56.66(5) \\
    annthyroid & 66.25(8) & 57.23(12) & 70.20(7) & 62.26(10) & 65.92(9) & 60.15(11) & 71.69(6) & 77.38(3) & 76.80(4) & 78.03(2) & 76.62(5) & 56.53(13) & 41.02(14) & \textbf{82.01(1)} \\
    backdoor & 80.13(7) & 86.20(2) & 85.68(3) & 81.16(4) & 73.03(8) & 71.43(10) & 80.82(6) & 69.54(11) & 80.97(5) & \textbf{86.33(1)} & 55.16(14) & 56.26(13) & 69.22(12) & 72.15(9) \\
    breastw & 95.13(8) & 80.30(10) & 40.61(12) & 96.81(7) & 38.84(13) & 98.94(3) & 97.01(6) & 93.97(9) & \textbf{99.68(1)} & 99.17(2) & 65.66(11) & N/A(N/A) & 98.49(4) & 98.32(5) \\
    campaign & 72.78(4) & 65.52(9) & 58.85(10) & 66.61(8) & 57.26(11) & \textbf{78.61(1)} & 72.10(5) & 69.04(7) & 77.69(2) & 76.78(3) & 48.70(14) & 56.08(12) & 51.43(13) & 71.71(6) \\
    cardio & \textbf{95.55(1)} & 93.91(3) & 66.33(13) & 89.93(7) & 71.41(12) & 84.67(8) & 76.64(9) & 73.25(11) & 92.35(5) & 94.44(2) & 58.96(14) & 75.01(10) & 90.34(6) & 93.19(4) \\
    Cardiotocography & 74.67(2) & \textbf{77.86(1)} & 59.51(10) & 64.54(7) & 53.77(12) & 60.86(9) & 56.23(11) & 51.69(14) & 67.02(6) & 68.92(4) & 53.53(13) & 62.01(8) & 73.65(3) & 67.57(5) \\
\bottomrule
    \end{tabular}%
}
  \label{tab:roc_unsup}%
\end{table}

%% file: tables/AUCPR_unsup.tex
\begin{table}[b]
\scriptsize
  \centering
  \caption{AUCPR of \nunsup unsupervised algorithms on \ndatasets benchmark datasets. We show the performance rank in parenthesis (lower the better), and mark the best performing method(s) in \textbf{bold}.
  }
\scalebox{0.65}{

    \begin{tabular}{l|llllllllllllll}
    \toprule
    \textbf{Datasets}         & \textbf{PCA} & \textbf{OCSVM} & \textbf{LOF} & \textbf{CBLOF} & \textbf{COF} & \textbf{HBOS} & \textbf{KNN} & \textbf{SOD} & \textbf{COPOD} & \textbf{ECOD} & \textbf{\makecell{Deep\\SVDD}} & \textbf{\makecell{DA\\GMM}} & \textbf{LODA} & \textbf{IForest} \\
    \midrule
    ALOI  & 4.17(9) & 5.02(5) & \textbf{8.08(1)} & 4.46(7) & 6.85(2) & 3.69(13) & 6.02(3) & 5.97(4) & 3.62(14) & 3.90(11) & 4.01(10) & 4.33(8) & 4.53(6) & 3.90(12) \\
    annthyroid & 16.12(8) & 10.37(12) & 15.71(9) & 13.69(11) & 14.39(10) & 16.99(5) & 16.74(6) & 18.84(4) & 16.58(7) & 24.65(2) & 21.95(3) & 9.64(13) & 7.06(14) & \textbf{30.47(1)} \\
    backdoor & 31.29(3) & 9.69(9) & 26.14(4) & 6.96(11) & 24.68(5) & 4.91(13) & \textbf{45.22(1)} & 39.41(2) & 7.69(10) & 11.25(8) & 12.85(7) & 6.50(12) & 14.51(6) & 4.75(14) \\
    breastw & 95.11(6) & 82.70(10) & 28.55(12) & 91.54(8) & 27.60(13) & 97.71(3) & 92.19(7) & 84.88(9) & \textbf{99.40(1)} & 98.54(2) & 50.92(11) & N/A(N/A) & 97.04(4) & 96.04(5) \\
    campaign & 27.90(6) & 29.22(5) & 14.51(11) & 23.99(8) & 13.01(13) & 37.99(2) & 27.18(7) & 18.88(9) & \textbf{38.58(1)} & 37.40(3) & 11.60(14) & 14.62(10) & 13.47(12) & 32.26(4) \\
    cardio & 66.06(2) & 62.89(3) & 23.79(13) & 61.95(4) & 28.67(11) & 52.10(8) & 40.72(9) & 28.54(12) & 60.42(5) & \textbf{68.59(1)} & 22.50(14) & 28.92(10) & 53.41(7) & 59.95(6) \\
    Cardiotocography & 47.95(3) & \textbf{52.61(1)} & 30.66(11) & 45.44(4) & 28.21(13) & 38.28(8) & 34.79(9) & 27.99(14) & 40.46(7) & 43.57(5) & 34.03(10) & 30.61(12) & 48.00(2) & 41.47(6) \\
\bottomrule
    \end{tabular}%
}
  \label{tab:pr_unsup}%
\end{table}

%% file: tables/AUCROC_label_0.01.tex
\begin{table}[h]
\scriptsize
  \centering
  \caption{AUCROC of 16 label-informed algorithms on \ndatasets benchmark datasets, with labeled anomaly ratio $\gamma_{l}=1\%$. We show the performance rank in parenthesis (lower the better), and mark the best performing method(s) in \textbf{bold}.
  }
\scalebox{0.57}{

    \begin{tabular}{l|lllllll|lllllllll}
    \toprule
    \textbf{Datasets}         & \textbf{GANomaly} & \textbf{DeepSAD} & \textbf{REPEN} & \textbf{DevNet} & \textbf{PReNet} & \textbf{FEAWAD} & \textbf{XGBOD} & \textbf{NB} & \textbf{SVM} & \textbf{MLP} & \textbf{ResNet} & \textbf{\makecell{FTTrans\\former}}  & \textbf{RF} & \textbf{LGB} & \textbf{XGB} & \textbf{CatB} \\
    \midrule
    ALOI  & 55.53(3) & 59.13(2) & 54.53(5) & 47.03(15) & 46.47(16) & 55.06(4) & \textbf{60.53(1)} & 49.31(13) & 52.42(8) & 48.50(14) & 49.89(12) & 51.74(9) & 51.04(10) & 50.60(11) & 53.08(7) & 53.22(6) \\
    annthyroid & 75.67(9) & 76.82(7) & 72.20(11) & 74.78(10) & 75.95(8) & 77.66(5) & 92.89(2) & 80.62(4) & 59.25(15) & 62.84(12) & 52.04(16) & 76.97(6) & 60.93(14) & 62.62(13) & 89.91(3) & \textbf{95.58(1)} \\
    backdoor & 82.28(9) & 91.98(3) & 89.44(5) & 92.91(2) & \textbf{94.23(1)} & 82.60(7) & 85.67(6) & 63.10(13) & 80.03(10) & 89.69(4) & 71.15(12) & 62.89(14) & 58.05(15) & 37.26(16) & 82.55(8) & 76.47(11) \\
    breastw & 91.99(3) & 88.36(5) & 86.03(7) & 74.61(10) & 69.24(12) & 78.39(9) & 86.57(6) & 52.11(16) & 71.27(11) & 52.59(15) & 53.60(14) & 95.70(2) & 65.01(13) & 90.20(4) & 84.81(8) & \textbf{97.30(1)} \\
    campaign & 56.58(12) & 64.32(7) & 57.57(11) & 66.20(6) & 67.62(5) & 58.45(10) & 74.21(2) & 51.87(14) & 47.20(16) & 60.69(9) & 49.37(15) & 61.16(8) & 56.28(13) & 68.27(4) & 73.81(3) & \textbf{81.61(1)} \\
    cardio & 82.03(6) & 69.21(12) & 83.07(5) & 89.74(2) & 87.54(3) & 79.57(8) & 83.94(4) & 58.89(15) & 79.91(7) & 61.55(14) & N/A(N/A) & 75.38(9) & 63.98(13) & 73.56(11) & 74.67(10) & \textbf{94.01(1)} \\
    Cardiotocography & 53.99(15) & 64.94(12) & 81.22(3) & 81.91(2) & 79.33(4) & 70.67(10) & 71.54(9) & 76.00(7) & 54.85(14) & 48.18(16) & 65.16(11) & 76.31(6) & 60.95(13) & 75.31(8) & 78.81(5) & \textbf{85.89(1)} \\
\bottomrule
    \end{tabular}%
}
\end{table}

%% file: tables/AUCPR_label_0.01.tex
\begin{table}[h]
\scriptsize
  \centering
  \caption{AUCPR of 16 label-informed algorithms on \ndatasets benchmark datasets, with labeled anomaly ratio $\gamma_{l}=1\%$. We show the performance rank in parenthesis (lower the better), and mark the best performing method(s) in \textbf{bold}.
  }
\scalebox{0.55}{

    \begin{tabular}{l|lllllll|lllllllll}
    \toprule
    \textbf{Datasets}         & \textbf{GANomaly} & \textbf{DeepSAD} & \textbf{REPEN} & \textbf{DevNet} & \textbf{PReNet} & \textbf{FEAWAD} & \textbf{XGBOD} & \textbf{NB} & \textbf{SVM} & \textbf{MLP} & \textbf{ResNet} & \textbf{\makecell{FTTrans\\former}}  & \textbf{RF} & \textbf{LGB} & \textbf{XGB} & \textbf{CatB} \\
    \midrule
    ALOI  & 3.83(14) & 6.10(2) & 4.51(6) & 3.76(15) & 3.86(13) & 4.56(4) & \textbf{6.68(1)} & 3.29(16) & 4.99(3) & 4.08(11) & 4.33(7) & 3.95(12) & 4.25(8) & 4.12(9) & 4.08(10) & 4.53(5) \\
    annthyroid & 34.35(5) & 29.25(9) & 27.38(11) & 31.79(8) & 33.92(6) & 32.05(7) & \textbf{65.84(1)} & 43.64(4) & 13.79(15) & 12.99(16) & 15.88(14) & 29.12(10) & 22.34(12) & 21.96(13) & 54.84(3) & 61.79(2) \\
    backdoor & 6.95(16) & 38.49(7) & 28.84(10) & \textbf{87.83(1)} & 86.20(2) & 68.50(4) & 25.79(12) & 28.06(11) & 51.54(5) & 71.48(3) & 34.99(8) & 33.83(9) & 16.55(14) & 10.42(15) & 42.00(6) & 19.24(13) \\
    breastw & 85.14(3) & 79.62(9) & 81.19(7) & 77.74(10) & 72.83(12) & 80.86(8) & 83.82(5) & 54.37(14) & 76.04(11) & 56.04(13) & 51.70(16) & 90.82(2) & 54.04(15) & 85.12(4) & 81.93(6) & \textbf{93.23(1)} \\
    campaign & 23.13(7) & 19.87(9) & 15.09(13) & 23.18(6) & 23.62(5) & 20.35(8) & 32.98(2) & 11.83(16) & 13.13(15) & 19.21(10) & 13.51(14) & 18.71(11) & 16.55(12) & 25.70(4) & 28.63(3) & \textbf{36.63(1)} \\
    cardio & 38.52(10) & 30.03(12) & 54.24(4) & 66.86(2) & 63.47(3) & 52.19(5) & 50.41(6) & 20.48(14) & 41.60(9) & 14.11(15) & N/A(N/A) & 48.60(7) & 29.30(13) & 42.16(8) & 36.18(11) & \textbf{70.42(1)} \\
    Cardiotocography & 33.18(15) & 40.99(12) & 58.09(5) & 61.74(2) & 58.84(4) & 52.60(9) & 46.13(10) & 53.30(7) & 37.51(13) & 21.50(16) & 46.12(11) & 59.03(3) & 34.71(14) & 52.95(8) & 55.45(6) & \textbf{65.09(1)} \\
\bottomrule
    \end{tabular}%
}
\end{table}

%% file: tables/AUCROC_label_0.05.tex
\begin{table}[h]
\scriptsize
  \centering
  \caption{AUCROC of 16 label-informed algorithms on \ndatasets benchmark datasets, with labeled anomaly ratio $\gamma_{l}=5\%$. We show the performance rank in parenthesis (lower the better), and mark the best performing method(s) in \textbf{bold}.
  }
\scalebox{0.55}{

    \begin{tabular}{l|lllllll|lllllllll}
    \toprule
    \textbf{Datasets}         & \textbf{GANomaly} & \textbf{DeepSAD} & \textbf{REPEN} & \textbf{DevNet} & \textbf{PReNet} & \textbf{FEAWAD} & \textbf{XGBOD} & \textbf{NB} & \textbf{SVM} & \textbf{MLP} & \textbf{ResNet} & \textbf{\makecell{FTTrans\\former}}  & \textbf{RF} & \textbf{LGB} & \textbf{XGB} & \textbf{CatB} \\
    \midrule
    ALOI  & 56.71(4) & 57.88(2) & 52.13(9) & 47.46(16) & 48.51(13) & 52.77(8) & \textbf{65.99(1)} & 48.37(14) & 49.86(11) & 48.95(12) & 47.91(15) & 51.97(10) & 56.68(5) & 55.28(6) & 57.22(3) & 55.05(7) \\
    annthyroid & 76.22(14) & 79.92(11) & 78.91(12) & 81.64(10) & 82.42(7) & 89.05(6) & 97.24(2) & 81.99(9) & 49.80(16) & 71.31(15) & 77.35(13) & 95.13(4) & 82.25(8) & 93.14(5) & 95.24(3) & \textbf{98.51(1)} \\
    backdoor & 81.81(12) & 95.13(3) & 89.33(10) & 95.58(2) & 94.37(4) & \textbf{96.80(1)} & 91.07(8) & 79.06(13) & 86.80(11) & 93.12(7) & 66.31(16) & 89.84(9) & 76.68(14) & 76.63(15) & 93.34(6) & 94.34(5) \\
    breastw & 92.28(11) & 90.67(12) & 97.61(4) & 99.18(2) & 92.79(9) & 92.67(10) & 96.72(6) & \textbf{99.73(1)} & 72.44(15) & 95.42(7) & 49.28(16) & 88.59(13) & 85.59(14) & 97.53(5) & 95.12(8) & 99.10(3) \\
    campaign & 64.43(14) & 69.71(10) & 57.88(16) & 81.03(5) & 80.65(6) & 71.01(8) & 88.41(2) & 69.02(11) & 68.58(12) & 70.18(9) & 59.17(15) & 67.08(13) & 72.78(7) & 84.86(4) & 86.48(3) & \textbf{88.89(1)} \\
    cardio & 82.46(13) & 79.71(14) & 95.56(5) & \textbf{97.12(1)} & 96.36(3) & 87.96(10) & 95.73(4) & 71.04(15) & 83.30(12) & 90.85(7) & N/A(N/A) & 92.65(6) & 85.07(11) & 88.03(9) & 89.79(8) & 96.82(2) \\
    Cardiotocography & 54.77(16) & 81.24(12) & 89.88(3) & \textbf{91.17(1)} & 90.73(2) & 81.89(10) & 83.73(8) & 86.77(5) & 80.19(13) & 84.67(6) & 60.64(15) & 81.73(11) & 79.49(14) & 83.03(9) & 84.55(7) & 89.12(4) \\
\bottomrule
    \end{tabular}%
}
\end{table}

%% file: tables/AUCPR_label_0.05.tex
\begin{table}[h]
\scriptsize
  \centering
  \caption{AUCPR of 16 label-informed algorithms on \ndatasets benchmark datasets, with labeled anomaly ratio $\gamma_{l}=5\%$. We show the performance rank in parenthesis (lower the better), and mark the best performing method(s) in \textbf{bold}.
  }
\scalebox{0.55}{

    \begin{tabular}{l|lllllll|lllllllll}
    \toprule
    \textbf{Datasets}         & \textbf{GANomaly} & \textbf{DeepSAD} & \textbf{REPEN} & \textbf{DevNet} & \textbf{PReNet} & \textbf{FEAWAD} & \textbf{XGBOD} & \textbf{NB} & \textbf{SVM} & \textbf{MLP} & \textbf{ResNet} & \textbf{\makecell{FTTrans\\former}}  & \textbf{RF} & \textbf{LGB} & \textbf{XGB} & \textbf{CatB} \\
    \midrule
    ALOI  & 3.92(11) & 5.85(4) & 4.35(9) & 4.55(8) & 4.67(7) & 3.83(14) & \textbf{7.59(1)} & 3.49(15) & 3.39(16) & 3.88(12) & 4.02(10) & 3.85(13) & 5.88(3) & 4.77(6) & 6.00(2) & 5.45(5) \\
    annthyroid & 36.68(12) & 37.97(11) & 36.09(14) & 43.73(8) & 43.73(9) & 51.83(6) & 77.54(2) & 42.73(10) & 11.84(16) & 25.83(15) & 36.36(13) & 76.62(3) & 49.66(7) & 64.75(5) & 72.25(4) & \textbf{78.01(1)} \\
    backdoor & 7.05(16) & 62.79(7) & 28.48(15) & 88.48(3) & 88.58(2) & \textbf{88.99(1)} & 61.12(8) & 30.12(14) & 51.30(9) & 85.62(4) & 42.59(13) & 74.80(5) & 43.47(12) & 47.18(11) & 63.30(6) & 50.67(10) \\
    breastw & 85.59(12) & 83.65(13) & 95.32(4) & 98.65(2) & 92.91(8) & 93.09(7) & 91.96(9) & \textbf{99.49(1)} & 75.47(15) & 94.88(6) & 56.83(16) & 85.66(11) & 79.61(14) & 95.09(5) & 91.30(10) & 98.04(3) \\
    campaign & 23.50(14) & 26.68(10) & 15.33(16) & 39.15(6) & 39.32(5) & 29.98(8) & \textbf{47.53(1)} & 24.15(12) & 24.90(11) & 27.35(9) & 20.49(15) & 23.76(13) & 31.65(7) & 43.55(4) & 44.55(3) & 46.68(2) \\
    cardio & 39.02(15) & 45.81(13) & 82.12(4) & \textbf{85.56(1)} & 82.34(3) & 77.24(6) & 79.07(5) & 59.27(11) & 45.71(14) & 67.55(8) & N/A(N/A) & 76.95(7) & 64.82(9) & 57.80(12) & 63.75(10) & 83.34(2) \\
    Cardiotocography & 33.71(16) & 60.87(12) & 75.22(3) & \textbf{76.46(1)} & 75.97(2) & 64.45(8) & 62.98(10) & 73.90(4) & 55.11(14) & 65.29(6) & 45.15(15) & 63.22(9) & 58.32(13) & 61.57(11) & 64.77(7) & 70.42(5) \\
\bottomrule
    \end{tabular}%
}
\end{table}

%% file: tables/AUCROC_label_0.1.tex
\begin{table}[h]
\scriptsize
  \centering
  \caption{AUCROC of 16 label-informed algorithms on \ndatasets benchmark datasets, with labeled anomaly ratio $\gamma_{l}=10\%$. We show the performance rank in parenthesis (lower the better), and mark the best performing method(s) in \textbf{bold}.
  }
\scalebox{0.55}{

    \begin{tabular}{l|lllllll|lllllllll}
    \toprule
    \textbf{Datasets}         & \textbf{GANomaly} & \textbf{DeepSAD} & \textbf{REPEN} & \textbf{DevNet} & \textbf{PReNet} & \textbf{FEAWAD} & \textbf{XGBOD} & \textbf{NB} & \textbf{SVM} & \textbf{MLP} & \textbf{ResNet} & \textbf{\makecell{FTTrans\\former}}  & \textbf{RF} & \textbf{LGB} & \textbf{XGB} & \textbf{CatB} \\
    \midrule
    
    ALOI  & 56.59(7) & 56.63(6) & 54.54(8) & 49.55(13) & 49.50(14) & 53.11(9) & \textbf{72.10(1)} & 47.78(15) & 47.05(16) & 51.34(10) & 49.79(12) & 50.89(11) & 60.67(2) & 57.98(4) & 58.95(3) & 56.85(5) \\
    annthyroid & 82.83(11) & 87.41(9) & 82.05(14) & 83.31(10) & 82.78(12) & 93.07(7) & 98.21(3) & 82.54(13) & 40.43(16) & 80.53(15) & 90.88(8) & 98.67(2) & 93.72(6) & 97.39(5) & 97.56(4) & \textbf{98.85(1)} \\
    backdoor & 83.69(14) & 96.47(5) & 89.40(12) & \textbf{97.84(1)} & 96.39(6) & 96.73(3) & 96.17(7) & 79.46(15) & 94.03(9) & 95.82(8) & 58.20(16) & 93.93(10) & 87.71(13) & 93.93(11) & 97.22(2) & 96.69(4) \\
    breastw & 92.59(13) & 94.61(11) & 97.92(8) & 99.49(3) & 98.79(5) & 88.12(14) & 97.88(9) & \textbf{99.75(1)} & 79.54(15) & 99.54(2) & 60.44(16) & 98.18(7) & 93.69(12) & 99.01(4) & 97.43(10) & 98.78(6) \\
    campaign & 73.02(10) & 71.02(13) & 57.92(16) & 83.85(5) & 83.84(6) & 72.49(12) & \textbf{90.77(1)} & 75.62(9) & 68.88(14) & 72.51(11) & 62.26(15) & 79.61(8) & 80.15(7) & 88.25(4) & 88.72(3) & 90.08(2) \\
    cardio & 82.87(15) & 86.51(13) & 98.09(4) & 98.35(2) & 97.97(5) & 96.57(7) & 97.42(6) & 90.04(12) & 86.19(14) & 94.32(8) & N/A(N/A) & 98.27(3) & 91.53(11) & 93.39(10) & 94.30(9) & \textbf{98.46(1)} \\
    Cardiotocography & 55.12(16) & 86.65(10) & 91.80(4) & 93.15(2) & \textbf{93.33(1)} & 85.38(13) & 89.77(7) & 90.85(6) & 83.19(14) & 91.02(5) & 69.36(15) & 86.43(11) & 85.64(12) & 87.23(8) & 87.03(9) & 91.85(3) \\
\bottomrule
    \end{tabular}%
}
\end{table}

%% file: tables/AUCPR_label_0.1.tex
\begin{table}[h]
\scriptsize
  \centering
  \caption{AUCPR of 16 label-informed algorithms on \ndatasets benchmark datasets, with labeled anomaly ratio $\gamma_{l}=10\%$. We show the performance rank in parenthesis (lower the better), and mark the best performing method(s) in \textbf{bold}. 
  }
\scalebox{0.55}{

    \begin{tabular}{l|lllllll|lllllllll}
    \toprule
    \textbf{Datasets}         & \textbf{GANomaly} & \textbf{DeepSAD} & \textbf{REPEN} & \textbf{DevNet} & \textbf{PReNet} & \textbf{FEAWAD} & \textbf{XGBOD} & \textbf{NB} & \textbf{SVM} & \textbf{MLP} & \textbf{ResNet} & \textbf{\makecell{FTTrans\\former}}  & \textbf{RF} & \textbf{LGB} & \textbf{XGB} & \textbf{CatB} \\
    \midrule
    
    ALOI  & 3.92(12) & 6.36(5) & 5.12(9) & 4.65(10) & 5.30(7) & 3.70(14) & \textbf{8.15(1)} & 3.61(16) & 3.64(15) & 5.19(8) & 3.91(13) & 4.27(11) & 7.50(2) & 7.46(3) & 7.03(4) & 6.01(6) \\
    annthyroid & 45.28(11) & 49.95(9) & 40.81(14) & 46.43(10) & 44.91(12) & 63.56(7) & 80.54(2) & 43.62(13) & 8.28(16) & 40.68(15) & 54.35(8) & \textbf{82.40(1)} & 69.19(6) & 78.78(5) & 79.32(4) & 79.71(3) \\
    backdoor & 18.34(15) & 63.57(10) & 29.10(14) & \textbf{90.30(1)} & 89.60(2) & 77.61(4) & 65.85(8) & 11.04(16) & 64.49(9) & 79.78(3) & 33.28(13) & 73.24(6) & 62.71(11) & 66.54(7) & 74.83(5) & 57.75(12) \\
    breastw & 86.00(14) & 88.30(13) & 96.01(7) & 99.11(2) & 98.54(4) & 90.05(11) & 94.64(9) & \textbf{99.52(1)} & 80.51(15) & 98.93(3) & 66.65(16) & 94.55(10) & 89.89(12) & 97.84(5) & 95.06(8) & 96.69(6) \\
    campaign & 30.76(10) & 28.61(13) & 15.30(16) & 42.04(6) & 44.12(5) & 29.45(12) & \textbf{51.43(1)} & 33.02(9) & 27.11(14) & 29.75(11) & 23.69(15) & 33.84(8) & 39.61(7) & 47.00(4) & 47.80(3) & 50.86(2) \\
    cardio & 39.46(15) & 55.53(13) & 86.47(6) & \textbf{89.58(1)} & 87.31(5) & 88.78(2) & 83.34(7) & 69.42(11) & 48.91(14) & 75.61(9) & N/A(N/A) & 88.49(4) & 70.36(10) & 67.92(12) & 76.67(8) & 88.51(3) \\
    Cardiotocography & 34.00(16) & 69.51(9) & 81.18(2) & 80.91(3) & \textbf{82.19(1)} & 67.56(13) & 75.08(7) & 76.96(5) & 61.68(14) & 78.52(4) & 51.64(15) & 68.10(12) & 68.36(11) & 69.15(10) & 70.46(8) & 76.75(6) \\
\bottomrule
    \end{tabular}%
}
\end{table}

%% file: tables/AUCROC_label_0.25.tex
\begin{table}[h]
\scriptsize
  \centering
  \caption{AUCROC of 16 label-informed algorithms on \ndatasets benchmark datasets, with labeled anomaly ratio $\gamma_{l}=25\%$. We show the performance rank in  parenthesis (lower the better), and mark the best performing method(s) in \textbf{bold}.
  }
\scalebox{0.55}{

    \begin{tabular}{l|lllllll|lllllllll}
    \toprule
    \textbf{Datasets}         & \textbf{GANomaly} & \textbf{DeepSAD} & \textbf{REPEN} & \textbf{DevNet} & \textbf{PReNet} & \textbf{FEAWAD} & \textbf{XGBOD} & \textbf{NB} & \textbf{SVM} & \textbf{MLP} & \textbf{ResNet} & \textbf{\makecell{FTTrans\\former}}  & \textbf{RF} & \textbf{LGB} & \textbf{XGB} & \textbf{CatB} \\
    \midrule
    ALOI  & 55.25(8) & 59.29(6) & 53.09(11) & 51.65(13) & 51.44(14) & 57.54(7) & \textbf{75.96(1)} & 49.27(15) & 53.86(9) & 52.16(12) & 47.74(16) & 53.45(10) & 71.83(2) & 65.23(3) & 64.77(4) & 62.78(5) \\
    annthyroid & 80.96(15) & 93.06(9) & 82.78(12) & 82.39(13) & 82.11(14) & 96.95(7) & 98.78(4) & 82.95(11) & 50.73(16) & 91.29(10) & 96.08(8) & \textbf{99.05(1)} & 98.58(6) & 98.70(5) & 98.83(3) & 98.97(2) \\
    backdoor & 82.96(15) & 97.07(8) & 89.49(13) & 97.49(6) & 96.45(9) & 98.00(4) & 97.20(7) & 76.77(16) & 94.51(12) & 95.98(10) & 86.10(14) & 94.98(11) & 97.92(5) & 98.40(2) & 98.39(3) & \textbf{98.92(1)} \\
    breastw & 93.13(15) & 97.44(12) & 98.72(8) & 99.67(2) & 99.27(5) & 93.83(14) & 99.14(6) & \textbf{99.75(1)} & 94.57(13) & 99.42(3) & 80.40(16) & 98.45(10) & 98.23(11) & 99.30(4) & 98.72(9) & 99.11(7) \\
    campaign & 66.57(15) & 73.12(14) & 57.92(16) & 85.83(7) & 85.95(6) & 76.42(11) & \textbf{92.20(1)} & 79.15(9) & 73.59(13) & 77.29(10) & 75.45(12) & 84.62(8) & 87.85(5) & 90.41(3) & 89.16(4) & 91.23(2) \\
    cardio & 83.68(15) & 93.98(12) & 98.65(2) & 98.02(7) & 98.26(5) & 97.11(8) & 98.46(3) & 93.74(13) & 91.48(14) & 95.21(11) & N/A(N/A) & 98.24(6) & \textbf{98.73(1)} & 95.73(10) & 96.02(9) & 98.37(4) \\
    Cardiotocography & 56.41(16) & 92.52(12) & 94.43(6) & 94.91(3) & 94.92(2) & 90.40(13) & 93.81(7) & 93.20(10) & 87.28(14) & 94.80(4) & 82.53(15) & 94.48(5) & 93.34(9) & 93.50(8) & 92.89(11) & \textbf{95.13(1)} \\
\bottomrule
    \end{tabular}%
}
\end{table}

%% file: tables/AUCPR_label_0.25.tex
\begin{table}[h]
\scriptsize
  \centering
  \caption{AUCPR of 16 label-informed algorithms on \ndatasets benchmark datasets, with labeled anomaly ratio $\gamma_{l}=25\%$. We show the performance rank in  parenthesis (lower the better), and mark the best performing method(s) in \textbf{bold}.
  }
\scalebox{0.55}{

    \begin{tabular}{l|lllllll|lllllllll}
    \toprule
    \textbf{Datasets}         & \textbf{GANomaly} & \textbf{DeepSAD} & \textbf{REPEN} & \textbf{DevNet} & \textbf{PReNet} & \textbf{FEAWAD} & \textbf{XGBOD} & \textbf{NB} & \textbf{SVM} & \textbf{MLP} & \textbf{ResNet} & \textbf{\makecell{FTTrans\\former}}  & \textbf{RF} & \textbf{LGB} & \textbf{XGB} & \textbf{CatB} \\
    \midrule
    
    ALOI  & 3.83(16) & 6.13(6) & 4.54(9) & 4.27(11) & 4.19(12) & 4.71(8) & 10.55(2) & 3.97(14) & 5.67(7) & 4.30(10) & 3.92(15) & 4.07(13) & \textbf{13.19(1)} & 9.32(3) & 7.79(5) & 8.97(4) \\
    annthyroid & 41.61(15) & 62.35(9) & 42.15(14) & 44.72(12) & 45.01(11) & 69.16(8) & 80.46(6) & 44.24(13) & 16.63(16) & 58.00(10) & 70.09(7) & \textbf{83.26(1)} & 81.17(5) & 82.87(2) & 82.02(4) & 82.03(3) \\
    backdoor & 8.12(15) & 68.46(13) & 29.46(14) & \textbf{90.47(1)} & 90.23(2) & 74.19(10) & 71.48(12) & 6.55(16) & 73.80(11) & 76.47(8) & 74.24(9) & 78.55(7) & 87.34(4) & 82.65(6) & 85.95(5) & 87.85(3) \\
    breastw & 86.88(15) & 93.23(13) & 97.22(8) & 99.38(2) & 98.96(4) & 94.72(12) & 97.90(7) & \textbf{99.53(1)} & 92.66(14) & 99.00(3) & 82.92(16) & 95.91(11) & 96.94(10) & 98.46(5) & 96.98(9) & 97.96(6) \\
    campaign & 26.41(15) & 32.04(14) & 15.31(16) & 46.25(6) & 47.11(5) & 32.96(12) & \textbf{55.33(1)} & 36.83(9) & 32.30(13) & 34.32(10) & 33.18(11) & 38.30(8) & 48.60(4) & 49.83(3) & 45.65(7) & 52.73(2) \\
    cardio & 41.03(15) & 75.75(13) & \textbf{93.38(1)} & 87.72(7) & 88.74(4) & 88.74(3) & 88.19(6) & 77.51(12) & 58.80(14) & 82.56(9) & N/A(N/A) & 88.20(5) & 87.29(8) & 80.24(11) & 81.58(10) & 88.93(2) \\
    Cardiotocography & 35.04(16) & 80.55(10) & 84.14(5) & 84.58(2) & 84.51(3) & 75.01(13) & 82.53(6) & 80.64(8) & 64.33(15) & \textbf{85.69(1)} & 64.76(14) & 78.34(12) & 81.11(7) & 80.61(9) & 79.73(11) & 84.49(4) \\
\bottomrule
    \end{tabular}%
}
\end{table}

%% file: tables/AUCROC_label_0.5.tex
\begin{table}[h]
\scriptsize
  \centering
  \caption{AUCROC of 16 label-informed algorithms on \ndatasets benchmark datasets, with labeled anomaly ratio $\gamma_{l}=50\%$. We show the performance rank in parenthesis (lower the better), and mark the best performing method(s) in \textbf{bold}. 
  }
\scalebox{0.55}{

    \begin{tabular}{l|lllllll|lllllllll}
    \toprule
    \textbf{Datasets}         & \textbf{GANomaly} & \textbf{DeepSAD} & \textbf{REPEN} & \textbf{DevNet} & \textbf{PReNet} & \textbf{FEAWAD} & \textbf{XGBOD} & \textbf{NB} & \textbf{SVM} & \textbf{MLP} & \textbf{ResNet} & \textbf{\makecell{FTTrans\\former}}  & \textbf{RF} & \textbf{LGB} & \textbf{XGB} & \textbf{CatB} \\
    \midrule
    ALOI  & 54.52(11) & 64.14(6) & 53.63(12) & 52.88(14) & 52.95(13) & 57.95(7) & \textbf{79.27(1)} & 52.70(15) & 52.07(16) & 57.46(8) & 56.16(9) & 54.72(10) & 78.90(2) & 70.12(5) & 70.28(4) & 70.28(3) \\
    annthyroid & 76.85(16) & 95.54(9) & 83.31(11) & 82.81(13) & 82.58(14) & 91.43(10) & 99.21(5) & 83.01(12) & 82.03(15) & 96.62(8) & 98.13(7) & 98.97(6) & 99.34(3) & \textbf{99.38(1)} & 99.37(2) & 99.27(4) \\
    backdoor & 86.48(15) & 98.24(6) & 89.60(14) & 98.11(7) & 94.07(12) & 98.48(5) & 97.27(9) & 85.15(16) & 97.21(10) & 97.69(8) & 92.79(13) & 96.56(11) & \textbf{99.69(1)} & 99.23(4) & 99.31(3) & 99.64(2) \\
    breastw & 94.15(16) & 98.44(13) & 98.84(12) & 99.67(2) & 99.59(4) & 98.97(11) & 99.17(9) & \textbf{99.75(1)} & 98.32(14) & 99.62(3) & 95.52(15) & 99.10(10) & 99.19(8) & 99.42(5) & 99.40(6) & 99.38(7) \\
    campaign & 58.36(15) & 78.11(13) & 57.94(16) & 86.53(8) & 87.42(7) & 77.21(14) & \textbf{92.93(1)} & 81.14(11) & 78.45(12) & 82.95(10) & 84.07(9) & 89.47(6) & 91.25(5) & 92.44(3) & 91.77(4) & 92.86(2) \\
    cardio & 85.52(15) & 96.64(12) & 98.68(6) & 98.75(4) & 98.62(7) & 96.56(13) & \textbf{99.38(1)} & 94.69(14) & 97.54(10) & 96.88(11) & N/A(N/A) & 98.70(5) & 99.31(2) & 98.52(8) & 98.28(9) & 99.18(3) \\
    Cardiotocography & 58.27(16) & 94.69(11) & 95.80(6) & 95.00(10) & 95.22(9) & 94.30(12) & 96.00(4) & 93.07(14) & 94.29(13) & 95.79(7) & 89.09(15) & 95.52(8) & 96.23(3) & 96.42(2) & 95.95(5) & \textbf{96.53(1)} \\
\bottomrule
    \end{tabular}%
}
\end{table}

%% file: tables/AUCPR_label_0.5.tex
\begin{table}[h]
\scriptsize
  \centering
  \caption{AUCPR of 16 label-informed algorithms on \ndatasets benchmark datasets, with labeled anomaly ratio $\gamma_{l}=50\%$. We show the performance rank in parenthesis (lower the better), and mark the best performing method(s) in \textbf{bold}.
  }
\scalebox{0.55}{

    \begin{tabular}{l|lllllll|lllllllll}
    \toprule
    \textbf{Datasets}         & \textbf{GANomaly} & \textbf{DeepSAD} & \textbf{REPEN} & \textbf{DevNet} & \textbf{PReNet} & \textbf{FEAWAD} & \textbf{XGBOD} & \textbf{NB} & \textbf{SVM} & \textbf{MLP} & \textbf{ResNet} & \textbf{\makecell{FTTrans\\former}}  & \textbf{RF} & \textbf{LGB} & \textbf{XGB} & \textbf{CatB} \\
    \midrule
    
    ALOI  & 3.80(16) & 7.79(6) & 4.68(11) & 4.29(13) & 4.29(14) & 5.80(10) & 14.42(3) & 4.04(15) & 6.38(7) & 5.97(9) & 6.00(8) & 4.58(12) & \textbf{19.73(1)} & 12.29(4) & 12.26(5) & 15.21(2) \\
    annthyroid & 36.48(16) & 69.96(9) & 43.17(15) & 45.48(12) & 45.07(14) & 56.75(11) & 85.86(5) & 45.09(13) & 58.00(10) & 71.27(8) & 77.50(7) & 83.77(6) & 89.09(3) & \textbf{89.70(1)} & 89.60(2) & 86.92(4) \\
    backdoor & 32.07(14) & 85.28(11) & 29.70(15) & 90.44(7) & 89.20(8) & 90.83(5) & 72.87(13) & 17.69(16) & 90.70(6) & 85.44(10) & 86.98(9) & 83.55(12) & \textbf{97.24(1)} & 92.06(4) & 93.08(3) & 96.60(2) \\
    breastw & 88.43(16) & 96.18(13) & 97.24(12) & 99.33(2) & 99.12(4) & 98.43(8) & 97.60(11) & \textbf{99.52(1)} & 96.17(14) & 99.20(3) & 94.77(15) & 97.73(10) & 98.08(9) & 98.61(6) & 98.58(7) & 98.67(5) \\
    campaign & 17.93(15) & 37.50(13) & 15.35(16) & 46.79(8) & 48.68(6) & 33.08(14) & \textbf{58.11(1)} & 38.49(12) & 38.90(11) & 42.72(9) & 40.12(10) & 47.50(7) & 53.96(3) & 53.83(4) & 52.12(5) & 55.79(2) \\
    cardio & 44.50(15) & 86.44(12) & 93.47(2) & 91.40(7) & 91.59(6) & 87.30(11) & \textbf{93.69(1)} & 79.41(14) & 82.90(13) & 87.82(10) & N/A(N/A) & 91.72(5) & 93.46(3) & 87.83(9) & 89.77(8) & 93.07(4) \\
    Cardiotocography & 37.08(16) & 84.10(11) & 87.98(2) & 84.13(10) & 84.69(8) & 80.77(13) & 86.08(6) & 78.74(14) & 84.36(9) & 87.26(4) & 76.85(15) & 82.05(12) & 87.86(3) & 87.24(5) & 85.69(7) & \textbf{88.35(1)} \\
\bottomrule
    \end{tabular}%
}
\end{table}

%% file: tables/AUCROC_label_0.75.tex
\begin{table}[h]
\scriptsize
  \centering
  \caption{AUCROC of 16 label-informed algorithms on \ndatasets benchmark datasets, with labeled anomaly ratio $\gamma_{l}=75\%$. We show the performance rank in  parenthesis (lower the better), and mark the best performing method(s) in \textbf{bold}.
  }
\scalebox{0.55}{

    \begin{tabular}{l|lllllll|lllllllll}
    \toprule
    \textbf{Datasets}         & \textbf{GANomaly} & \textbf{DeepSAD} & \textbf{REPEN} & \textbf{DevNet} & \textbf{PReNet} & \textbf{FEAWAD} & \textbf{XGBOD} & \textbf{NB} & \textbf{SVM} & \textbf{MLP} & \textbf{ResNet} & \textbf{\makecell{FTTrans\\former}}  & \textbf{RF} & \textbf{LGB} & \textbf{XGB} & \textbf{CatB} \\
    \midrule
    ALOI  & 56.22(11) & 66.49(6) & 51.94(15) & 52.85(14) & 50.82(16) & 63.28(7) & \textbf{80.73(1)} & 53.69(12) & 58.62(9) & 59.71(8) & 57.35(10) & 53.21(13) & 79.83(2) & 74.51(3) & 73.38(5) & 73.46(4) \\
    annthyroid & 81.51(16) & 97.38(10) & 83.51(13) & 82.15(15) & 83.14(14) & 93.21(11) & 99.38(5) & 84.48(12) & 98.10(8) & 97.96(9) & 98.61(7) & 99.10(6) & 99.46(2) & 99.44(3) & \textbf{99.47(1)} & 99.42(4) \\
    backdoor & 86.89(16) & 98.69(6) & 89.62(15) & 97.91(9) & 94.93(13) & 98.56(8) & 97.52(10) & 93.90(14) & 98.56(7) & 98.88(5) & 96.33(12) & 96.96(11) & \textbf{99.95(1)} & 99.73(2) & 99.60(4) & 99.64(3) \\
    breastw & 95.04(16) & 99.22(14) & 99.46(12) & 99.73(5) & 99.74(4) & 98.76(15) & 99.68(10) & 99.69(9) & 99.55(11) & 99.75(3) & 99.39(13) & 99.70(8) & \textbf{99.79(1)} & 99.70(7) & 99.71(6) & 99.78(2) \\
    campaign & 66.24(15) & 81.12(13) & 58.00(16) & 87.18(9) & 88.34(7) & 76.61(14) & \textbf{93.54(1)} & 81.89(12) & 84.32(11) & 86.75(10) & 88.03(8) & 90.05(6) & 91.84(5) & 93.29(3) & 92.47(4) & 93.52(2) \\
    cardio & 87.07(15) & 98.39(11) & 99.27(7) & 98.80(9) & 98.56(10) & 97.84(12) & \textbf{99.57(1)} & 94.77(14) & 99.25(8) & 97.39(13) & N/A(N/A) & 99.42(4) & 99.48(2) & 99.35(5) & 99.34(6) & 99.47(3) \\
    Cardiotocography & 60.68(16) & 96.59(8) & 96.10(10) & 95.11(13) & 95.25(12) & 94.09(15) & 97.60(4) & 94.31(14) & 96.57(9) & 96.93(7) & 95.84(11) & 97.48(5) & 97.96(2) & 97.79(3) & 97.45(6) & \textbf{98.13(1)} \\
\bottomrule
    \end{tabular}%
}
\end{table}

%% file: tables/AUCPR_label_0.75.tex
\begin{table}[h]
\scriptsize
  \centering
  \caption{AUCPR of 16 label-informed algorithms on \ndatasets benchmark datasets, with labeled anomaly ratio $\gamma_{l}=75\%$. We show the performance rank in parenthesis (lower the better), and mark the best performing method(s) in \textbf{bold}. 
  }
\scalebox{0.55}{

    \begin{tabular}{l|lllllll|lllllllll}
    \toprule
    \textbf{Datasets}         & \textbf{GANomaly} & \textbf{DeepSAD} & \textbf{REPEN} & \textbf{DevNet} & \textbf{PReNet} & \textbf{FEAWAD} & \textbf{XGBOD} & \textbf{NB} & \textbf{SVM} & \textbf{MLP} & \textbf{ResNet} & \textbf{\makecell{FTTrans\\former}}  & \textbf{RF} & \textbf{LGB} & \textbf{XGB} & \textbf{CatB} \\
    \midrule
    ALOI  & 3.90(16) & 8.77(7) & 4.49(12) & 4.19(14) & 4.17(15) & 5.82(10) & 16.99(5) & 4.25(13) & 9.13(6) & 7.17(8) & 6.56(9) & 5.05(11) & \textbf{24.07(1)} & 18.92(3) & 18.05(4) & 19.77(2) \\
    annthyroid & 40.97(16) & 73.75(10) & 42.93(15) & 44.64(14) & 45.04(13) & 59.76(11) & 88.78(5) & 47.44(12) & 79.51(8) & 77.15(9) & 80.98(7) & 86.84(6) & \textbf{90.61(1)} & 89.80(4) & 90.59(2) & 90.44(3) \\
    backdoor & 32.90(15) & 92.07(7) & 29.77(16) & 90.62(9) & 88.94(12) & 92.35(6) & 75.37(14) & 81.09(13) & 94.44(5) & 91.79(8) & 89.33(11) & 89.82(10) & \textbf{98.71(1)} & 95.25(4) & 95.46(3) & 97.00(2) \\
    breastw & 89.92(16) & 98.37(15) & 98.79(13) & 99.47(5) & 99.48(4) & 98.80(12) & 99.29(8) & 99.43(6) & 98.95(11) & 99.53(3) & 98.60(14) & 99.28(9) & \textbf{99.57(1)} & 99.26(10) & 99.37(7) & 99.56(2) \\
    campaign & 28.98(15) & 41.70(12) & 15.38(16) & 48.78(8) & 49.98(7) & 34.27(14) & \textbf{60.85(1)} & 38.41(13) & 46.84(9) & 46.58(10) & 45.04(11) & 50.01(6) & 57.47(4) & 58.70(3) & 54.62(5) & 58.93(2) \\
    cardio & 49.13(15) & 91.99(9) & 95.64(3) & 91.60(11) & 91.91(10) & 91.06(13) & \textbf{96.20(1)} & 79.33(14) & 95.71(2) & 91.16(12) & N/A(N/A) & 95.27(5) & 95.36(4) & 94.32(8) & 94.87(7) & 95.11(6) \\
    Cardiotocography & 40.53(16) & 88.95(9) & 88.93(10) & 84.29(13) & 85.47(12) & 79.88(15) & 92.45(3) & 80.81(14) & 89.52(7) & 90.49(6) & 87.42(11) & 89.39(8) & \textbf{93.58(1)} & 92.04(4) & 91.23(5) & 93.36(2) \\
\bottomrule
    \end{tabular}%
}
\end{table}

%% file: tables/AUCROC_label_1.tex
\begin{table}[h]
\scriptsize
  \centering
  \caption{AUCROC of 16 label-informed algorithms on \ndatasets benchmark datasets, with labeled anomaly ratio $\gamma_{l}=100\%$. We show the performance rank in  parenthesis (lower the better), and mark the best performing method(s) in \textbf{bold}.
  }
\scalebox{0.55}{

    \begin{tabular}{l|lllllll|lllllllll}
    \toprule
    \textbf{Datasets}         & \textbf{GANomaly} & \textbf{DeepSAD} & \textbf{REPEN} & \textbf{DevNet} & \textbf{PReNet} & \textbf{FEAWAD} & \textbf{XGBOD} & \textbf{NB} & \textbf{SVM} & \textbf{MLP} & \textbf{ResNet} & \textbf{\makecell{FTTrans\\former}}  & \textbf{RF} & \textbf{LGB} & \textbf{XGB} & \textbf{CatB} \\
    \midrule
    ALOI  & 57.04(12) & 68.95(6) & 52.22(16) & 53.84(14) & 53.32(15) & 65.82(7) & 82.38(2) & 54.35(13) & 62.88(9) & 64.54(8) & 60.40(10) & 57.47(11) & \textbf{82.85(1)} & 79.37(3) & 76.55(5) & 78.54(4) \\
    annthyroid & 82.58(16) & 98.17(10) & 84.54(13) & 82.70(15) & 82.81(14) & 89.96(12) & 99.57(3) & 91.67(11) & 98.72(8) & 98.62(9) & 98.98(7) & 99.13(6) & \textbf{99.60(1)} & 99.56(4) & 99.57(2) & 99.55(5) \\
    backdoor & 88.57(16) & 99.54(5) & 89.65(15) & 98.32(12) & 95.39(14) & 98.73(9) & 99.20(7) & 97.53(13) & 98.62(10) & 99.69(4) & 98.94(8) & 98.50(11) & \textbf{99.97(1)} & 99.47(6) & 99.87(3) & 99.97(2) \\
    breastw & 97.28(16) & 99.82(9) & 99.01(14) & 99.72(11) & 99.77(10) & 99.10(13) & 99.90(5) & 98.87(15) & 99.66(12) & 99.82(8) & \textbf{99.95(1)} & 99.86(7) & 99.92(4) & 99.92(3) & 99.87(6) & 99.93(2) \\
    campaign & 57.15(16) & 84.72(12) & 58.03(15) & 87.35(11) & 88.73(9) & 79.81(14) & 93.74(3) & 82.61(13) & 87.84(10) & 89.10(8) & 89.30(7) & 91.27(6) & 92.79(5) & 93.79(2) & 93.35(4) & \textbf{93.93(1)} \\
    cardio & 87.96(15) & 99.20(10) & 99.42(9) & 98.96(11) & 98.87(12) & 98.21(13) & 99.81(2) & 97.64(14) & 99.71(4) & 99.44(8) & N/A(N/A) & \textbf{99.82(1)} & 99.73(3) & 99.64(5) & 99.49(7) & 99.61(6) \\
    Cardiotocography & 64.62(16) & 96.95(9) & 96.09(11) & 94.82(13) & 95.38(12) & 94.44(14) & 98.14(6) & 93.75(15) & 96.87(10) & 97.21(8) & 98.01(7) & 98.20(5) & 98.33(3) & 98.54(2) & 98.28(4) & \textbf{98.57(1)} \\
\bottomrule
    \end{tabular}%
}
\end{table}

%% file: tables/AUCPR_label_1.tex
\begin{table}[h]
\scriptsize
  \centering
  \caption{AUCPR of 16 label-informed algorithms on \ndatasets benchmark datasets, with labeled anomaly ratio $\gamma_{l}=100\%$. We show the performance rank in parenthesis (lower the better), and mark the best performing method(s) in \textbf{bold}. 
  }
\scalebox{0.55}{

    \begin{tabular}{l|lllllll|lllllllll}
    \toprule
    \textbf{Datasets}         & \textbf{GANomaly} & \textbf{DeepSAD} & \textbf{REPEN} & \textbf{DevNet} & \textbf{PReNet} & \textbf{FEAWAD} & \textbf{XGBOD} & \textbf{NB} & \textbf{SVM} & \textbf{MLP} & \textbf{ResNet} & \textbf{\makecell{FTTrans\\former}}  & \textbf{RF} & \textbf{LGB} & \textbf{XGB} & \textbf{CatB} \\
    \midrule
    
    ALOI  & 3.92(16) & 9.31(7) & 4.57(12) & 4.30(13) & 4.19(15) & 6.02(10) & 18.12(5) & 4.21(14) & 11.46(6) & 9.06(8) & 7.68(9) & 5.01(11) & \textbf{31.35(1)} & 26.05(2) & 23.15(4) & 25.44(3) \\
    annthyroid & 45.77(13) & 78.04(10) & 45.07(15) & 45.35(14) & 44.64(16) & 53.95(12) & 93.23(2) & 60.64(11) & 80.97(8) & 79.61(9) & 85.09(7) & 86.50(6) & \textbf{93.25(1)} & 92.97(3) & 92.44(4) & 92.36(5) \\
    backdoor & 39.08(15) & 96.79(5) & 29.75(16) & 91.06(11) & 89.83(12) & 91.65(10) & 85.58(13) & 84.92(14) & 95.00(8) & 95.84(7) & 96.30(6) & 92.72(9) & \textbf{99.33(1)} & 98.22(3) & 97.87(4) & 99.33(2) \\
    breastw & 94.78(16) & 99.66(9) & 97.83(14) & 99.46(11) & 99.56(10) & 98.91(13) & 99.81(5) & 96.36(15) & 99.22(12) & 99.67(8) & \textbf{99.90(1)} & 99.71(7) & 99.83(4) & 99.85(3) & 99.74(6) & 99.85(2) \\
    campaign & 20.82(15) & 46.45(12) & 15.39(16) & 49.78(11) & 50.98(9) & 37.76(14) & \textbf{61.45(1)} & 38.67(13) & 51.95(7) & 51.60(8) & 50.33(10) & 54.05(6) & 60.31(3) & 59.83(4) & 58.63(5) & 60.96(2) \\
    cardio & 53.07(15) & 95.14(10) & 96.27(9) & 92.91(13) & 93.03(12) & 94.60(11) & 98.46(2) & 81.01(14) & 97.85(3) & 96.40(8) & N/A(N/A) & \textbf{98.65(1)} & 97.79(4) & 97.41(5) & 96.96(7) & 97.40(6) \\
    Cardiotocography & 44.34(16) & 90.49(9) & 88.97(11) & 82.95(14) & 85.97(12) & 83.16(13) & 94.35(5) & 79.80(15) & 89.57(10) & 91.69(8) & 93.19(7) & 94.66(4) & 95.12(3) & 95.21(2) & 93.50(6) & \textbf{95.38(1)} \\
\bottomrule
    \end{tabular}%
}
\end{table}